# Synthetic Data Generation for Augmenting Small Samples

Dan Liu[1,2], Samer El Kababji[1,2], Nicholas Mitsakakis[1], Lisa Pilgram[1,2,3], Thomas Walters[4], Mark Clemons[5,6], Greg Pond[7], Alaa El-Hussuna[8], Khaled El Emam[1,2]

[1]*CHEO Research Institute, Ottawa, Canada*
[2]*University of Ottawa, Ottawa, Canada*
[3]*Department of Nephrology and Medical Intensive Care, Charité - Universitaetsmedizin Berlin, Berlin, Germany*
[4]*Hospital for Sick Children, Toronto, Canada*
[5]*Ottawa Hospital Research Institute, Ottawa, Canada*
[6]*Division of Medical Oncology, Department of Medicine, University of Ottawa, Ontario, Canada*
[7]*McMaster University, Hamilton, Ottawa*
[8]*OpenSourceResearch, Aalborg, Denmark*

Corresponding Author:
Khaled El Emam
Children's Hospital of Eastern Ontario Research Institute
401 Smyth Road
Ottawa, Ontario K1H 8L1
Canada

kelemam@ehealthinformation.ca

# Abstract

**Background:** Small datasets are common in health research. However, the generalization performance of machine learning models is suboptimal when the training datasets are small. To address this, data augmentation is one solution and is often used for imaging and time series data, but there are no evaluations on its potential benefits for tabular health data. Augmentation increases sample size and is seen as a form of regularization that increases the diversity of small datasets, leading them to perform better on unseen data.

**Objectives:** Evaluate data augmentation using generative models on tabular health data and assess the impact of diversity versus increasing the sample size.

**Methods:** Using 13 large health datasets, we performed a simulation to evaluate the impact of data augmentation on the prediction performance (as measured by the AUC) on binary classification gradient boosted decision tree models. Four different synthetic data generation models were evaluated. We also developed a decision support model to help analysts determine if augmentation will improve model performance and demonstrate the application of the decision model for augmentation on seven small real datasets. A comparison of augmentation with resampling (which is a proxy for a larger dataset with minimal impact on diversity) was performed.

**Results:** Augmentation improves prognostic performance for datasets that: have fewer observations, with smaller baseline AUC, have higher cardinality categorical variables, and have more balanced outcome variables. No specific generative model consistently outperformed the others. Our decision support model had an AUC of 0.77 and can be used to inform analysts if augmentation would be useful. For the seven small application datasets, augmenting the existing data results in an increase in AUC between 4.31% (AUC from 0.71 to 0.75) and 43.23% (AUC from 0.51 to 0.73), with an average 15.55% relative improvement, demonstrating the nontrivial impact of augmentation on small datasets (p=0.0078). Augmentation AUC was higher than resampling only AUC (p=0.016). The diversity of augmented datasets was higher than the diversity of resampled datasets (p=0.046).

**Conclusions:** This study demonstrates that data augmentation using generative models can have a marked benefit in terms of improved predictive performance for machine learning models, but only for datasets that meet baseline data size and complexity criteria. Our decision model can help analysts decide if augmentation can be useful, and if it is, then they can follow our recommended method for finding the appropriate level of augmentation to apply. Furthermore, augmentation performed better than having a larger dataset, which is consistent with the argument that greater data diversity due to augmentation is beneficial.

# 1. Introduction

Many machine learning (ML) clinical prediction models are being trained on datasets that are too small. Specifically, a median of 12.5 events per predictor variable has been reported in the literature [1] and 1.7 for oncology ensemble models [2]. However, to achieve stability while training ML models more than 200 events per predictor variable are often required [3], and the vast majority of ML modeling studies in oncology did not meet the minimum recommended sample sizes [4].

To address this data scarcity problem, there is a growing interest in using data augmentation to simulate additional observations from existing data [5]. This augmentation process increases the sample size of the dataset, which by itself is expected to improve ML model prognostic performance [3]. Augmentation can also be seen as a form of regularization [6], where the simulated data increase the diversity of the original dataset by generating more and different examples from the same population. Therefore, augmentation could improve the prediction accuracy on the unseen data and enable ML models trained on augmented data to achieve better generalization performance.

Data augmentation is a common practice for imaging data [5,7,8] where additional synthetic imaging samples can boost model accuracy [9–17]. Augmentation has also been applied to time series datasets [18,19]. However, there is a dearth of comprehensive evaluations, applications of, and guidance on augmentation methods for tabular data. Tabular data are ubiquitous in practice, including in the health domain [20].

Augmentation techniques have been applied to address different types of data scarcity problems. Oversampling methods are commonly used, such as the synthetic minority over-sampling technique (SMOTE) [21] and a few variants of SMOTE [22–25], and these enlarge the original data by interpolation. However, they are typically applied in the case of outcome imbalance. When there is covariate imbalance, with certain groups under-represented in the data, generative models have been used to mitigate the representation bias that is introduced [26]. For augmenting overall records, methods such as sampling with replacement, sequential synthesis using decision trees [27], generative adversarial networks (GAN) [28], and variational autoencoders [29] have been evaluated with encouraging results [30–34], although some deep learning architectures were found to be unstable [35]. Augmentation methods have also been applied to small clinical trial datasets [36–38].

In the current paper, we develop and evaluate an augmentation scheme for tabular health data. Specifically, we make two contributions represented as the two parts of the study. First, we examine the augmentation performance of four synthetic data generation (SDG) methods on the predictive performance of ensemble ML models using binary gradient boosted decision trees (GBDTs) on a heterogeneous set of 13 datasets, develop a decision support tool to recommend whether augmentation would be useful, and define a process to find the level of augmentation that would maximize performance if the recommendation is positive. Second, we test the benefit of diversity by examining seven different datasets that are deemed good candidates for augmentation and evaluate whether greater data diversity due to augmentation improves the GBDT model performance over simply increasing the sample size.

## 2. Methods

Our study consists of a simulation and evaluation of the extent to which augmentation can improve the predictive performance of GBDTs by answering the following questions:

Q-1. Is augmentation always beneficial or are there situations where augmentation does not add value or is even detrimental to machine learning model performance?

Q-2. Is there a type of generative model that consistently produces better data augmentation outcomes ?

Q-3. Is the value of augmentation due to larger dataset sizes or increased data diversity ?

### 2.1 Overview of Simulation and Evaluation Processes

The overall workflows for part 1 is shown in Figure 1, and part 2 in Figure 2.

For part 1, we start with a large population dataset P and randomly split it into a training dataset T and a test dataset P\T with a 70:30 split for train:test. The test set represents unseen patients that we would use to evaluate augmented data on.

We then draw a simple random sample (step A) of size $n_0$ from the training dataset (the *base* dataset), which is augmented using a generative model, also called a synthesizer (step B), with a set of additional n' records. The augmented dataset of size $n = n_0 + n'$ (step C) is used to train a binary GBDT model (step D) [39]. Tree-based models are the most common type of ML prognostic methods used in clinical research [1], they perform better than linear models, such as logistic regression [40–44], and have also been found to perform better than deep learning models on tabular datasets [45,46]. The performance of that trained model is evaluated on the test dataset using the AUC (step E). This process is repeated for multiple values of $n_0$.

Since augmentation does not always positively contribute to prognostic model performance improvement, a decision support model is required to allow end-users to decide whether to attempt augmentation or not for a particular dataset. The decision support model would save analysts significant time and resources in augmentation if it will not add value. Using all the data generated from these simulations (step G) as well as specific characteristics of each dataset (step F), we train another binary classification model that predicts whether augmentation for any dataset is likely to improve prognostic performance (step H).

Seven new datasets with realistic sizes seen in clinical research are used to illustrate the application of the decision support model and augmentation. In all cases the decision support model is used to recommend whether generative models should be used to augment the datasets. We can then evaluate predictive performance improvement that one can expect to see through augmentation (comparing N vs Q).

We take the same seven datasets and augment them using sampling with replacement (step J). This is intended to increase the sample size but not impact data diversity (comparing M vs. P). By comparing the predictive performance of GBDT on augmented datasets with higher diversity to that of resampled datasets with minimal impact on diversity, we can determine whether any benefits from augmentation are due to diversity or due to larger sample sizes (comparing L vs N).

In the remainder of the methods section we provide the details of these steps.

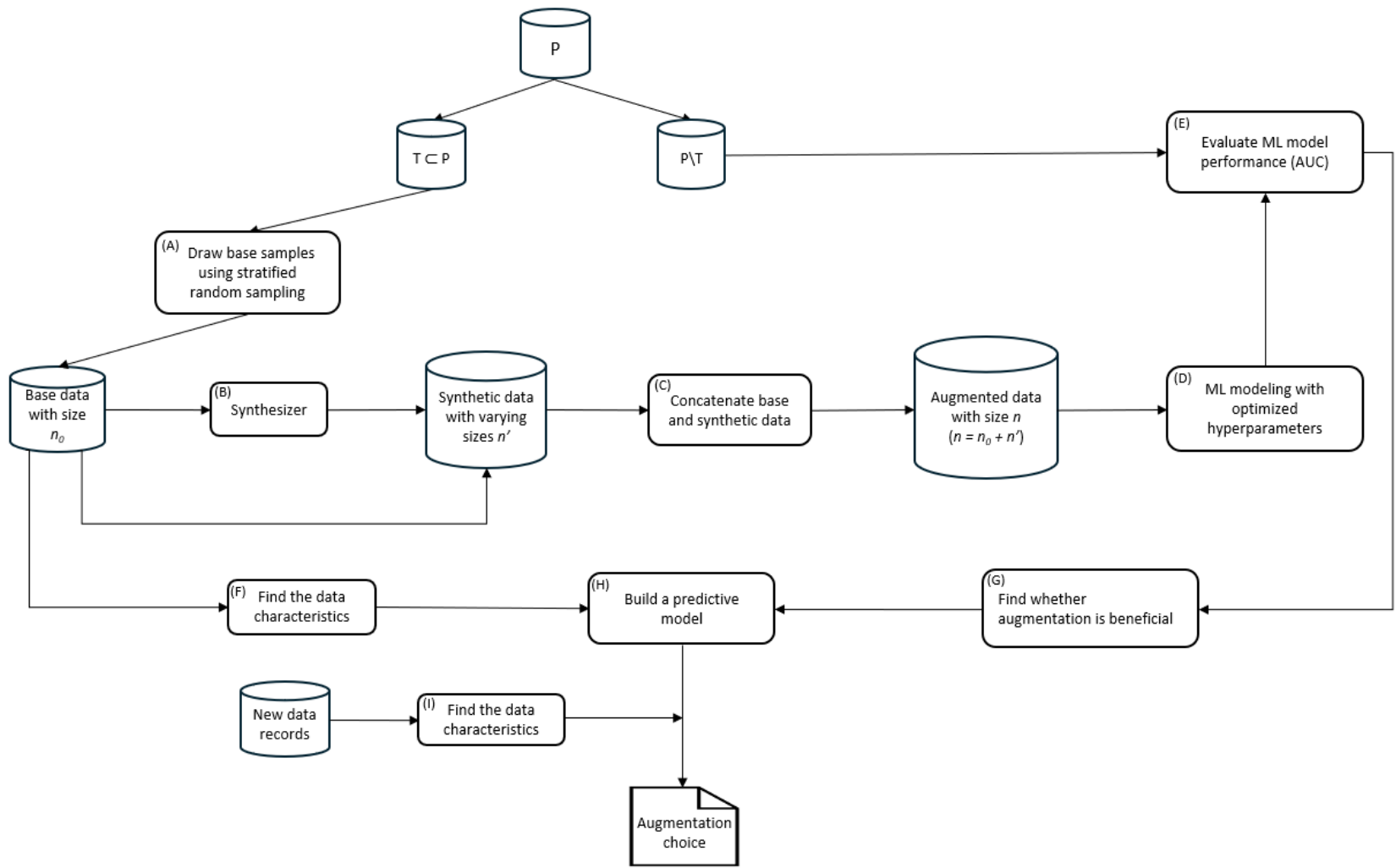

**Figure 1:** The methods workflow for part 1 of the study.

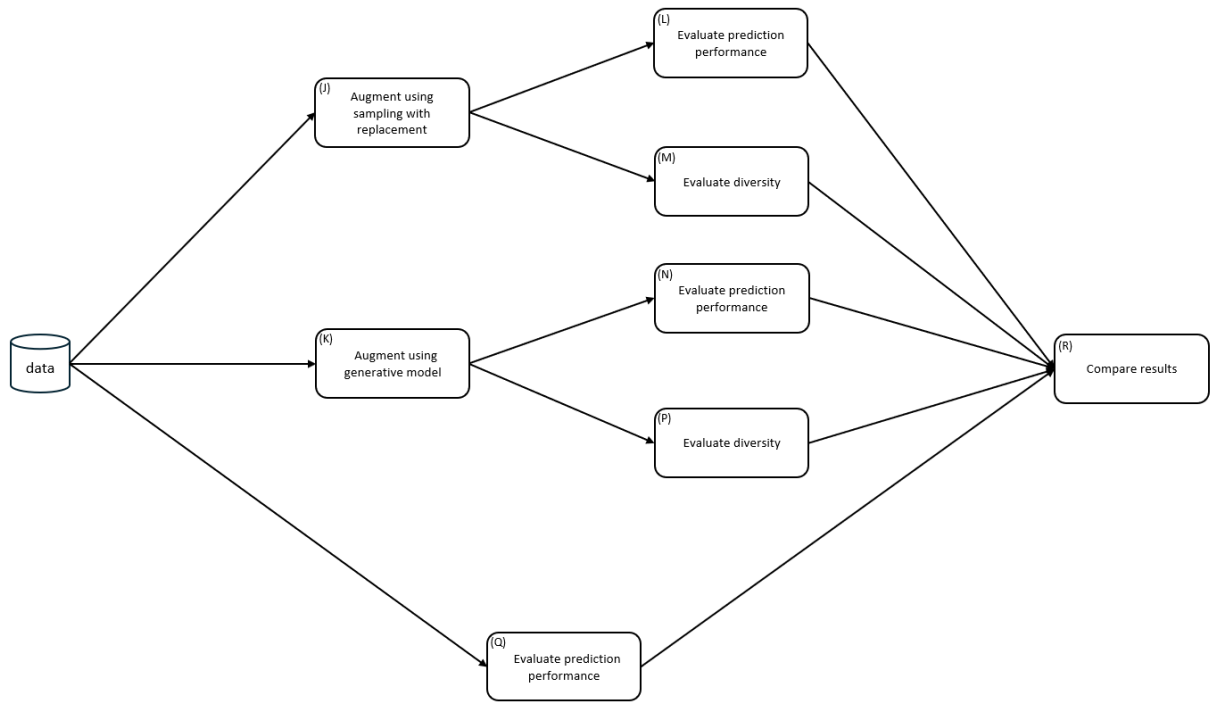

**Figure 2:** The methods workflow for part 2 of the study.

## 2.2 Datasets

We have two sets of data corresponding to the two parts of our study.

The population real world datasets that were used are summarized in Table 1A. These datasets cover heterogeneous domains, including public health, hospital discharge, infant and maternal health, adverse events, ICU, population health surveys, and insurance claims. The table provides an overview of the datasets, the original number of observations, the number of observations after removing those with any missing values in the outcome variable and the number of variables included in the binary classification models used to predict the outcome. A detailed description of each preprocessed dataset and the binary workload used for modeling can be found in the appendix. The number of predictor variables in the workloads is consistent with what is seen in the clinical research literature [1].

For the second part, we show the seven smaller datasets that we use for our application case studies and comparisons in Table 1B.

| Dataset | Description | No. observations (original) | No. observations* | No. Variables used in the analysis |
|---|---|---|---|---|
| COVID (COVID) | COVID-19 health records of Canadians collected by Esri Canada | 1,384,881 | 745,623 | 7 |
| Canadian Community Health Survey (CCHS) | A pooled version of survey data across multiple years that gathers health information for the Canadian population | 904,813 | 752,472 | 8 |
| COVID Survival (Nexoid) | A web-based survey dataset on COVID-19 survival prediction collected by the Nexoid company in London, UK | 968,408 | 968,394 | 19 |
| FDA Adverse Event Reporting System (FAERS) | Adverse event and medication error reports submitted to FDA | 881,204 | 251,409 | 7 |
| Texas Inpatient Data (Texas) | Discharges from Texas hospitals | 745,999 | 745,997 | 11 |
| Washington State Hospital Discharge (Washington2007) | Hospital discharge information from the HCUP state inpatient database for 2007 | 644,902 | 644,901 | 8 |
| Basic Stand Alone (BSA) Inpatient Claims | Claim-level information from 2008 Medicare inpatient claims | 588,415 | 588,415 | 6 |
| Washington State Hospital Discharge (Washington2008) | Hospital discharge information from the HCUP state inpatient database for 2008 | 652,340 | 652,340 | 18 |
| California Hospital Discharge (California2007) | Hospital discharge information from the HCUP state inpatient database for 2007 | 4,016,573 | 4,016,573 | 16 |
| Florida Hospital Discharge (Florida2007) | Hospital discharge information from the HCUP state inpatient database for 2007 | 2,327,563 | 2,327,563 | 12 |
| New York Hospital Discharge (New York2007) | Hospital discharge information from the HCUP state inpatient database for 2007 | 2,666,541 | 2,666,541 | 14 |
| Better Outcomes Registry & Network (BORN) | A population registry containing comprehensive perinatal, newborn and child information in Ontario | 963,083 | 963,083 | 18 |
| Medical Information Mart for Intensive Care III (MIMIC-III) | Health-related information for patients who stayed in critical care units of the Beth Israel Deaconess Medical Center between 2001 and 2012 | 540,482 | 540,482 | 13 |

* After data transformation / removing observations with missing values on the outcome variable.

**Table 1A:** A description of the thirteen datasets used in the first simulation part of the study.

| Dataset | Description | No. observations (original) | No. observations* | No. Variables used in the analysis |
|---|---|---|---|---|
| Hot Flashes | A survey contains health information related to vasomotor symptoms for early breast cancer patients between 2020 and 2021 | 373 | 360 | 18 |
| Danish Colorectal Cancer Group (DCCG) | Registry of all patients with colorectal cancer in Denmark since 2001 | 12,855 | 7,948 (700**) | 11 |
| Breast Cancer Coimbra | Registry of women with breast cancer in Portugal between 2009 and 2013 | 116 | 116 | 10 |
| Breast Cancer | Health information related to breast cancer recurrence in Yugoslavia | 277 | 277 | 11 |
| Colposcopy/Schiller | One of three modality dataset related to subjective quality assessment of digital colposcopies | 92 | 92 | 63 |
| Diabetic Retinopathy | Messidor image information related to signs of diabetic retinopathy | 1151 | 600 | 20 |
| Thoracic Surgery | Post-operative life expectancy of patients who went through surgery for lung cancer between 2007 and 2011 | 470 | 470 | 17 |

\* After data transformation / removing observations with missing values on the outcome variable.
\*\* The sample drawn for the evaluation which is different from the full clean dataset.

**Table 1B:** A description of the seven datasets used in the case studies and evaluations in the second part of the study.

## 2.3  Augmentation Scheme

Given a population dataset, the first step is to split it into training and testing datasets, where the training dataset is used for subsequent sampling, augmentation and ML modeling, while the testing data is retained for performance evaluation. In our augmentation scheme, outcome stratified random sampling is applied to draw 40 samples (base datasets) of sizes $n_0$ without replacement, from the training data, where $n_0 \in \{20, 30, 40, 50, 60, 70, 80, 90, 100, 150, 200, 250, 300, 350, 400, 450, 500, 550, 600, 650, 700, 750, 800, 850, 900, 950, 1000, 2000, 3000, 4000, 5000, 6000, 7000, 8000, 9000, 10000, 20000, 30000, 40000, 50000\}$. Then, each of the 40 base datasets is used to train a specific generative model. Subsequently, the synthetic records were simulated from that generative model with sizes according to the following geometric series. Let $b \sim N(\mu = 1.5, \sigma^2 = 0.005)$ be a random variable that follows a normal distribution with a mean of 1.5 and a standard deviation of 0.005. The geometric series has more samples at low values and less at higher values as we expect there will be more variability at the lower end of the range.

A series contains 30 elements, and each element represents a size of synthetic datasets to be generated, denoted as $n_i' = [b^{i+4}]$ (i = 1, …, 30), where [x] denotes rounding to the closest integer to x. Following this procedure, a total of 10 geometric series were created. The augmented dataset has a size of $n_i = n_0 + n_i'$ which means that for each of the 40 values of $n_0$, a total of 300 augmented datasets are generated of different sizes (i.e., different degrees of augmentation). Each of the augmented datasets is used to train an ML model. To ensure the comparability of the results, the same testing dataset is used for all the augmented datasets for evaluation. In total 12,000 augmented datasets were therefore generated and evaluated for each of the 13 datasets.

## 2.4 Machine Learning Analytic Workload

In this study, the chosen workload ML model is a light gradient boosting machine (LGBM) [39]. Model tuning used 5-fold cross-validation and Bayesian optimization [47]. The range for the tuning parameters was previously suggested [48–51], and these are summarized in the appendix. High cardinality variables were converted to embeddings [52] using a scheme similar to target encoding.

## 2.5 Synthetic Data Generation Methods

We used four commonly applied generative modeling methods to generate new observations for structured tabular data, namely, sequential decision trees [27,53–55], Bayesian networks [56–59], conditional generative adversarial network [60] and variational autoencoders [60]. The first method was implemented using Aetion® Generate, a commercial product from Aetion[1], and the last three methods were implemented using an open-sourced Python package Synthcity [61]. Our implementation, which is publicly available, provides further pre-processing and post-processing on top of Synthcity. In the experiments, the variables to be synthesized in each dataset are only those that were used in the analysis (last column in Table 1).

### 2.5.1 Sequential Decision Trees

Similar to using a chaining method for multi-label classification problems, sequential decision trees (SEQ) generate synthetic data using conditional trees in a sequential fashion [27,62,63]. It has been commonly employed in the healthcare and social science domains for data synthesis [30,53,54,64–69]. The details of the implementation procedures can be referred to [27].

---

[1]See <https://aetion.com/products/generate/>

10/34

### 2.5.2 Bayesian Networks

Bayesian Networks (BN) are models based on Directed Acyclic Graphs that consist of nodes representing the random variables and arcs representing the dependencies among these variables. To construct the BN model, the first step is to find the optimal network topology, and then to estimate the optimal parameters [56]. Starting with a random initial network structure, the Hill Climb heuristic search is used to find the optimal structure. Then, the conditional probability distributions are estimated using the maximum a posteriori estimator [70]. Once the network structure and the parameters are estimated, we can initialize the nodes with no incoming arcs by sampling from their marginal distributions and predict the rest of the connected variables using the estimated parameters.

### 2.5.3 Conditional Generative Adversarial Network

A basic generative adversarial network (GAN) consists of two artificial neural networks (ANNs), a generator and a discriminator [28]. The generator and the discriminator play a min-max game. The input to the generator is noise, while its output is synthetic data. The discriminator has two inputs: the real training data and the synthetic data generated by the generator. The output of the discriminator indicates whether its input is real or synthetic. The generator is trained to 'trick' the discriminator by generating samples that look real. On the other hand, the discriminator is trained to maximize its discriminatory capability.

Among all the variations of GAN architectures, the conditional tabular GAN (CTGAN) is often used in tabular data synthesis [71]. CTGAN builds on conditional GANs by addressing the multimodal distributions of continuous variables and the highly imbalanced categorical variables [60]. CTGAN solves the first problem by proposing a per-mode normalization technique. For the second problem, each category of a categorical variable serves as the condition passed to the GAN.

### 2.5.4 Variational Autoencoder

Variational autoencoders (VAE) use ANNs and involve two steps (encoding and decoding) to generate new samples [29]. First, an encoder is generated to compress input data into a lower-dimensional latent space, in which the data points are represented by distributions. The second step is a decoding process, in which new data samples are reconstructed as output from the latent space. The neural network is optimized by minimizing the reconstruction loss between the output and the input. VAEs are known to generate complex data of various types due to its ability to learn more complex distributions [72]. Many variants have been proposed as an extension of VAE, such as triplet-based VAE [73], conditional VAE [74], and Gaussian VAE [75]. In particular, the tabular VAE (TVAE) was proposed as an adaption of standard VAE to model and generate mixed-type tabular data with a modified loss function [60].

## 2.6   Decision Support Model for Augmentation

Based on the characteristics of the input base datasets, a decision support model was trained to recommend whether extra synthetic data should be simulated and added to the base dataset (the "augmentation recommendation"). The decision support model would be used by an analyst to determine whether augmentation is likely to improve the performance of their prognostic ML model. Given a dataset, if the decision model recommends augmentation, then the four generative models would be used to create the additional synthetic data for multiple values of $n'$, and the augmented dataset with the highest AUC gain would then be chosen. If the decision model does not recommend augmentation, then the analyst can save resources as augmentation is not likely to be beneficial.

The outcome for this decision support model was determined by examining all of the simulation results for each $n_0$ value for every dataset and every generative model, and a binary value was selected to indicate that for this {$n_0$, dataset, generative model} combination augmentation improved AUC over the baseline (a one outcome) or not (a zero outcome). This resulted in 520 observations.

Whether a dataset will benefit from a certain amount of augmentation will depend on its complexity. For example, a simple dataset, which conceptually can mean a small dataset with few low cardinality categorical variables, is unlikely to have a marked increase in diversity after augmentation. This is because the space of possible values on the categorical variables is small. Whereas a more complex dataset with many high cardinality variables is likely to experience much more increases in diversity with augmentation, and hence would perform better on unseen data.

Previous work on data complexity metrics [76,77] and methods for sample size calculation that take data complexity into account [78,79] have defined a set of metrics that we considered for our augmentation decision model. We propose that dataset complexity can be characterized by the following variables: the base dataset size $n_0$, the number of predictor parameters, outcome distribution, standardized entropy, mutual information, separability measure and the AUC of the base dataset. These additional variables are defined as:

- **Base dataset size $n_0$.** The number of records in the original dataset.

- **Degrees of freedom.** This is given a value of 1 for a numeric variable, and a categorical variable with k levels gives k – 1.

- **Imbalance factor.** The outcome distribution is represented by the imbalance. It describes the imbalance between the positive and negative classes in the binary outcome and is quantified as the maximum of prevalence/(1 – prevalence) and (1 – prevalence)/prevalence, where prevalence is the proportion of individuals who have a positive outcome. A lower imbalance factor implies a more balanced distribution of outcome classes in the dataset.

- **Standardized entropy.** This is calculated as the information for each predictor and the whole dataset. We take the mean of the standardized entropy across all predictors to reflect the average amount of information produced by the variables.

- **Mutual information.** This is the coefficient of variation across the mutual information calculated across all predictor pairs.

- **The separability measure.** This is defined as the ratio of the distance of intraclass nearest neighbors to the distance of interclass nearest neighbors to reflect the magnitude of distinguishability between two samples from different classes. To accommodate various types of variables for the intraclass and interclass distances, we further modify this measure by replacing the Euclidean distance with the Gower distance.

The simulated data are clustered, with the dataset constituting the clustering factor. While modeling methods for clustered data would be a natural choice (e.g., mixed effects models), their benefits when prediction (and not explanation) is the modeling objective are questionable [80].

Several combinations of characteristics were evaluated and compared for variable selection, and the baseline AUC was found to be the most impactful feature. In addition, the combination of $n_0$, the imbalance factor, the degrees of freedom for the predictors and the baseline AUC yielded the best results. Thus, these four characteristics were computed for each of the base datasets and then taken as input features to construct the decision support model for the augmentation recommendation. The model is a binary prediction to recommend whether augmentation should be performed or not (with 1 being augmentation is recommended and zero not). The binary outcome was determined by examining the maximum AUC from the simulation results for each generative model by dataset combination. Augmentation is recommended if the predicted probability from the decision model is greater than 0.5.

Five types of predictive models were considered to develop the decision model: logistic regression (LR), random forest (RF), support vector machine (SVM), light gradient boosting machine (LGBM), and extreme gradient boosting (XGB). All the models were tuned and optimized using Bayesian optimization [47]. The range for the tuning parameters was previously suggested [48–51]. Two metrics were utilized to evaluate and compare the performance of the models, 1) the AUC, which is computed based on the true classes and probabilities of the predicted classes, and 2) the prediction accuracy, which is measured by the overall accuracy of correctly predicting the true classes.

We adopted leave-one-out cross-validation (LOOCV) to find the model with the optimal performance [81]. In this implementation we are not leaving out observations but datasets. The implementation of LOOCV involves splitting the 13 datasets into 12 datasets for model training and the remaining dataset for testing in one iteration, in which the model is trained on 12 training datasets and predicts the probabilities using the testing dataset. Then, the AUC and accuracy of the predicted classes are calculated for the testing dataset. This process is repeated 13 times until each dataset has been used as the testing set once. The final evaluation results are obtained by averaging the AUC and prediction accuracy across the 13 iterations.

## 2.7 Evaluation of Augmentation

We illustrate the proposed decision support model and augmentation method by applying it in real situations with small datasets and assess whether this results in an improvement in the performance of the ML model.

Seven real datasets were used: the Hot Flashes dataset, Danish Colorectal Cancer Group dataset, Breast Cancer Coimbra dataset, Breast Cancer dataset, Colposcopy/Schiller dataset, Diabetic Retinopathy dataset and Thoracic Surgery dataset. These datasets vary across dimensions and complexity. The detailed descriptions of the datasets are summarized in the Appendix.

A nested 5-fold cross-validation (CV) approach was applied for model training and prediction, which has been shown to yield almost unbiased estimates [82–84]. Each original dataset was first preprocessed for the recommendation of augmentation from the decision support tool and if that was positive it randomly divided into 5-folds of training and testing sets. The characteristics of the analysis dataset were measured, including the sample size, imbalance factor and degrees of freedom. The baseline AUC was determined as the average value of AUC obtained from the 5-fold training sets. We train an LGBM model to examine the association between the outcome of interest and the data complexity measures. The hyperparameters of LGBM models were tuned and optimized using Bayesian optimization [47]. The range for the tuning parameters, specific to each model, was previously suggested [48–51]. Note that augmentation was performed separately for each training partition in the outer loop to avoid data leakage that would result in optimistic model performance. A range of values for *n'* from 7 to 1 million was evaluated and remained the same for each iteration. The final AUC result was the averaged value of AUC across five iterations of the outer loop, and the *n'* value that provided the maximum AUC was deemed optimal.

To assess the improvements in AUC from the augmented datasets relative to the original datasets, we performed an exact permutation one-tailed test for the mean paired difference at an alpha level of 0.05.

## 2.8 Evaluation of Diversity

The objective here was to determine if improvements in the AUC of augmented data were due to the larger sample size or due to the generative models increasing the diversity of the datasets (which is the mechanism described in the literature).

### 2.8.1 Measuring Diversity

Diversity is an important evaluation metric to assess the quality of generated synthetic data and is sometimes defined as the proportion of real data covered by the synthetic data [61,62]. However, in our study, we are more interested in identifying synthetic data records that are significantly different from the original samples. In other words, a new data record is defined to be diverse if it is different (i.e., the extent to which it is an outlier or an anomaly) from the original sample. It is necessary to find an effective approach to detect the anomaly records in one dataset with reference to another one.

Since diversity is measured at the dataset level rather than an individual record level, one way to conceptualize diversity is to compare the multivariate variation in the original data and the augmented data. If augmentation results in greater variation, then that would be an indicator of greater data diversity, several versions of multivariate coefficients of variation were introduced to measure the variability of populations using the characteristics of the numeric variables [63–66]. Another study proposed a method to determine the variability specifically for categorical data [67]. However, these methods are restricted to one type of variable and our datasets have both categorical and numeric variables. An alternative approach is to examine methods for assessing data shift. Kamulete developed a data-driven approach, called D-SOS, to detect non-negligible adverse shifts in a sample using outlier scores [68]. In contrast to other statistical tests, D-SOS focuses on identifying distributions that are not benign but significantly shifting from the reference sample by placing more weights on instances in the outlying regions of the sample data. However, the contamination rate that aims to detect non-negligible adverse shifts is distribution-based and therefore, unsuitable to our context, which is to capture the amount of new and diverse observations.

Inspired by this idea, we designed a new metric to measure the diversity using outlier data records in the augmented dataset compared to the base dataset. A record in the augmented dataset is deemed to be an outlier using a score obtained from an extended isolation forest model trained on the base dataset. The extended isolation forest, an extension of isolation forest, addresses the bias problem during the tree branching that arises in the standard isolation forest and therefore, is more robust in detecting anomalies [69,70]. Then, the trained isolation forest model is applied to both the base and augmented datasets to predict the outlier score for each observation, where a larger predicted score indicates a higher possibility of an outlier record. An incremental sequence of thresholds $\tau_j$ is created from 0.01 to 1 with a step size of 0.01. Then, we calculate a threshold-dependent contamination rate that is quantified as the proportion of outliers in the data, which are the records with outlier scores equal to or exceeding $\tau_j$ at step j. For a given threshold, a higher contamination rate implies a greater percentage of outlier records, and consequently, the data are more diverse. The are difference between the two contamination curves of the augmented and base datasets is the additional amount of diverse data records contributed to the original data. We are only interested in the positive difference as the negative difference means the contamination rate of the augmented data does not provide any meaningful increment in the diversity. Thus, the diversity metric is defined as follows.

$$diversity = \frac{\sum_{j=1}^{100}\{1(x_j \geq 0) \cdot \left(x_j(2-x_j)\right) + 1(x_j < 0) \cdot 0\}}{100}, \qquad (1)$$

where 1(·) is the indicator function, and $x_j$ represents the difference between the contamination rates of the augmented and base datasets at the threshold $\tau_j$. Thus, if the difference in the contamination rates is zero or positive, we calculate the diversity using a weighted contamination rate difference, which is always non-negative.

The steps of the calculation are included in the appendix.

### 2.8.2 Evaluating Impact of Diversity

In addition to the four generative models, we included the bootstrap method as another approach to augment the base dataset by resampling the original records. The purpose of including the bootstrap method is to rule out the influence of increasing data size. The size of the additional data that were bootstrapped was the same as the amount of synthetic data generated from the generative model that led to the optimal performance.

We also performed the diversity analysis for the seven datasets to show the quantity of the diverse records in the augmented data that were generated using either the generative models or bootstrap. For each dataset, the diversity was averaged across five iterations as the final diversity values for both the best generative model and bootstrap.

One-tailed exact permutation tests of the mean paired difference were performed to compare the diversity of the datasets and of the AUC results with resampling and generation. An alpha level of 0.05 was used.

## 3. Results
### 3.1 Overall Augmentation Performance

In this section, the performance of data augmentation against the size of synthetic data $n'$ in 40 different $n_0$ scenarios is summarized. To make the trends more interpretable and visible, the scales for the y-axis are varied, and the logarithm is taken for n'. Local regression was used to fit a smooth curve for each generative model.

In the main body of the paper, we present results for the BSA and FAERS datasets. The results for the remaining datasets are included in the appendix. These two datasets were selected for inclusion in the main body since the former is simple data and the latter is quite a complex dataset (with multiple variables with high cardinality). They illustrate the findings across the range of data complexity. The conclusions drawn from these two datasets are consistent with those from the other datasets.

In Figure 2 and Figure 3, it can be clearly seen that the augmentation can improve the performance measured by AUC, as more synthetic data are incorporated, especially for small and medium $n_0$. In fact, the improvements in model performance as measured by the AUC can be nontrivial, in some cases exceeding absolute increases of 0.1. For the large $n_0$, the improvement from augmentation is less or there is even a deterioration. In addition, the performance of SDG models varies significantly across different $n_0$ and base datasets, demonstrating the importance of identifying the most appropriate model in a specific situation. Moreover, compared to the BSA dataset, the FAERS dataset benefits more from data augmentation, as the highest $n_0$ with noticeable improvement is relatively larger, around $n_0$ = 3000, whereas the highest $n_0$ with noticeable improvement for the BSA dataset is approximately 650. Since the FAERS dataset is more complex with higher cardinality variables, further augmentation may generate more plausible values from the population, which leads to a more diverse augmented dataset compared to the BSA dataset.

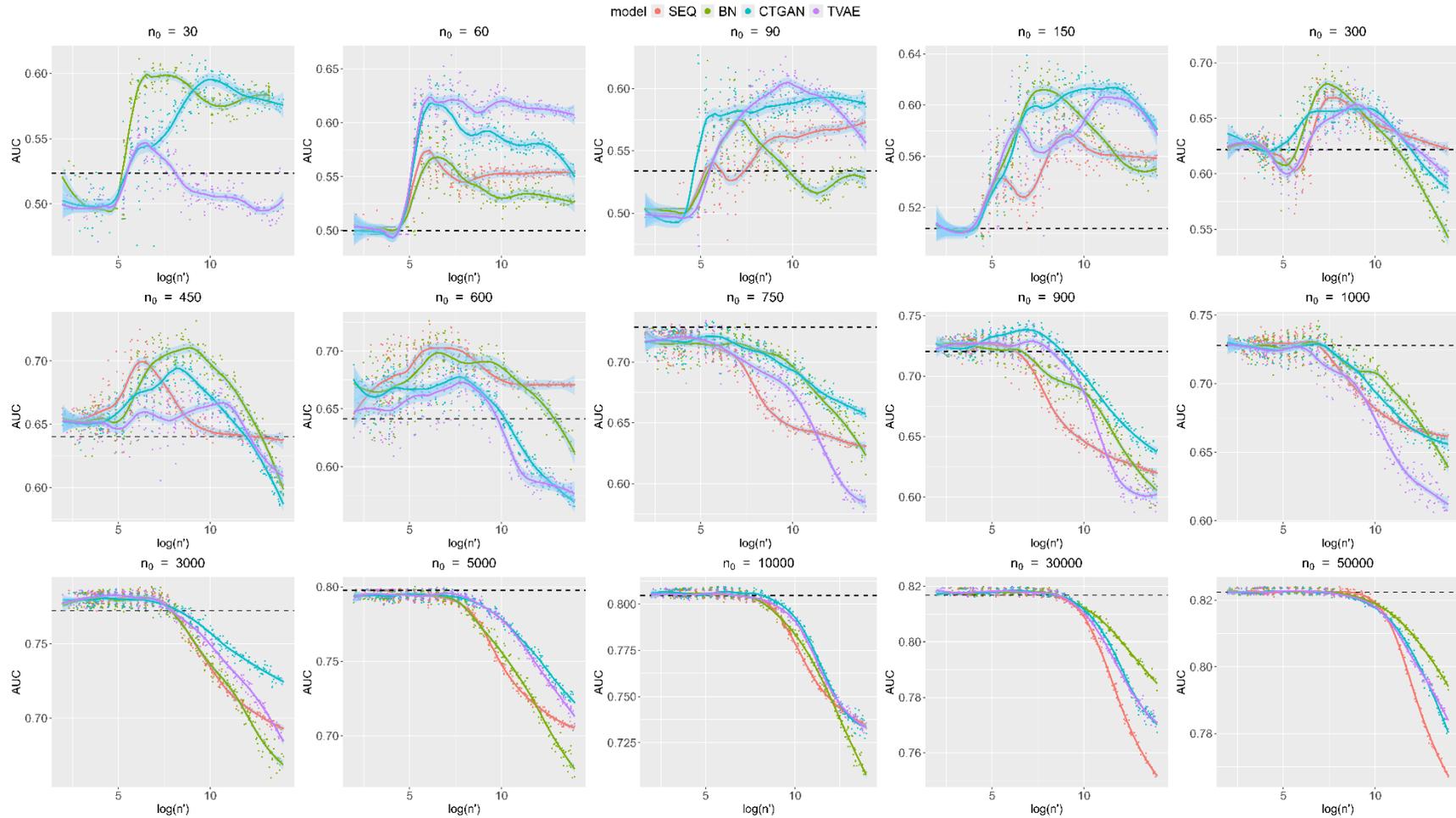

**Figure 3:** Augmentation performance of AUC against log($n'$) for the BSA dataset for a subset of the baseline data sizes. The black dotted line is the baseline AUC for the base dataset of size $n_0$.

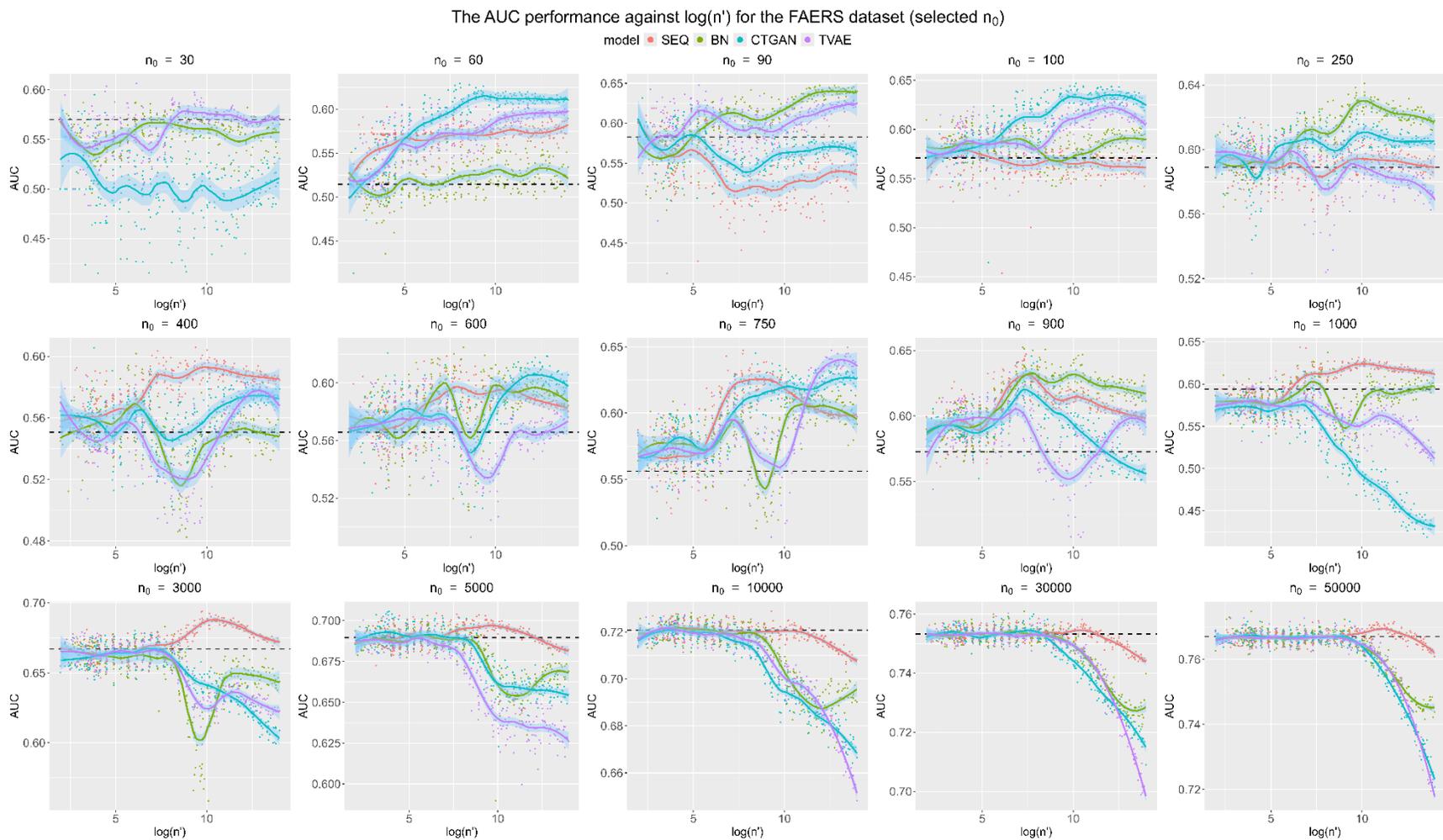

**Figure 4:** Augmentation performance of AUC against log($n'$) for the FAERS dataset for a subset of the baseline data sizes. The black dotted line is the baseline AUC for the base dataset of size $n_0$.

## 3.2 Augmentation Decision Support Tool

Table 2 presents the LOOCV results of AUC and prediction accuracy for the five methods on the decision support model. The best model is logistic regression, which is highlighted in bold. It results in the highest AUC of 0.7661 and a prediction accuracy of 76.15%, indicating its superior performance over other ML models in terms of the ability to recommend augmentation (or not). One possible reason for LR performing best might be that the underlying relationships between the data characteristics and augmentation indicator tend to be linear, and logistic regression is well suited to capture such linear relationships. Random forest and LGBM perform similarly in ranking after logistic regression, while the rest of the ML models are less favorable in predicting the augmentation choice for new datasets.

| Model | AUC | Prediction accuracy (%) |
| --- | --- | --- |
| **LR** | **0.7661** | **76.15** |
| RF | 0.7397 | 72.88 |
| LGBM | 0.7330 | 72.12 |
| XGB | 0.7009 | 71.92 |
| SVM | 0.7006 | 71.54 |

**Table 2:** The AUC and prediction accuracy results for the five models that predict whether augmentation should be performed using LOOCV. LR: logistic regression. RF: random forest. SVM: support vector machine. LGBM: light gradient boosting machine. XGB: extreme gradient boosting.

As a result, logistic regression is selected as the final model with the parameters in Table 3 to implement the decision support model that should be applied to a new dataset. The outcome is a binary indicator of whether there would be a benefit from augmentation using any of the four generative models. Table 5 presents both the unstandardized and standardized coefficients in the LR model.

| Variable | Unstandardized coefficients | Standardized coefficients |
|---|---|---|
| intercept | 6.75 | |
| $n_0$ | $-4.79 \times 10^{-5}$ | -0.52 |
| imbalance factor | $-4.94 \times 10^{-2}$ | -0.18 |
| degrees of Freedom | $5.12 \times 10^{-4}$ | 0.62 |
| baseline AUC | -7.63 | -0.97 |

**Table 3:** The logistic regression model parameters for predicting whether augmentation would be beneficial. The standardized coefficients are the unstandardized ones multiplied by the standard deviation of the respective variables.

The LR model shows that the baseline AUC has the biggest impact on whether to recommend augmentation, with lower baseline AUC baseline datasets benefiting more from augmentation. Moreover, the datasets that are smaller, more balanced, and more complex with higher cardinality are more likely to benefit from augmentation.

### 3.3   Evaluation of Augmentation and Diversity

After calculating the data characteristics for the analysis datasets, our decision model suggested that augmentation should be performed for all the case study datasets. The four generative models were employed to simulate the additional datasets. Table 4 presents the augmentation results for each dataset, the generative model that leads to the optimal performance, the amount of synthetic data records needing to be generated to achieve the optimal performance and the performance using the bootstrap method.

The baseline AUC values are within the range from poor to good. The additional synthetic data sizes vary depending on both the generative model that was used and the dataset. As expected, the best generative model is not uniform.

The relative improvement in AUC due to generative model augmentation is remarkably high ranging from 4.3% to 43.23% (average 15.55%), indicating a substantial gain in the model performance after augmentation (baseline AUC vs augmented AUC: p=0.0078). The resampling augmentation generally yields a much lower AUC, compared to the synthetic data generative models and on some occasions is even worse than the baseline scenario without augmentation (augmented AUC vs resampled AUC: p=0.016). Increasing the sample size by resampling the original data often does not contribute to the improvement of model performance as much as the other synthetic data generative models.

|  |  |  | AUC Results |  |  |  | Diversity Results |  |
|---|---|---|---|---|---|---|---|---|
| Dataset | Model | $n'_{max}$ | Baseline AUC | Augmented AUC | Relative AUC (%) | Resampled AUC | Diversity generative | Diversity resample |
| Hot Flashes | CTGAN | 720 | 0.7161 | 0.7668 | 7.08 | 0.6477 | 0.0023 | 0.0013 |
| Danish Colorectal Cancer Group | TVAE | 720 | 0.7171 | 0.778 | 8.50 | 0.7077 | 0.0000 | 0.0008 |
| Breast Cancer Coimbra | BN | 53 | 0.7392 | 0.8722 | 18.00 | 0.8291 | 0.0061 | 0.0019 |
| Breast Cancer | CTGAN | 25 | 0.7143 | 0.7451 | 4.31 | 0.6729 | 0.0017 | 0.0008 |
| Colposcopy/Schiller | CTGAN | 2,205 | 0.5125 | 0.7341 | 43.23 | 0.6116 | 0.0883 | 0.0004 |
| Diabetic Retinopathy | BN | 11,534 | 0.74 | 0.7974 | 7.75 | 0.7299 | 0.1177 | 0.0002 |
| Thoracic Surgery | TVAE | 6,602 | 0.5584 | 0.67 | 19.98 | 0.6914 | 0.0000 | 0.0003 |

**Table 4:** Analysis results of augmentation performance for the seven datasets. The augmented dataset size is $n_o + n'_{max}$. for example, it would be 1008 observations for the Hot Flashes dataset. $n'_{max}$: n' that leads to maximum AUC. Baseline AUC: baseline AUC from the base data. Augmented AUC: maximum AUC from the augmented data. Resampled AUC: AUC from the augmented data with a size of $n'_{max}$ using resampling with replacement method. Diversity generative: diversity of data augmented using a generative model. Diversity resample: diversity of data augmented using the bootstrap method.

The diversity results for the resampled data are generally lower than those for the data augmented using the generative models (generative diversity vs. resampled diversity: p=0.046). Therefore, augmentation using the generative models does increase the diversity of the datasets beyond just a simple increase in the sample size from the original data distribution.

## 4. Discussion

### 4.1 Summary

The availability of health data for research purposes is limited, and these datasets are often small. However, training of ML models requires large amounts of data to obtain optimal performance on unseen data, and training on datasets that are too small can lead to model instability [85], and to overfitting and an inability to generalize predictions to unseen data [3,86] even under ideal conditions (e.g., no data shift or drift). Consequently, the conclusions drawn from such models may be unstable and inaccurate. In such cases, data augmentation can be beneficial by simulating more, and more diverse, data based on the existing data.

Although it has been receiving increasing attention in recent years, especially in imaging data and time series data applications, tabular data augmentation has not been extensively evaluated despite data augmentation being one of the primary use cases for synthetic data generation methods [87]. In this study, we fill this gap by evaluating the benefits of data augmentation for tabular health data.

Our simulations show that augmenting existing data can enhance the ML performance as measured by AUC, especially for smaller, more balanced, more complex datasets or datasets with lower baseline AUC. Excessive augmentation is not necessarily beneficial. The appropriate level of augmentation that maximizes performance differs for each dataset. However, the benefits of augmentation are less obvious or even detrimental for large datasets.

Our interpretation of this phenomenon is that with small or moderate data size to start, the simulated data positively contributes by increasing the size and diversity, and thus, more likely adds the information that is similar to the unseen dataset. In contrast, for a large base dataset, the increase in size has less marginal prognostic benefit and the dataset may already contain sufficiently diverse information, and incorporating more simulated data is less likely to provide useful diversity. In fact, it may be increasing the unnecessary noise in the current dataset. Moreover, the simpler datasets with fewer categorical variables and lower cardinality were found to benefit less from augmentation, and this is arguably because the space to increase diversity is limited (i.e., simulated records will look more like current records rather than be different). That lower baseline AUC benefits more from augmentation can be attributed to a ceiling effect where higher AUC values will likely benefit less from augmentation.

Different generative models perform best depending on the dataset itself and its baseline size. Therefore, it is not possible to a priori say that a particular generative model is consistently superior for the augmentation task. Multiple SDG models need to be evaluated to find the best one to augment a particular dataset.

Consequently, we developed a predictive decision support model to help an analyst decide whether a new dataset should be augmented. This is of particular relevance given the computational load of evaluating multiple generative models. In practice, this model only requires basic characteristics of the base dataset and will make a recommendation on whether augmentation would be beneficial or not.

Our application of augmentation to seven small datasets further confirms the model performance improvement through augmenting the original dataset. These datasets resulted in model performance improvement ranging from 4.31% to 43.23% using the generative models (average 15.55%), whereas the datasets with only resampling did not consistently perform better than that. Diversifying the existing data through augmentation plays an important role in enhancing the model performance, and therefore, increasing the sample size without making the data more diverse is not as beneficial.

A recent smaller-scale study of data augmentation on tabular data similarly did not find a predominant generative model [88], which is why our recommendation of evaluating multiple models on each dataset and selecting the best performing one gives more reliable augmentation outcomes. Furthermore, the previous study did not examine the relationship between base sample size and degree of augmentation and did not consider data complexity and baseline model performance, hence the conclusions of that study were quite limited in this regard.

Previous work on the augmentation of longitudinal EHR data using generative models demonstrated improvements in prognostic accuracy on a handful of datasets [89]. However, our results on tabular data show that augmentation depends on the data characteristics, the specific generative model used, and the degree of augmentation, and therefore will not always be beneficial. Therefore, having a decision support model to determine a priori if augmentation can be helpful is important. In addition, our findings suggest that dynamic selection among multiple generative models to identify the best one given the specific data parameters provides better results.

## 4.2 Recommendations for Practice and Research

For datasets where the baseline AUC is high, augmentation may not provide a significant advantage. However, where the baseline AUC is medium or small, and where datasets are in the 100 to 3000 observations range, augmentation can potentially improve the performance of a model's AUC, sometimes by a considerable amount. Datasets with high cardinality categorical variables can also benefit from augmentation. In contrast, augmentation will likely be less beneficial for large and simple datasets with strong relationships with the outcome (i.e., higher baseline AUC).

Analysts can try different degrees of augmentation using multiple generative models and evaluate them on holdout data to determine which amount of augmentation can maximize the prognostic performance. We recommend using the data characteristic measures and the logistic regression decision support model to determine the necessity of augmentation for a given dataset.

It should be noted that the training dataset for the generative models should be separate from the dataset that is used for testing. This is easier to do in a simple train/test split scenario. However, if augmentation is used in the context of, say, 5-fold cross-validation then the generative models should be trained on the 4/5 training splits each time and evaluated on the remaining 1/5 split. This will ensure that there is no data leakage which would result in optimistic results that would not carry to unseen data in subsequent applications. For the final augmented dataset, the determined $n'\_max$ simulated records should be concatenated to the original dataset.

## 4.3 Limitations and Future Work

Evaluating the performance of each dataset at different levels of augmentation can be computationally intensive. This means that the processing time to determine the best level of augmentation may not be small in practice.

Our analysis assumed that resampling with replacement was a good proxy for increasing the sample size without increasing diversity. The reasoning was that adding observations from the same distribution would have a minimal impact on diversity.

When datasets are very small, the types of generative models that we used in our study have a higher risk of overfitting. However, the data dimensionality that was used has also tended to be low which is a mitigating factor. Nevertheless, future work should examine generative models that are suited for small datasets, such as those based on pre-trained models.

Synthetic data generation has been shown to introduce bias in the generated data relative to the training data [90], and these biases are propagated across multiple generations of generative models (where the output of one is used as training for the next one) [91]. Our study did not examine the impact of augmentation on fairness. The impact of augmentation on fairness is an open question that should be the subject of further studies.

Data amplification, which is when more synthetic data is generated relative to the base dataset that was used by the generative model, has been shown not to improve the quality of population inferences nor the replicability of results for statistical models [92]. Amplification is different from augmentation in that amplified data does not include any of the original data within it. Our results did not consider population inferences or replicability. However, it would be informative for future work to examine whether augmentation gives different conclusions with respect to population inferences.

Given our results showing augmented datasets with greater diversity have a higher improvement in predictive performance, further work can optimize generative models to specifically increase the diversity of the synthetic data to maximize the performance improvement for downstream ML workloads.

It is also of interest to examine the augmentation performance using other ML models in addition to LGBM (or boosted decision trees in general), such as random forests and support vector machines.

## Ethics

This project was approved by the Research Ethics Board of the Children's Hospital of Eastern Ontario Research Institute protocol 24/80x. The hot flashes data analysis was approved by the Ottawa Health Sciences Research Ethics Board protocols OHSN REB #20210727-01H and OHSN REB #20210827-01H. For the DCCG dataset, Danish Data Protection Agency (Datatilsynet) approval was obtained (RN-2018-94).

## Clinical Trial Registration

Clinical trial number: not applicable.

## Funding

This research is funded by the Canada Research Chairs program through the Canadian Institutes of Health Research, a Discovery Grant RGPIN-2022-04811 from the Natural Sciences and Engineering Research Council of Canada, and the Canadian Children Inflammatory Bowel Disease Network. LP is funded by the Deutsche Forschungsgemeinschaft (DFG, German Research Foundation) – 530282197.

## Competing Interests Statement

KEE is the Scholar-in-Residence at the Office of the Information and Privacy Commissioner of Ontario. KEE holds shares in Aetion, which provided the sequential synthesis generative model software that was used in this study.

## Author Contributions

DL, SK, NM. LP, and KEE designed the study and performed the analysis. TW, MC, AEH, and KEE provided and interpreted the datasets. GP consulted on the analysis. All authors contributed to writing the paper.

## Data Availability

The following provides information on the availability of each of the datasets used in this study:

1. Better Outcomes Registry & Network (BORN) |
   The BORN collects Ontario's prescribed perinatal, newborn and child registry with the role of facilitating quality care for families across the province. It can be accessed through a data request at https://bornontario.ca/en/data/data.aspx.
2. Basic Stand Alone (BSA)
   The BSA inpatient claims dataset is about claim-level information that each record is an inpatient claim incurred by a 5% sample of Medicare beneficiaries. The dataset is publicly available at https://www.cms.gov/data-research/statistics-trends-and-reports/basic-stand-alone-medicare-claims-public-use-files/bsa-inpatient-claims-puf.
3. California State Hospital Discharge
   The California dataset contains the patient's hospital 2008 discharge data from California, State Inpatient Databases (SID), Healthcare Cost and Utilization Project (HCUP), Agency for Healthcare Research and Quality [93], and is available for purchase at https://hcup-us.ahrq.gov/tech_assist/centdist.jsp.
4. Canadian Community Health Survey (CCHS)
   The CCHS data are Canadian population-level information concerning health status, health system utilization and health determinants collected by Statistics Canada through telephone survey. The availability of CCHS data is restricted and requires an access request at https://www150.statcan.gc.ca/n1/pub/82-620-m/2005001/4144189-eng.htm.
5. COVID-19
   The COVID-19 dataset collects Canadian health records of COVID-19 gathered by the Public Health Agency of Canada and is available at Esri Canada (https://resources-covid19canada.hub.arcgis.com/).
6. FDA Adverse Event Reporting System (FAERS)
   The FAERS is a database comprising the information on adverse events and medication error reports submitted to FDA and can be downloaded at https://open.fda.gov/data/faers/.
7. Florida State Hospital Discharge
   The Florida dataset contains the patient's hospital 2007 discharge data from Florida, State Inpatient Databases (SID), Healthcare Cost and Utilization Project (HCUP), Agency for Healthcare Research and Quality [93], and is available for purchase at https://hcup-us.ahrq.gov/tech_assist/centdist.jsp.
8. MIMIC-III
   MIMIC-III is a large database that contains deidentified health-related data associated with over forty thousand patients who stayed in critical care units of the Beth Israel Deaconess Medical Center between 2001 and 2012 [94,95]. The access to the MIMIC database is upon signing a data use agreement with PhysioNet at https://physionet.org/content/mimiciii/1.4/ [96].
9. New York State Hospital Discharge
   The New York dataset contains the patient's hospital 2007 discharge data from New York, State Inpatient Databases (SID), Healthcare Cost and Utilization Project (HCUP), Agency for Healthcare Research and Quality [93], and is available for purchase at https://hcup-us.ahrq.gov/tech_assist/centdist.jsp.

10. COVID-19 Survival (Nexoid)
    The COVID-19 survival dataset is a web-based survey data collected by a company called Nexoid in United Kingdom. It is publicly available at https://www.covid19survivalcalculator.com/en/download.
11. Texas Hospital Discharge
    The Texas dataset contains the patient's hospital discharge information for the first quarter of 2012 from Texas in the United States [97], and is publicly available at https://www.dshs.texas.gov/center-health-statistics/chs-data-sets-reports/texas-health-care-information-collection/health-data-researcher-information/texas-inpatient-public-use.
12. Washington State Hospital Discharge 2007
    The Washington dataset contains the patient's hospital 2007 discharge data from Washington, State Inpatient Databases (SID), Healthcare Cost and Utilization Project (HCUP), Agency for Healthcare Research and Quality [93], and is available for purchase at https://hcup-us.ahrq.gov/tech_assist/centdist.jsp.
13. Washington State Hospital Discharge 2008
    The Washington2008 dataset contains the patient's hospital 2008 discharge data from Washington, State Inpatient Databases (SID), Healthcare Cost and Utilization Project (HCUP), Agency for Healthcare Research and Quality [93], and is available for purchase at https://hcup-us.ahrq.gov/tech_assist/centdist.jsp.
14. Hot Flashes
    The Hot Flashes dataset stores the health information of patients with early breast cancer who experienced vasomotor symptoms, and the access request is available by contacting the senior authors of the original article.
15. Danish Colorectal Cancer Group
    The Danish Colorectal Cancer Group (DCCG) dataset comprises all patients with colorectal cancer in Denmark between 2001 and 2018. The DCCG dataset can be requested from the Danish Colon Cancer registry.
16. Breast Cancer Coimbra
    The Breast Cancer Coimbra dataset contains women with breast cancer recruited by the Gynaecology Department of the University Hospital Centre of Coimbra between 2009 and 2013 and is publicly available at https://archive.ics.uci.edu/dataset/451/breast+cancer+coimbra.
17. Breast Cancer
    The Breast Cancer dataset collects information by the University Medical Centre, Institute of Oncology, Ljubljana, Yugoslavia, and is publicly available at https://archive.ics.uci.edu/dataset/14/breast+cancer.
18. Colposcopy/schiller
    The Colposcopy/schiller dataset is one of three modality datasets that collects subjective quality assessment of digital colposcopies and is publicly available at https://archive.ics.uci.edu/dataset/384/quality+assessment+of+digital+colposcopies.
19. Diabetic Retinopathy
    The Diabetic Retinopathy dataset extracts health information from the Messidor image set and is publicly available at https://archive.ics.uci.edu/dataset/329/diabetic+retinopathy+debrecen.
20. Thoracic Surgery
    The Thoracic Surgery dataset describes the post-operative life expectancy of patients who underwent lung resections for primary lung cancer between 2007 and 2011 and is publicly available at https://archive.ics.uci.edu/dataset/277/thoracic+surgery+data.

## Code Availability

The code used in this analysis can be accessed as follows:

- The synthetic data generation code is available in the pysdg package, available from: <https://osf.io/xj9pr/>
- The machine learning modeling was performed using the R sdgm package available from: <https://osf.io/DCJM6>
- The R code for applying the results on a new dataset is available from <https://osf.io/4gu62/>.

# Appendix

# Synthetic Data Generation for Augmenting Small Samples

## Contents

# Appendix A - Steps for Calculating Diversity

The detailed implementation procedures of our proposed diversity metric are as follows.

1. Build an extended insolation forest model using the base dataset.
2. Apply the extended isolation forest model to the base dataset and calculate the outlier scores for the base dataset.
3. Apply the extended isolation forest model to the augmented dataset and calculate the outlier scores for the augmented dataset.
4. Generate a sequence of thresholds $\tau_j$ starting from 0.01 and increase to 1 by a step size of 0.01 such that $\tau_j \in \{0.01, 0,02, …, 1\}$.
5. At each j, compute the contamination rate as the proportion of the outlier scores in the base dataset that is equal to or greater than $\tau_j$.
6. At each j, compute the contamination rate as the proportion of the outlier scores in the augmented dataset that is equal to or greater than $\tau_j$.
7. Find the difference between the two contamination rates.
8. Compute the diversity according to the equation in the body of the paper.

Our proposed metric enjoys several advantages. First, the metric is fairly straightforward to understand and implement. Secondly, it is data-driven and adaptive to any given dataset. Another strength is that it allows mixed data to be processed, offering the flexibility to quantify the diversity between the two datasets.

# Appendix B - Results for augmentation performance (AUC)

## B.1 BORN Dataset

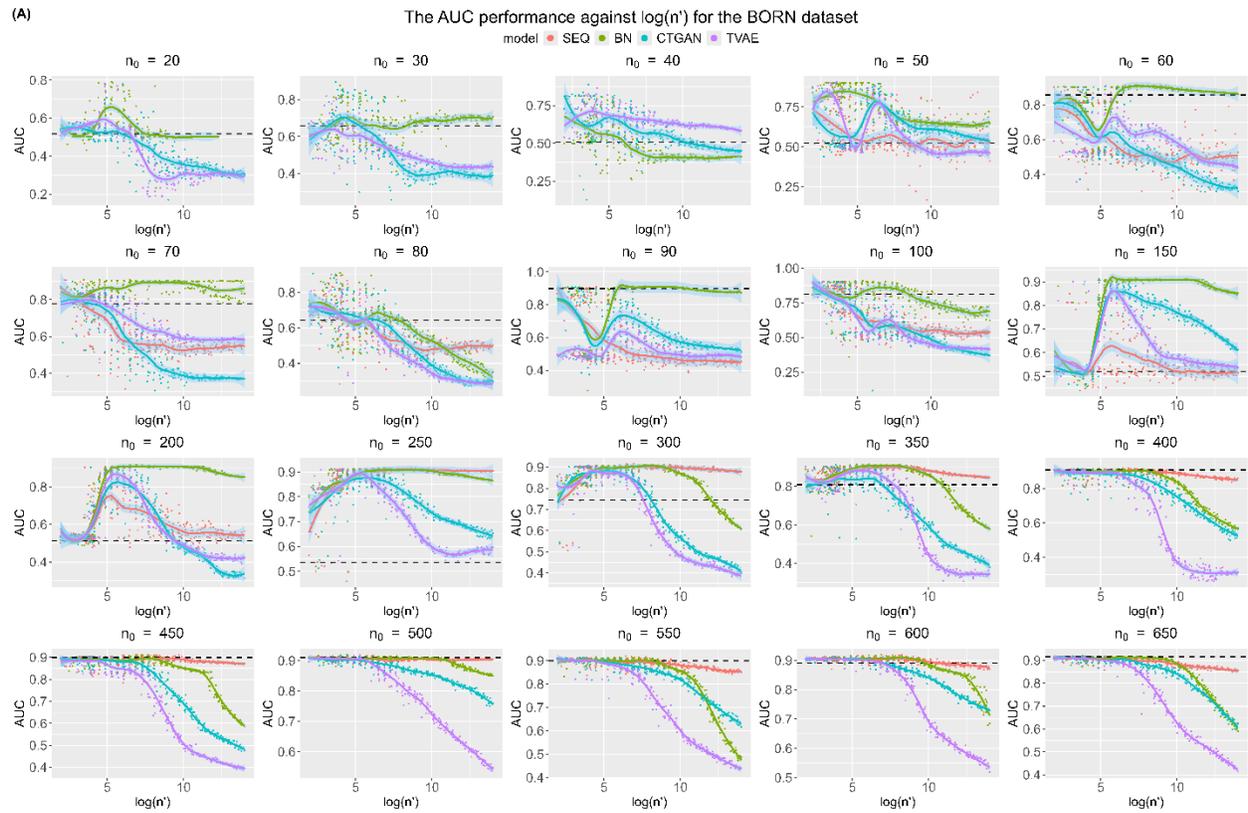

**Figure B.1:** Augmentation performance of AUC against log(*n'*) for the BORN dataset (A).

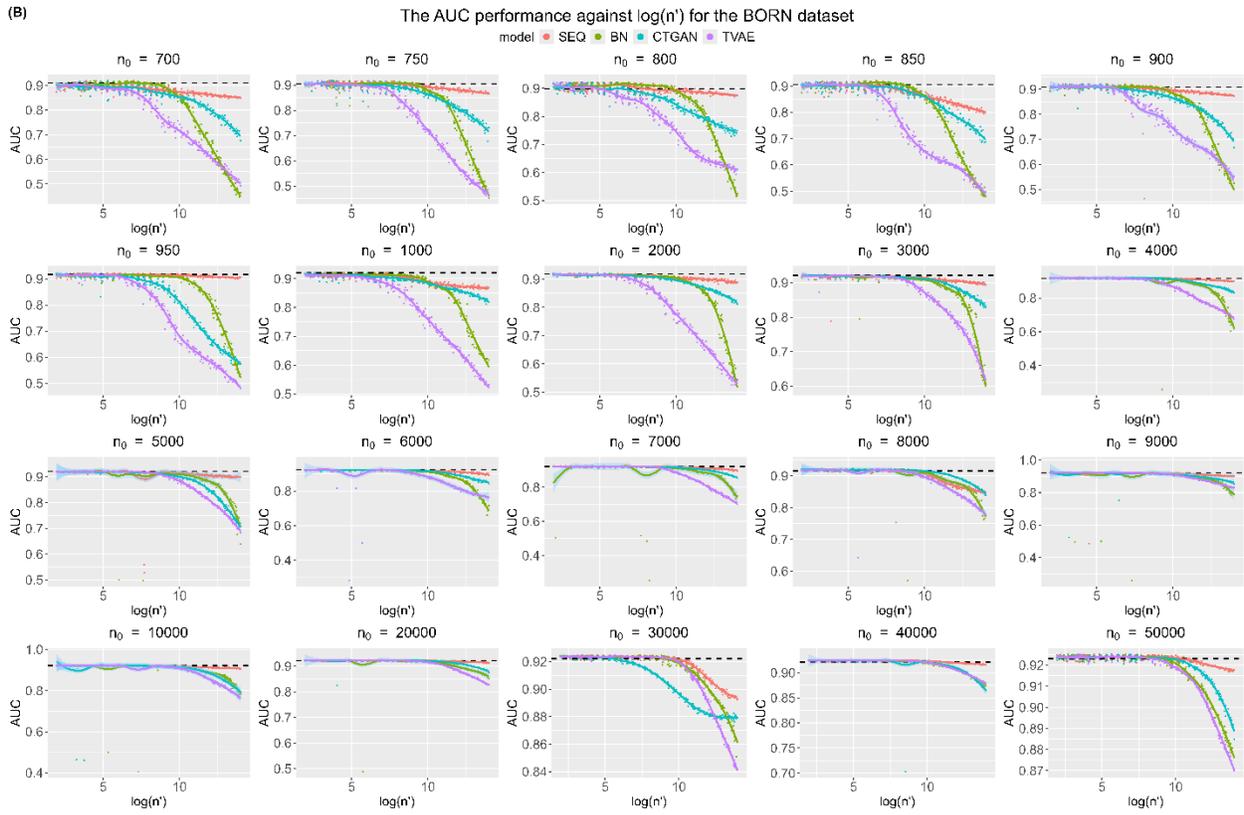

**Figure B.2:** Augmentation performance of AUC against log(*n'*) for the BORN dataset (B).

## B.2 BSA Dataset

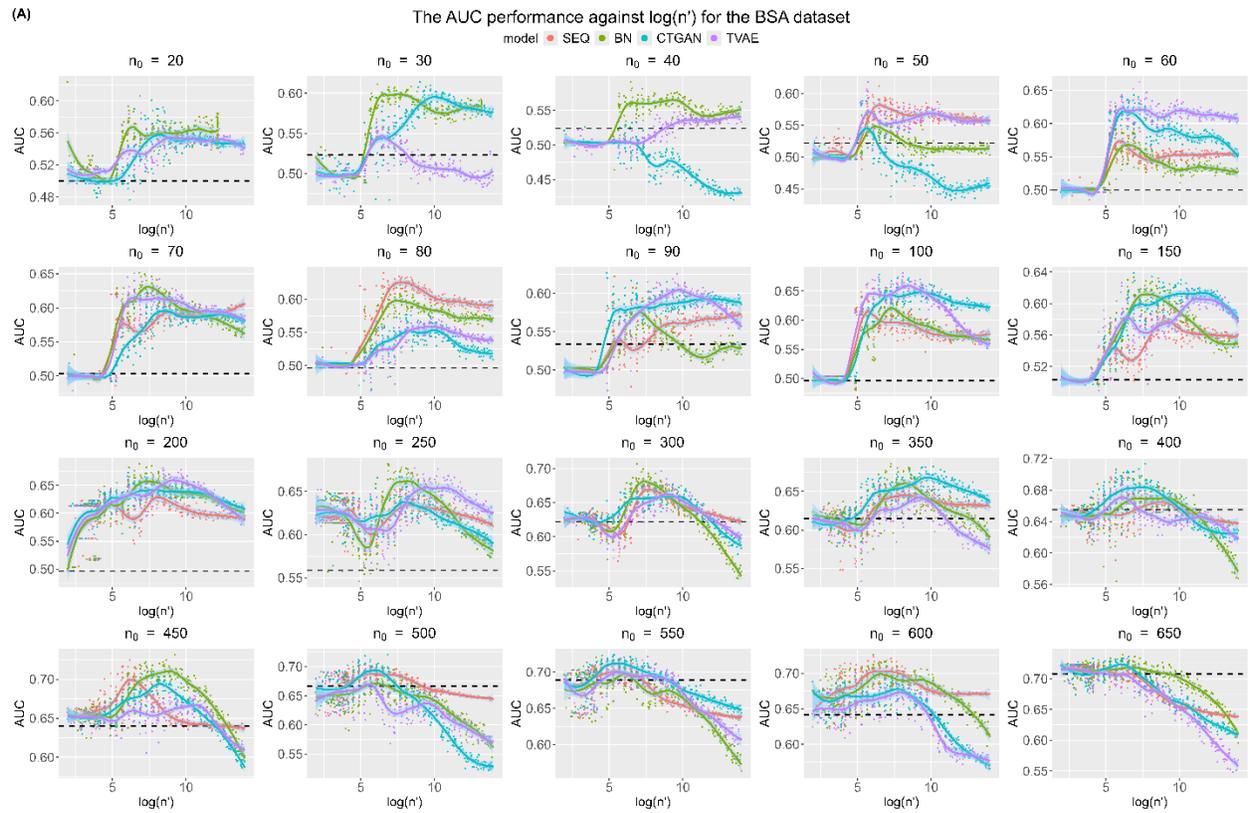

**Figure B.3:** Augmentation performance of AUC against log($n'$) for the BSA dataset (A).

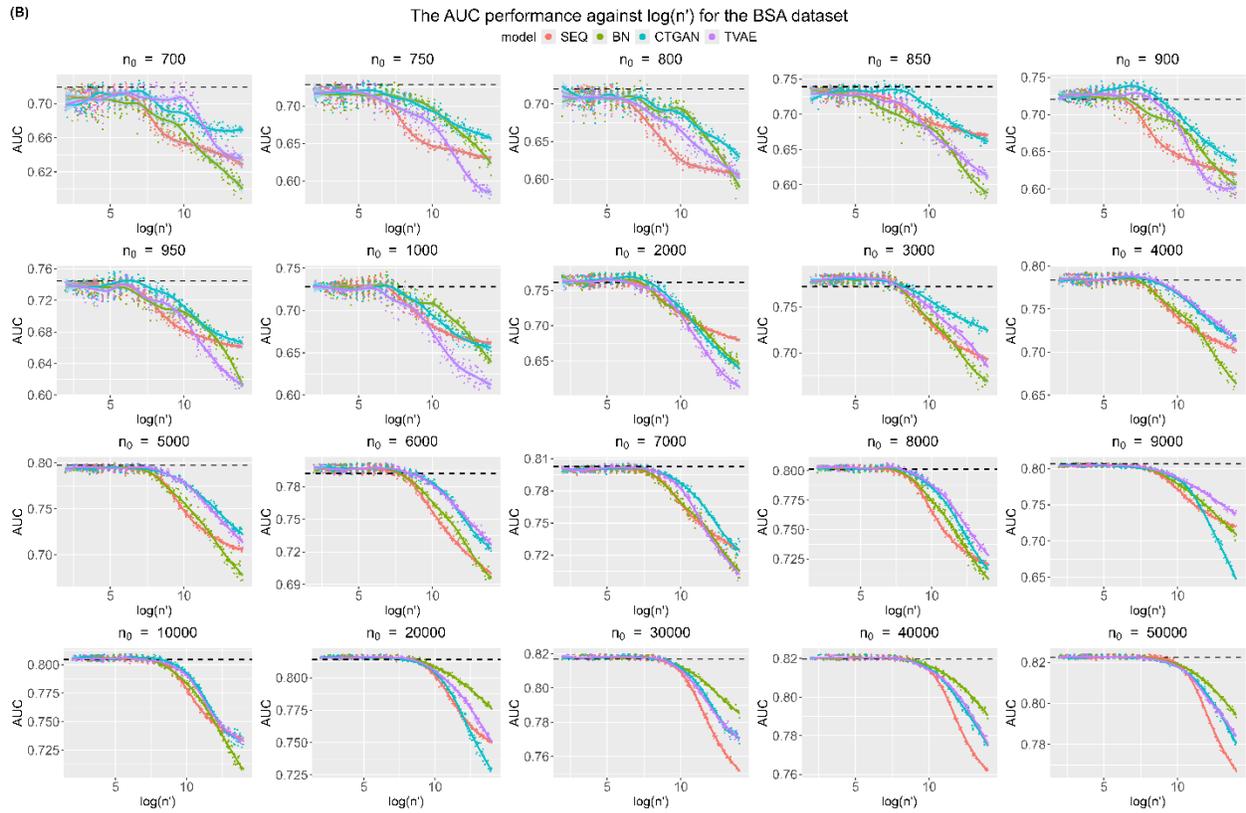

**Figure B.4:** Augmentation performance of AUC against log(*n'*) for the BSA dataset (B).

## B.3 California Dataset

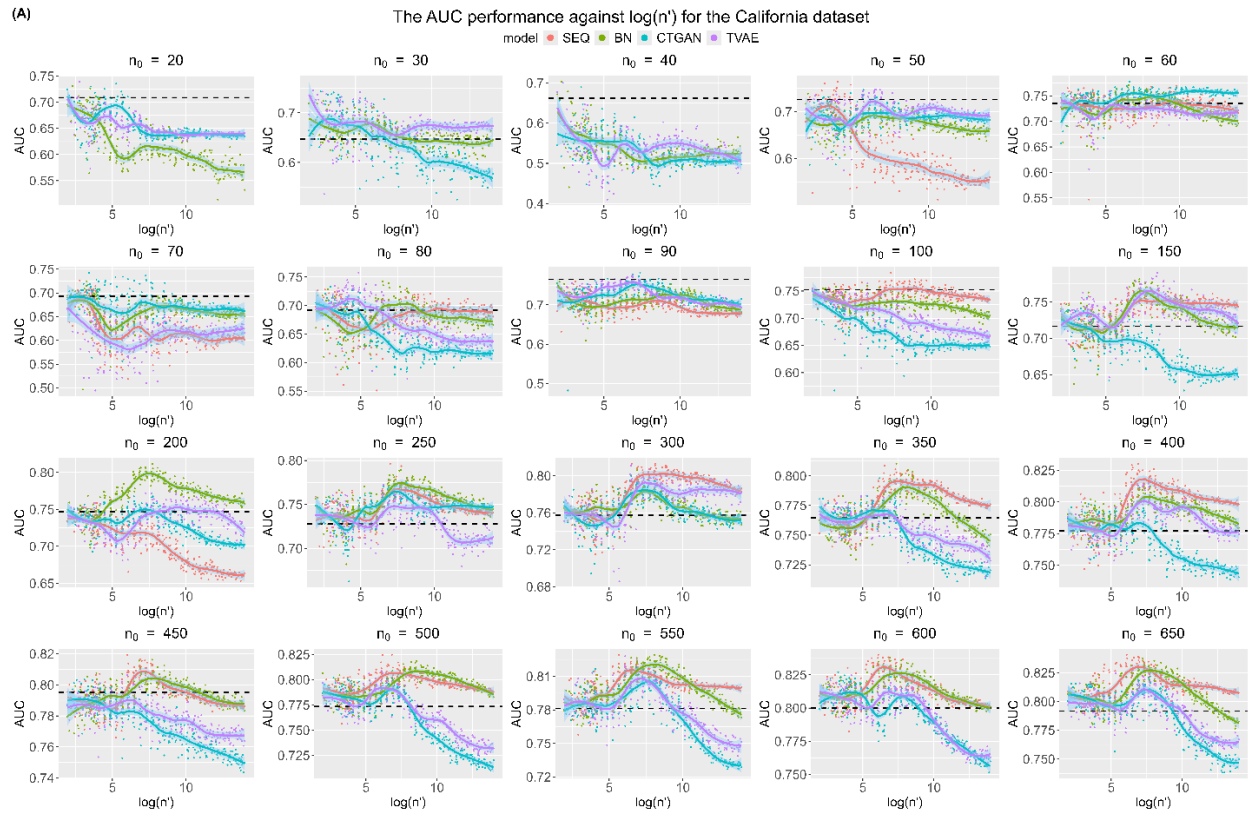

**Figure B.5:** Augmentation performance of AUC against log($n'$) for the California dataset (A).

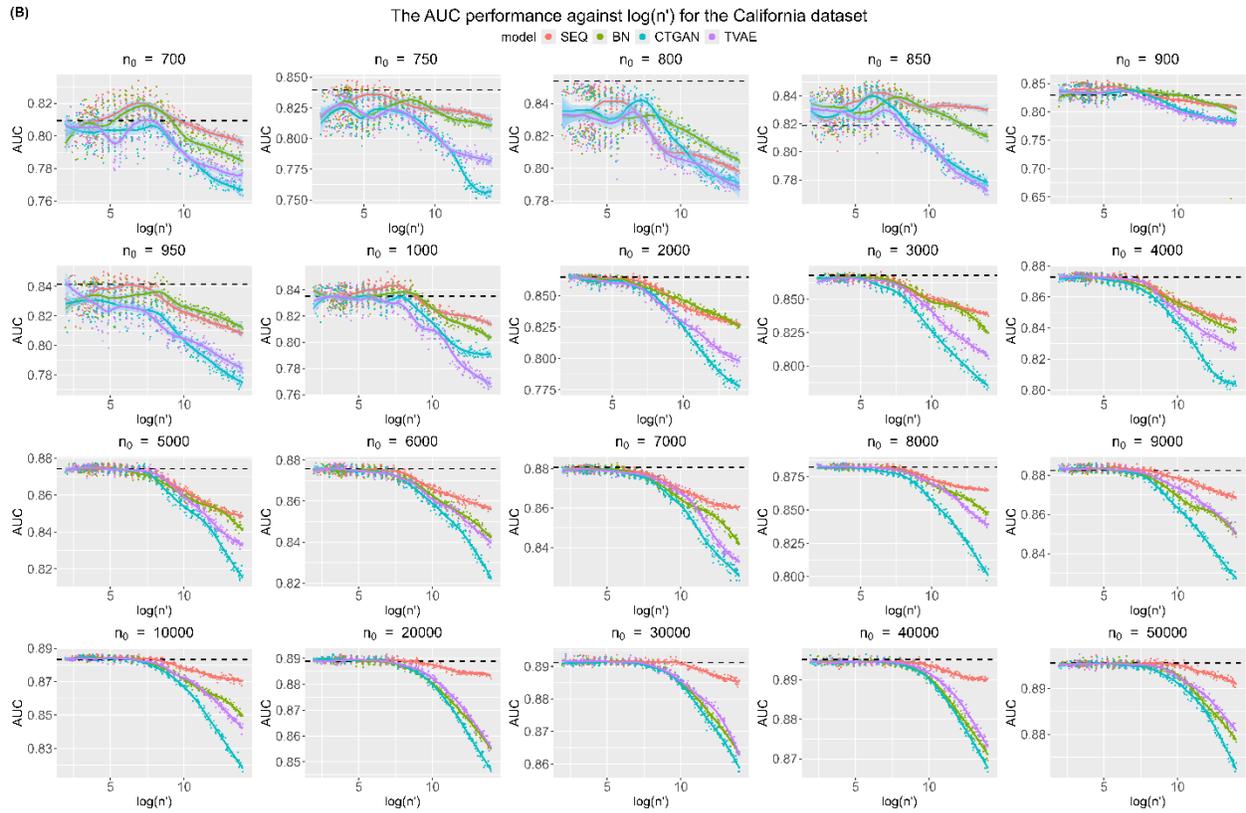

**Figure B.6:** Augmentation performance of AUC against log(*n'*) for the California dataset (B).

## B.4 CCHS Dataset

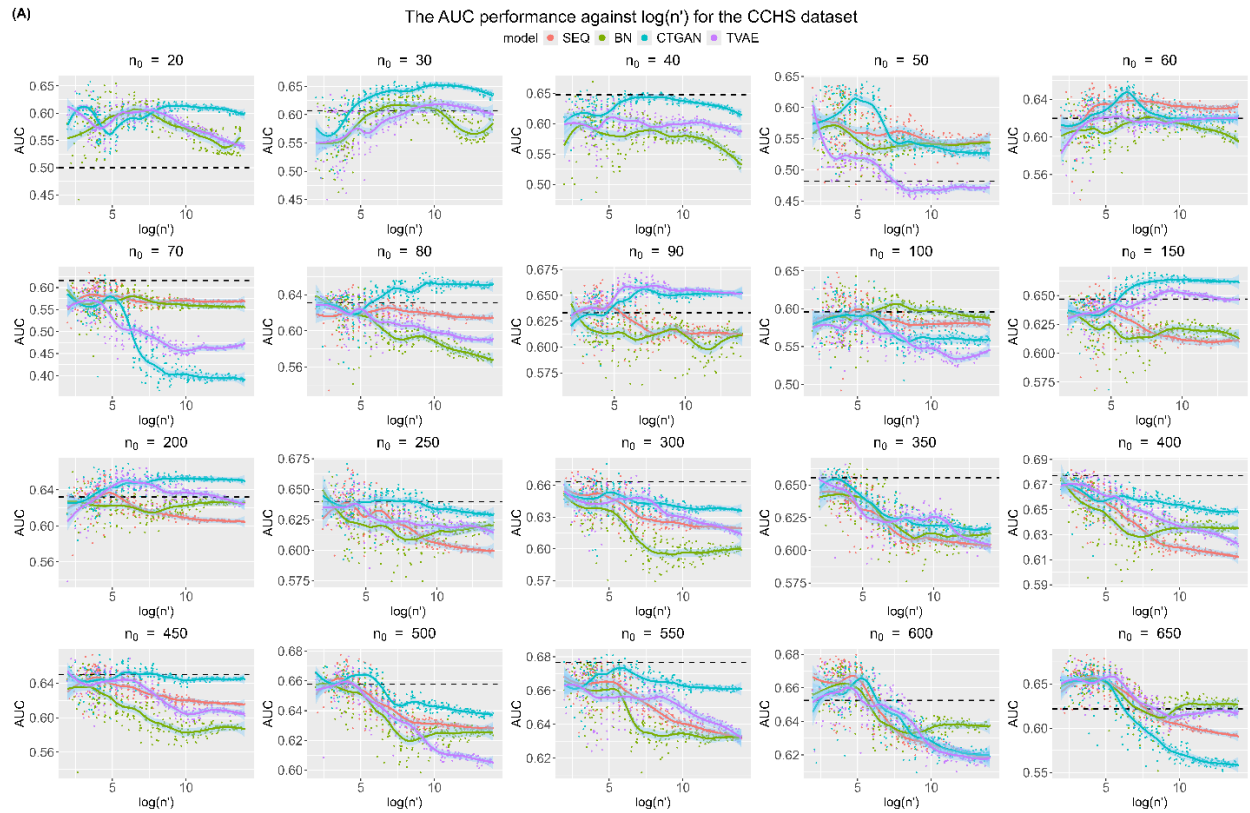

**Figure B.7:** Augmentation performance of AUC against log($n'$) for the CCHS dataset (A).

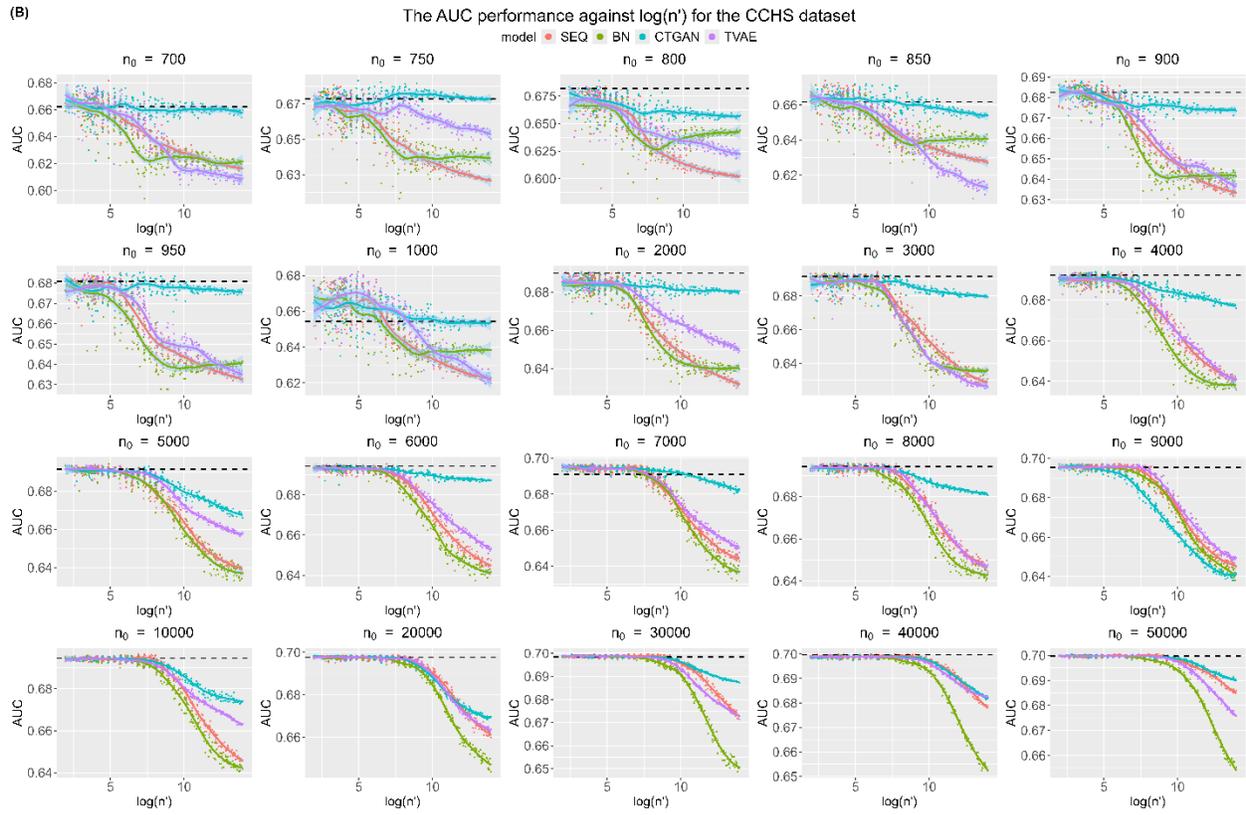

**Figure B.8:** Augmentation performance of AUC against log($n'$) for the CCHS dataset (B).

## B.5 COVID Dataset

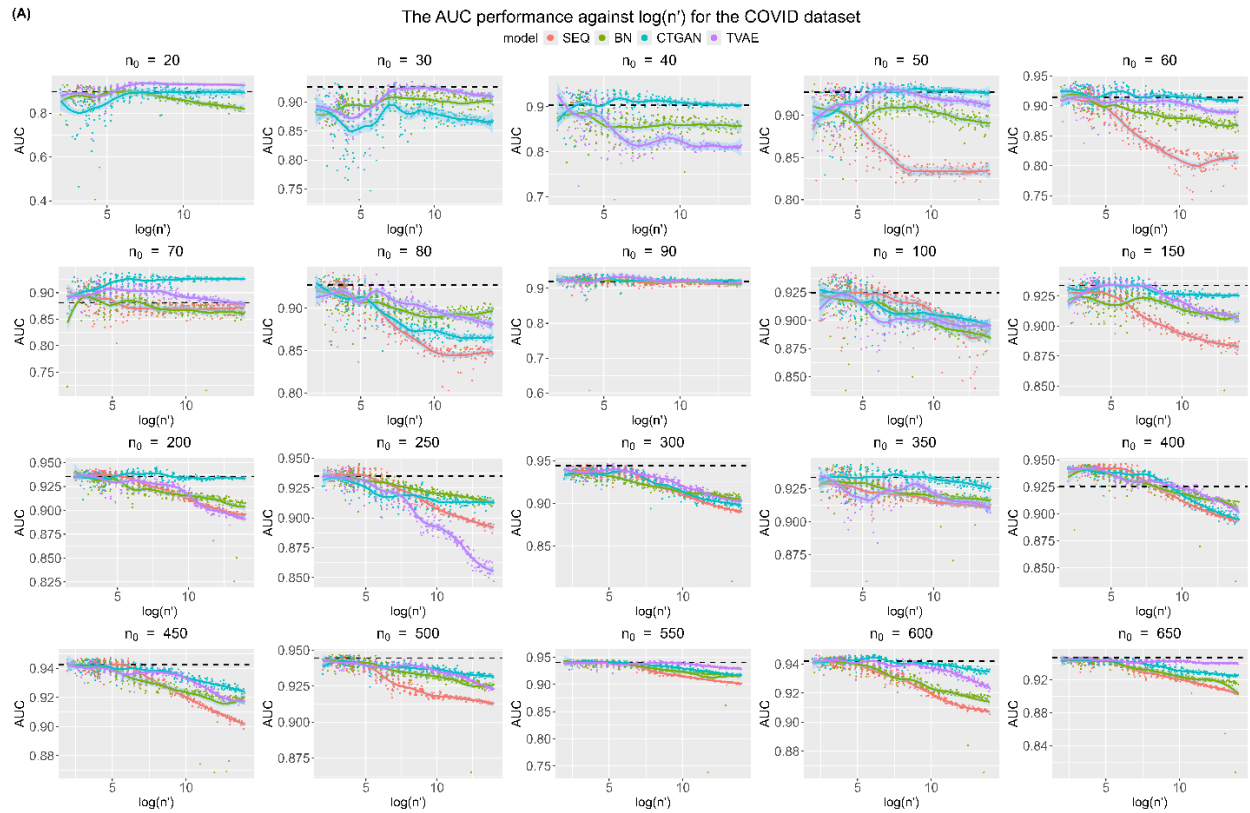

**Figure B.9:** Augmentation performance of AUC against log(*n'*) for the COVID dataset (A).

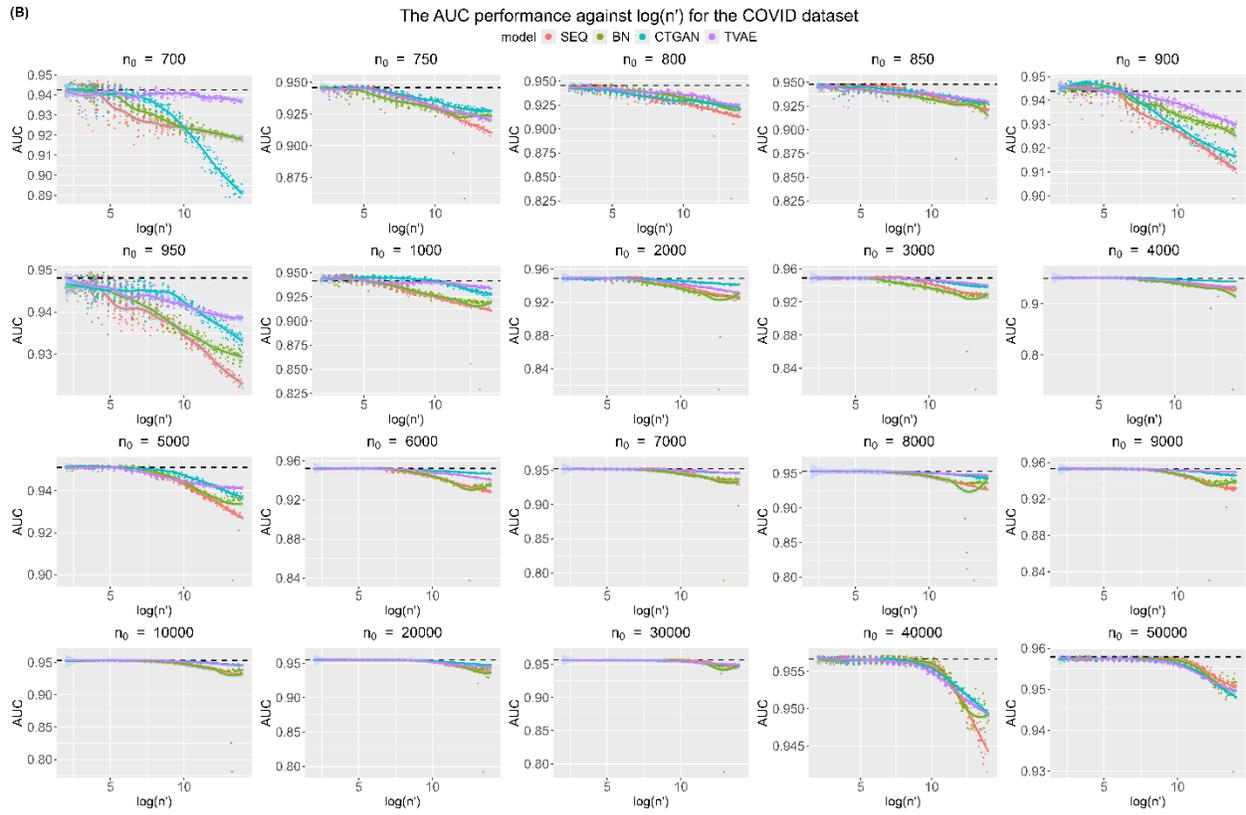

**Figure B.10:** Augmentation performance of AUC against log(*n'*) for the COVID dataset (B).

## B.6 FAERS Dataset

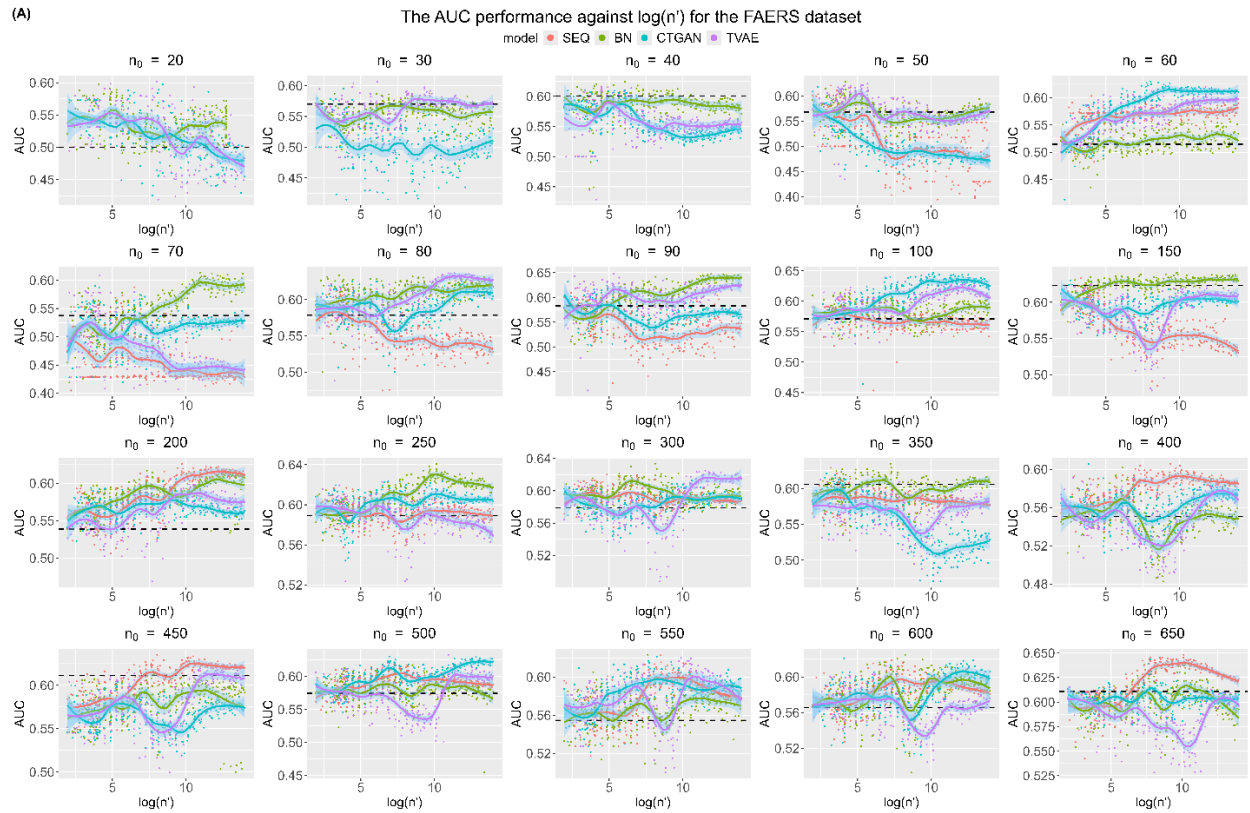

**Figure B.11:** Augmentation performance of AUC against log(*n'*) for the FAERS dataset (A).

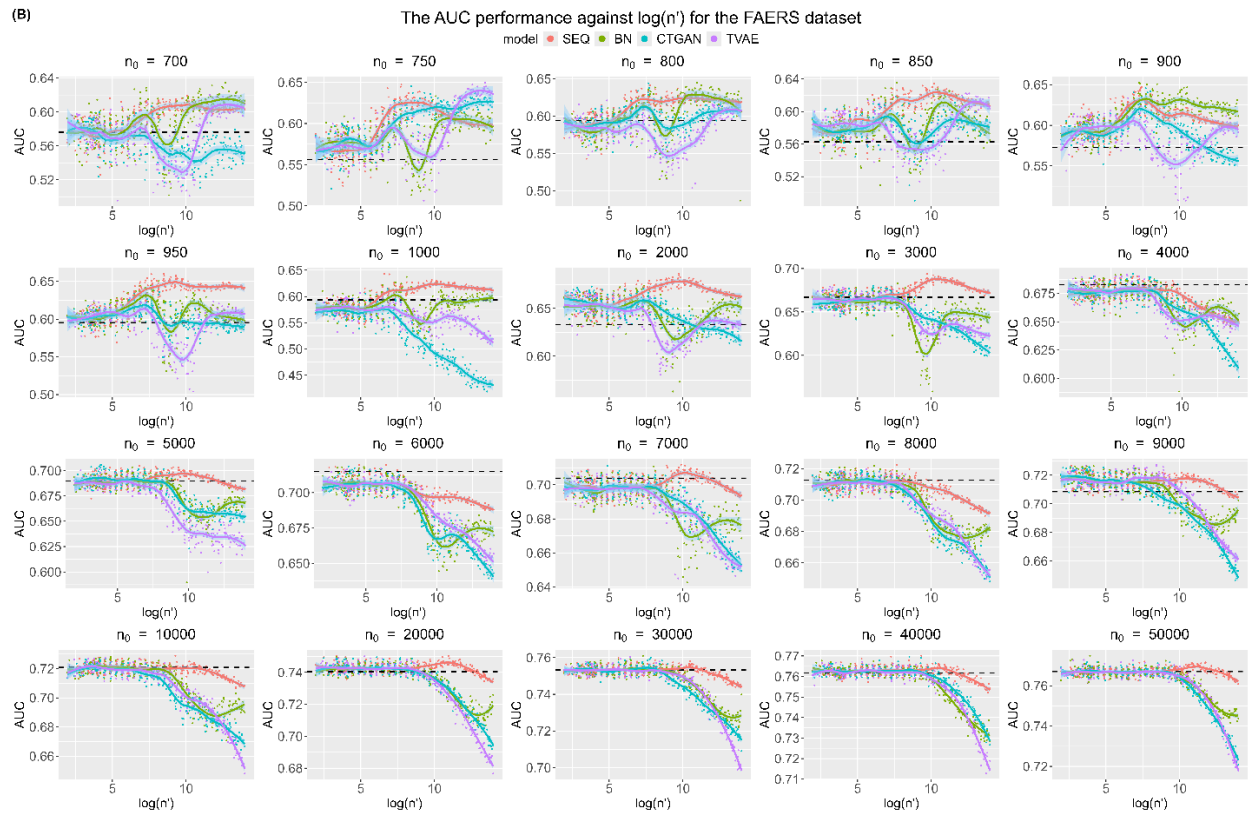

**Figure B.12:** Augmentation performance of AUC against log(*n'*) for the FAERS dataset (B).

## B.7 Florida Dataset

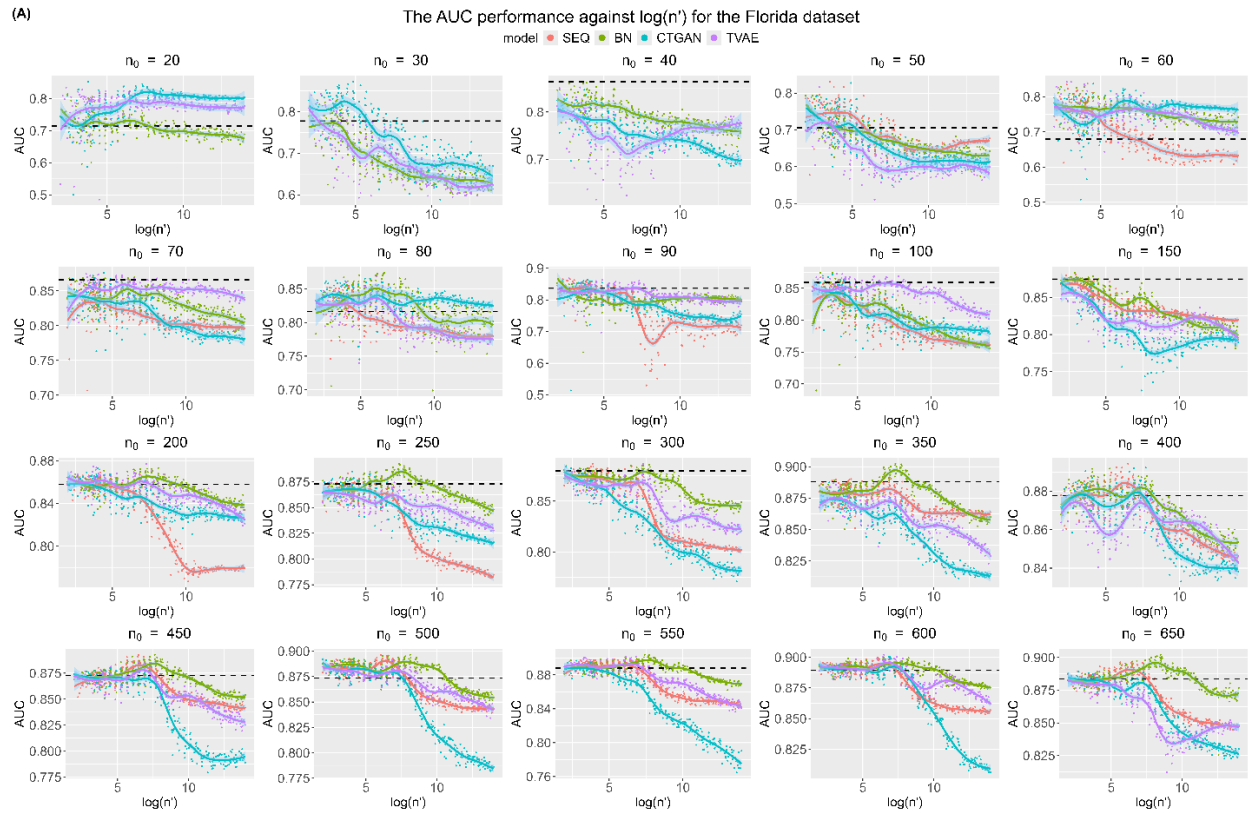

**Figure B.13:** Augmentation performance of AUC against log($n'$) for the Florida dataset (A).

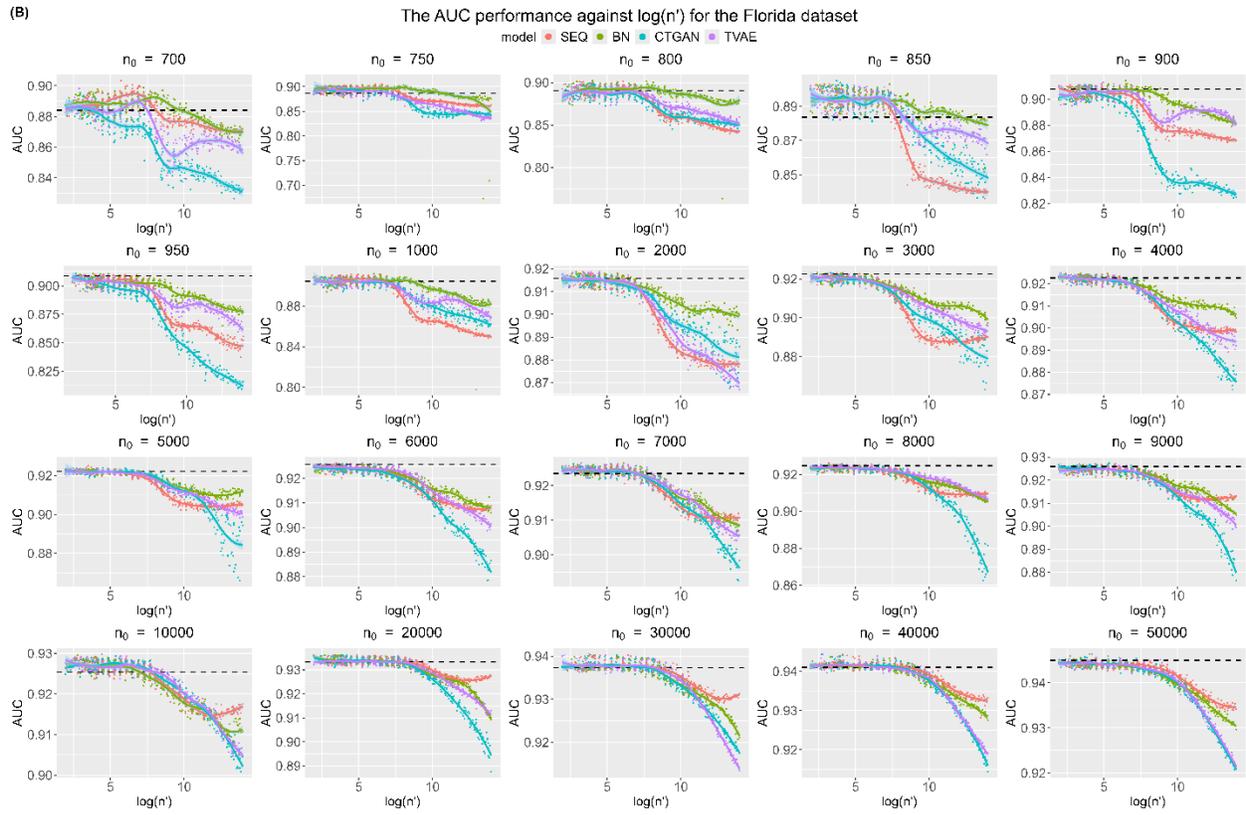

**Figure B.14:** Augmentation performance of AUC against log(*n'*) for the Florida dataset (B).

## B.8 MIMIC Dataset

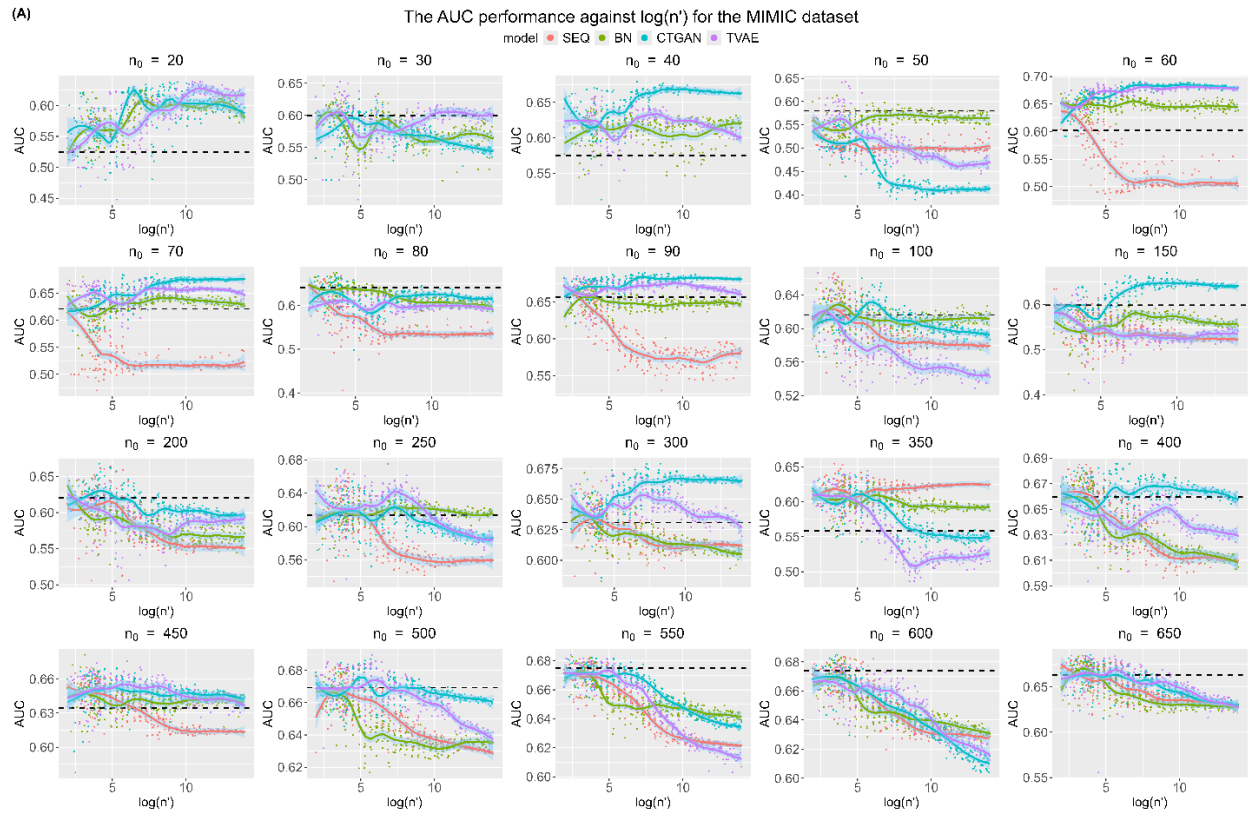

**Figure B.15:** Augmentation performance of AUC against log(*n'*) for the MIMIC dataset (A).

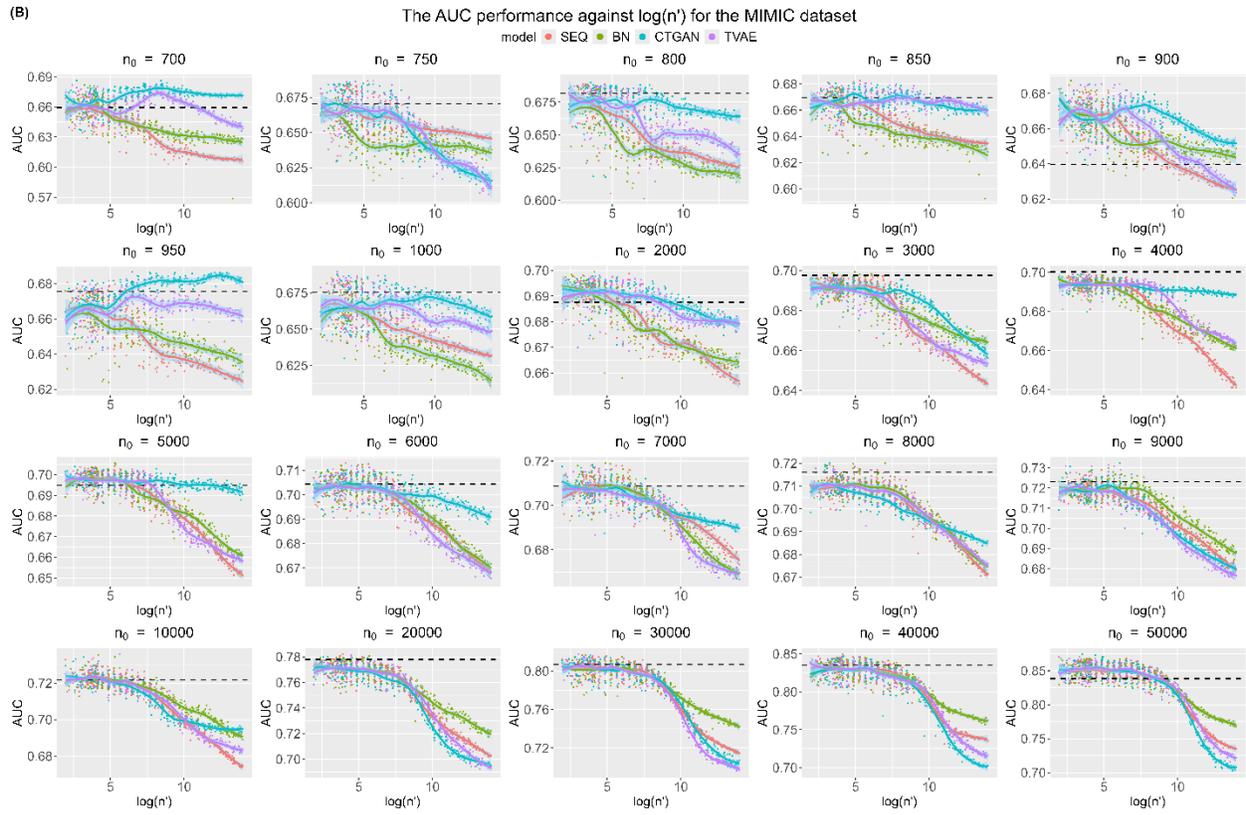

**Figure B.16:** Augmentation performance of AUC against log($n'$) for the MIMIC dataset (B).

## B.9 New York Dataset

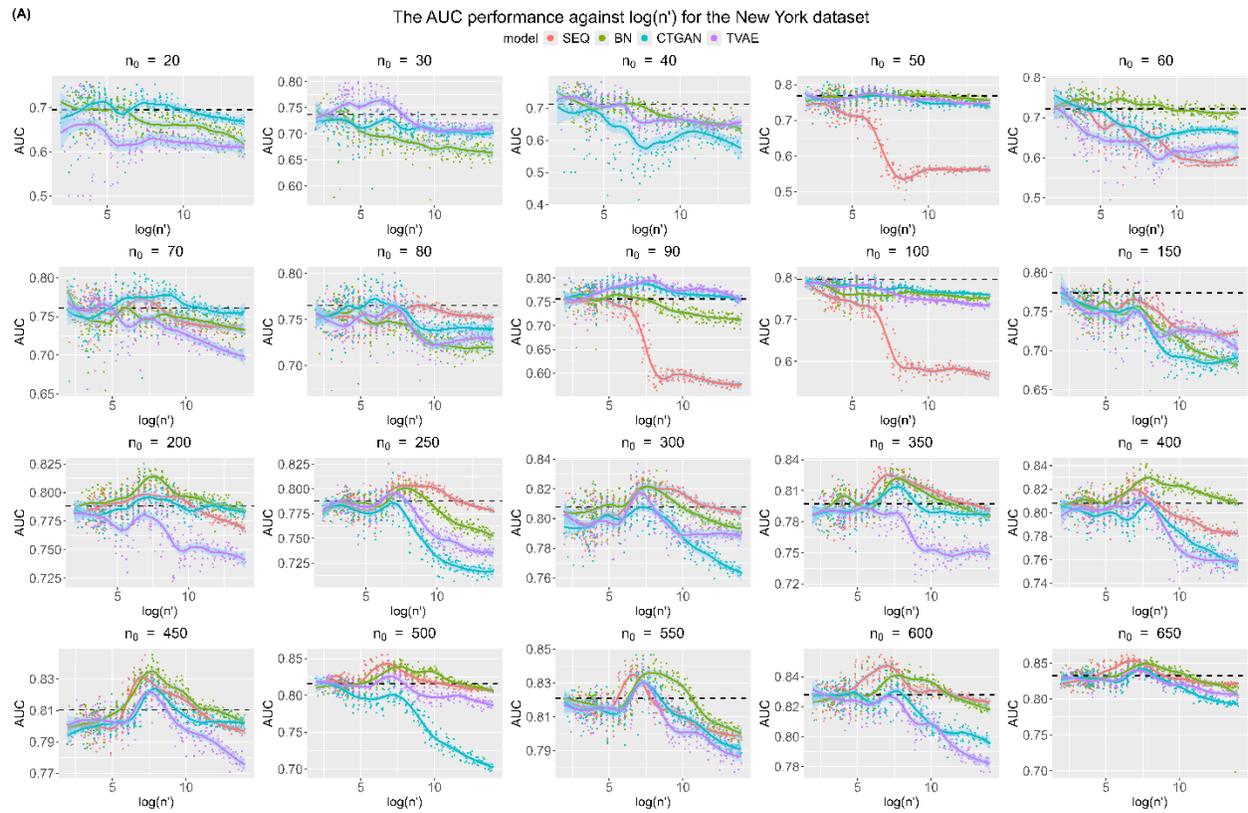

**Figure B.17:** Augmentation performance of AUC against log(*n'*) for the New York dataset (A).

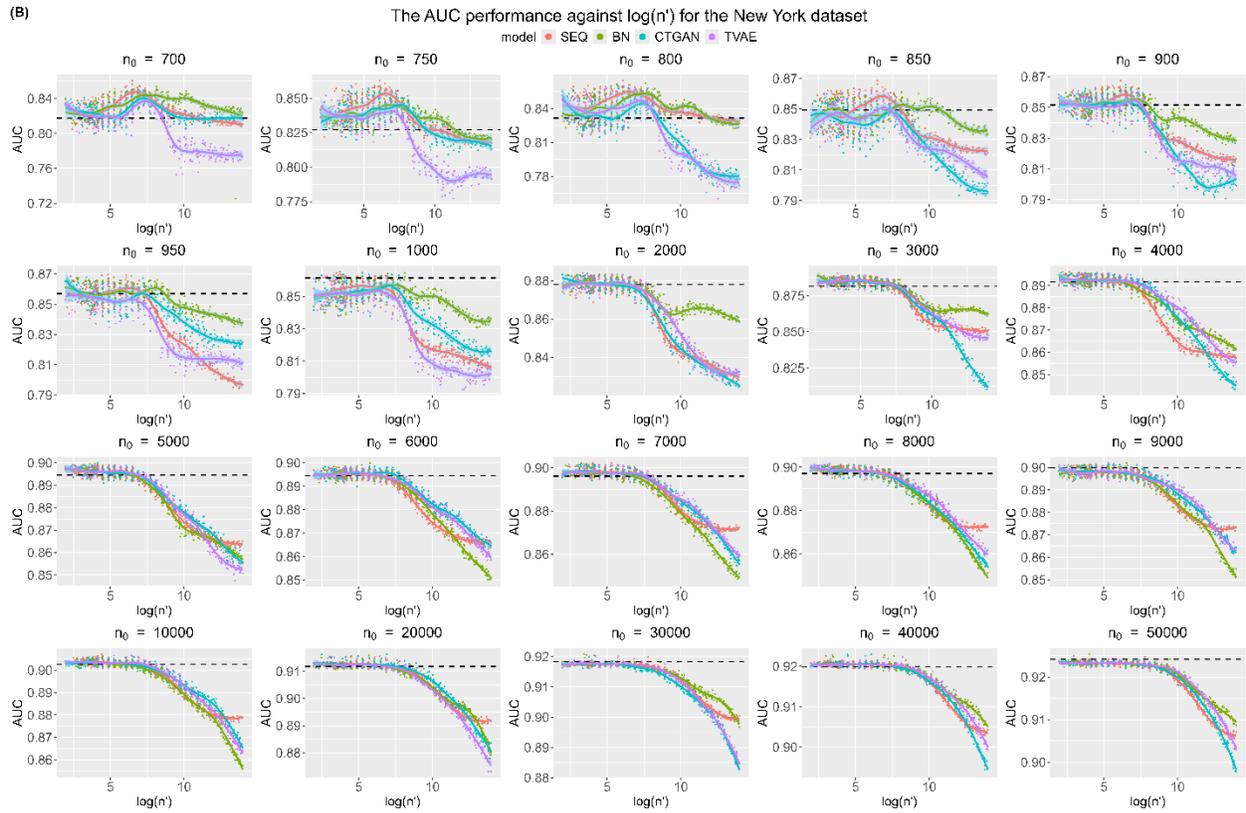

**Figure B.18:** Augmentation performance of AUC against log(*n'*) for the New York dataset (B).

## B.10 Nexoid Dataset

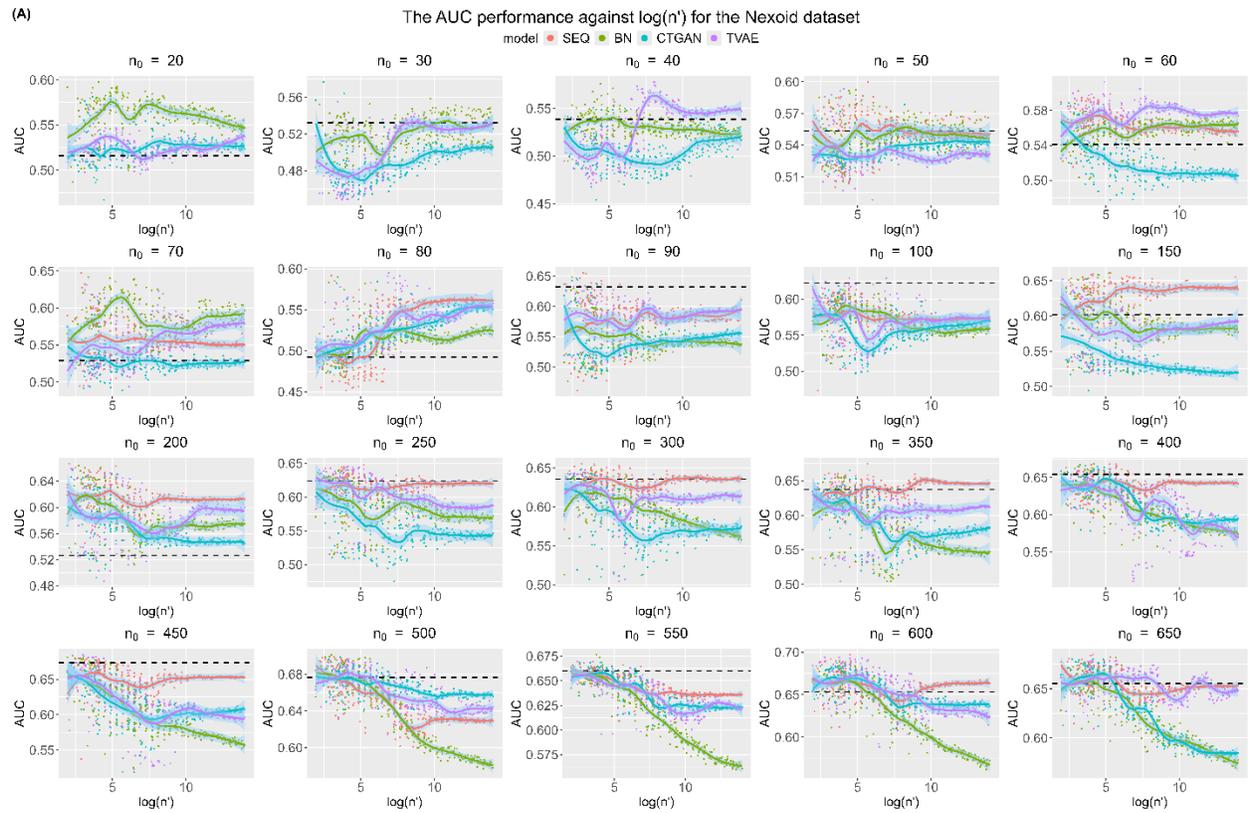

**Figure B.19:** Augmentation performance of AUC against log($n'$) for the Nexoid dataset (A).

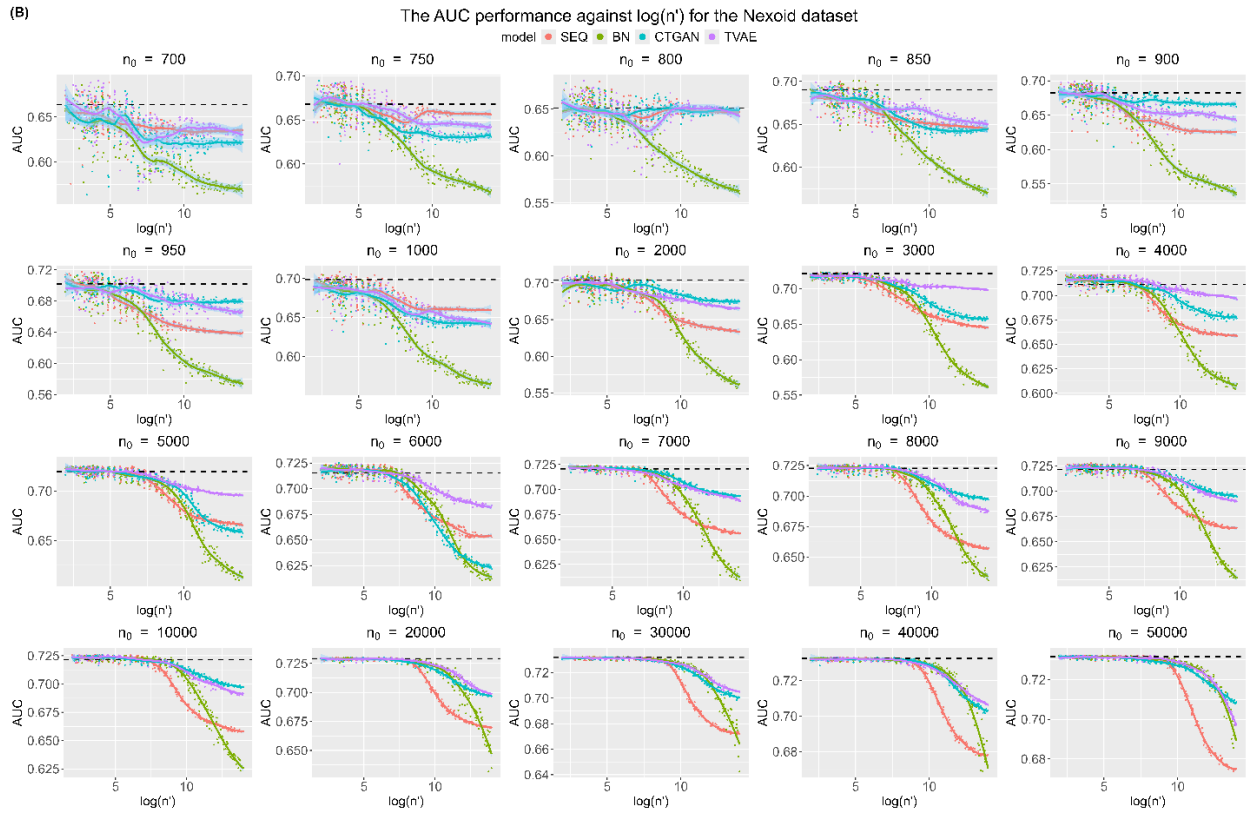

**Figure B.20:** Augmentation performance of AUC against log(*n'*) for the Nexoid dataset (B).

## B.11 Texas Dataset

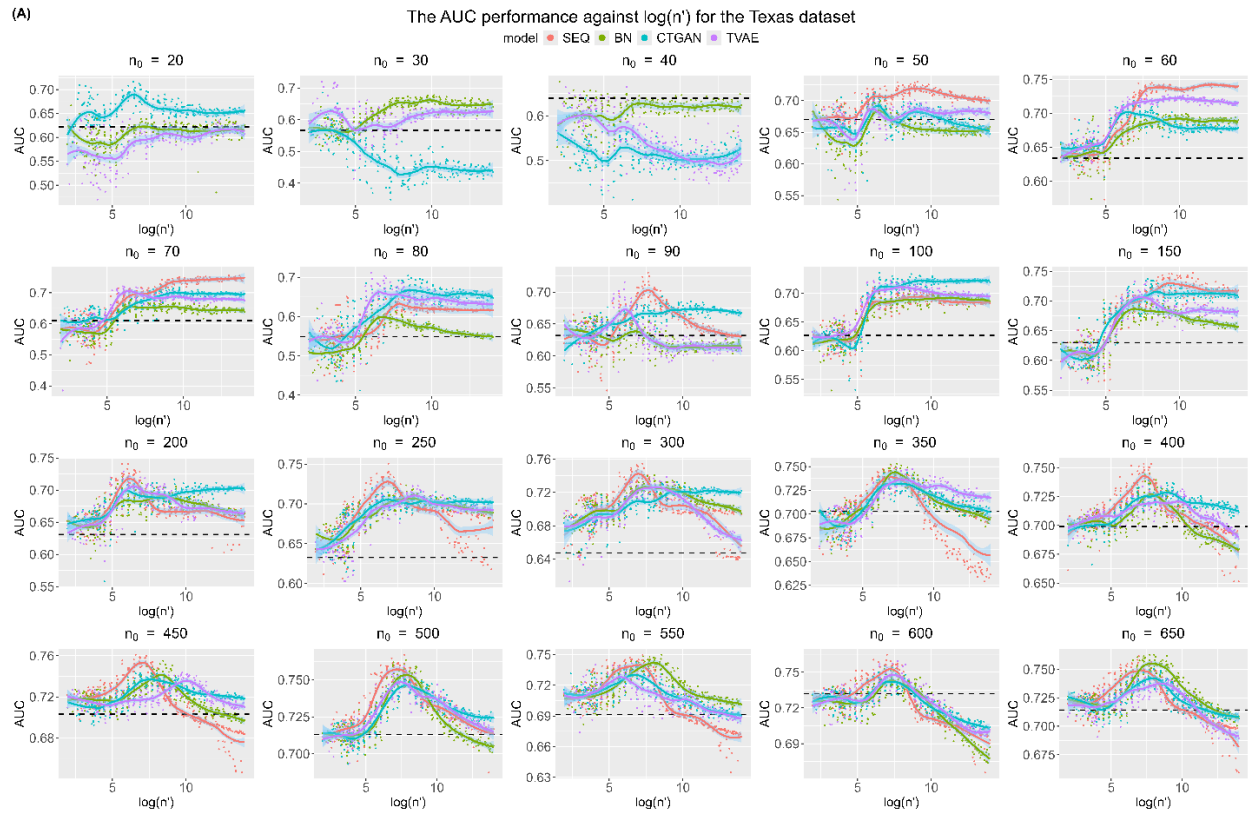

**Figure B.21:** Augmentation performance of AUC against log(*n'*) for the Texas dataset (A).

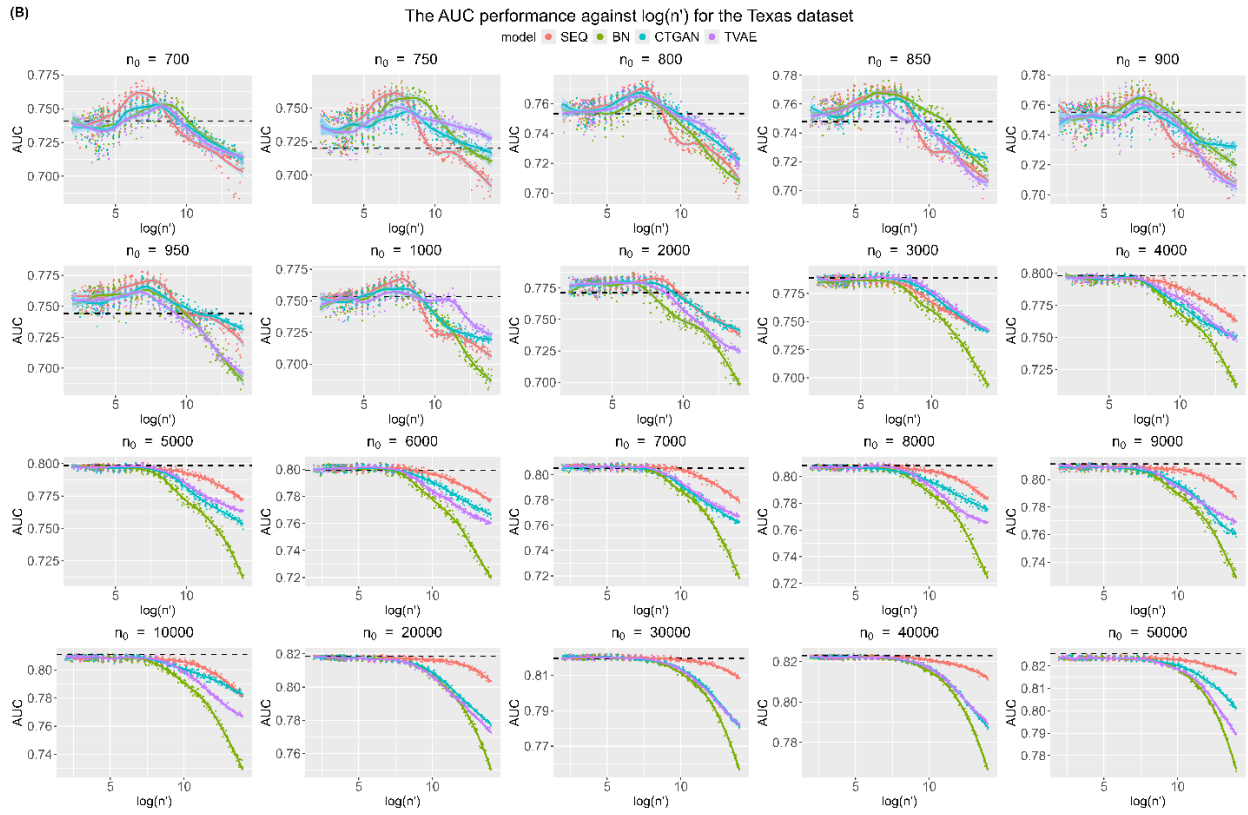

**Figure B.22:** Augmentation performance of AUC against log(*n'*) for the Texas dataset (B).

## B.12 Washington Dataset

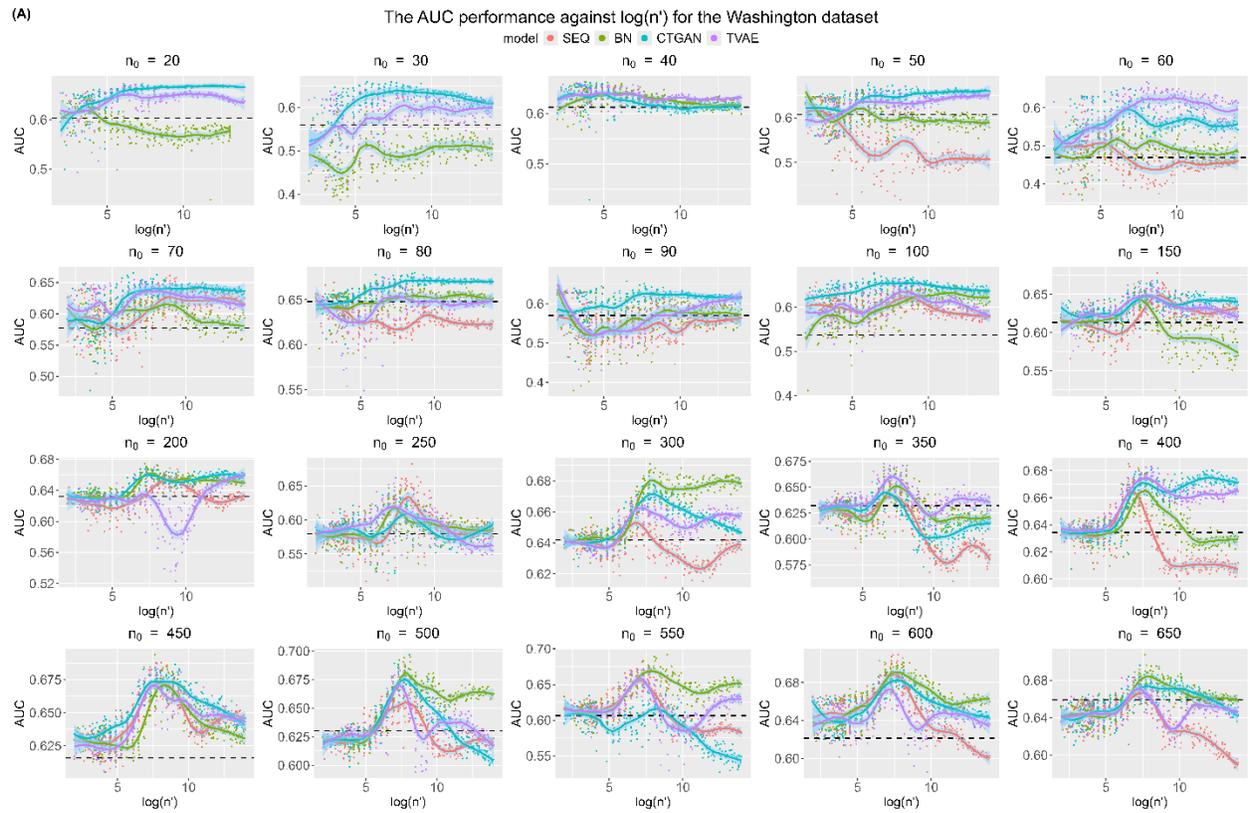

**Figure B.23:** Augmentation performance of AUC against log(*n'*) for the Washington dataset (A).

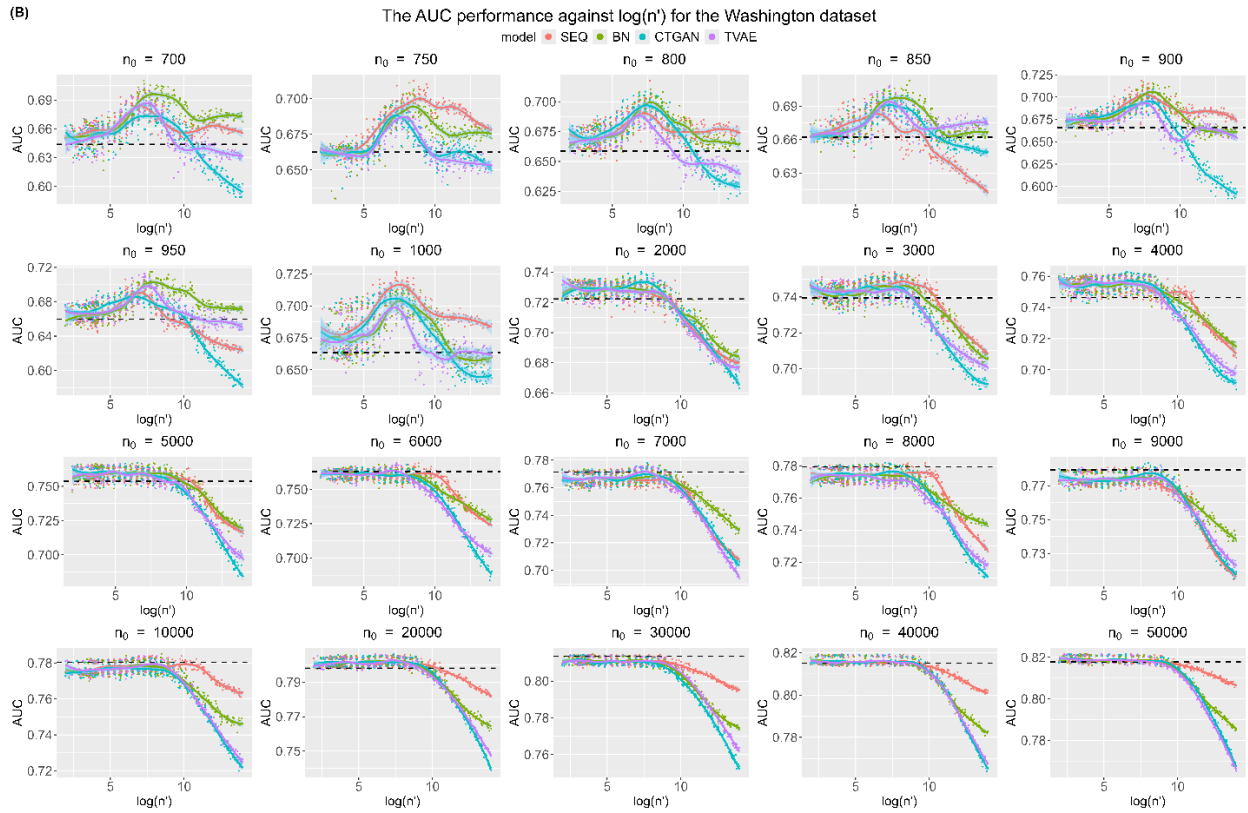

**Figure B.24:** Augmentation performance of AUC against log($n'$) for the Washington dataset (B).

## B.13 Washington2008 Dataset

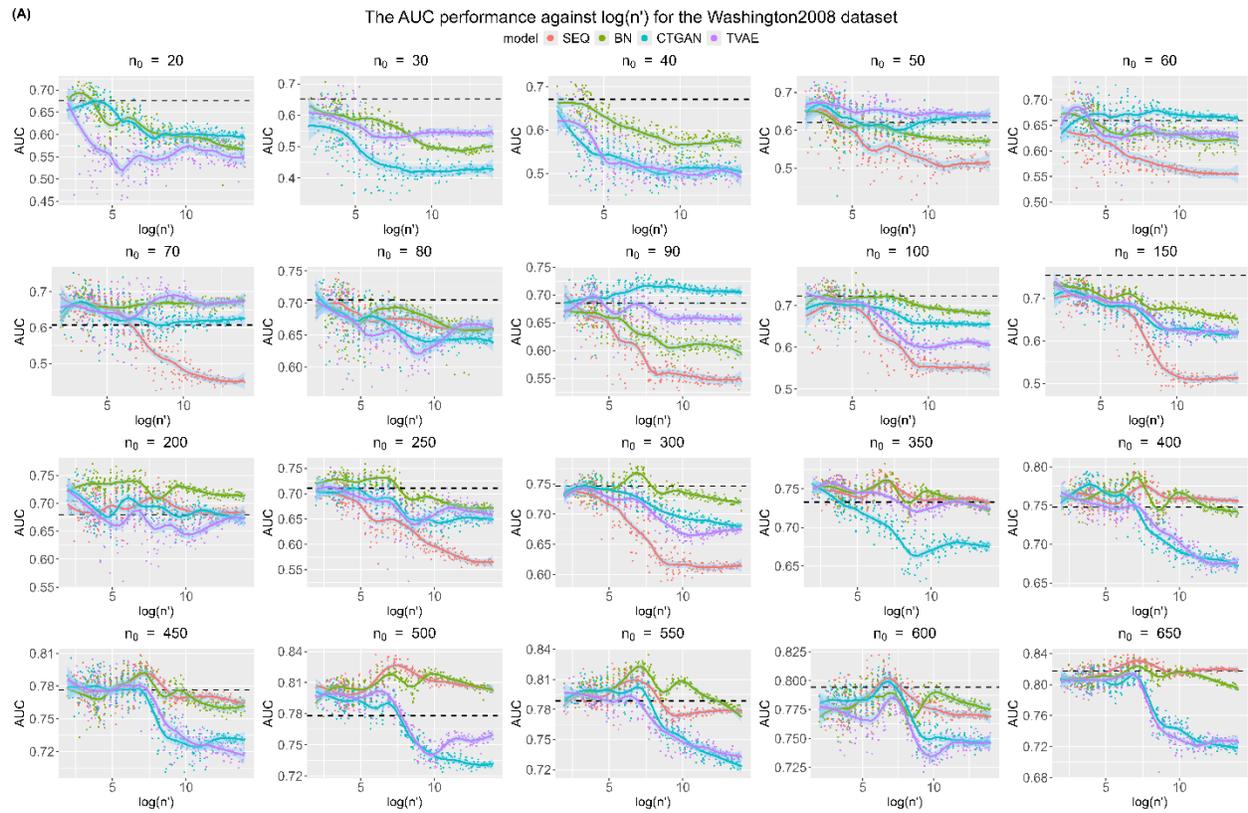

**Figure B.25:** Augmentation performance of AUC against log($n'$) for the Washington2008 dataset (A).

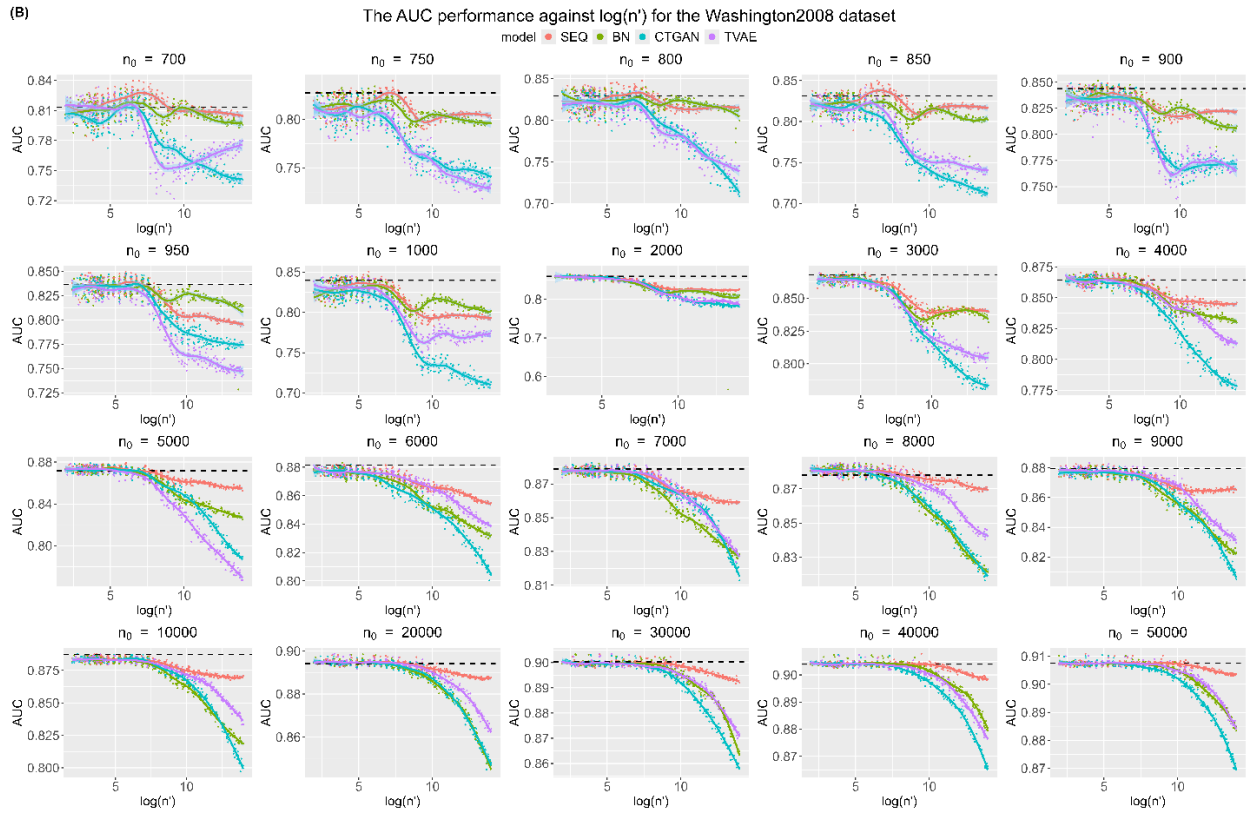

**Figure B.26:** Augmentation performance of AUC against log(*n'*) for the Washington2008 dataset (B).

# Appendix C - Results for case studies

## B.1  Application to Hot Flashes Dataset

The first dataset to be analyzed is a hot flashes survey dataset that was collected from patients with early breast cancer to understand the frequency and severity of vasomotor symptoms (VMS) and the effectiveness of previously applied interventions between June 5, 2020 and March 5, 2021 at two cancer in Ontario [1]. The original dataset contains 373 health records related to demographics, menopausal status, cancer-associated symptoms and treatments. The outcome is a binary variable that indicates whether the severity of the VMS problem is high or not, and the other seventeen variables were chosen as predictors, with detailed descriptions of these variables summarized in Table E.14. After preprocessing and removing the missing values in the outcome, the number of observations was reduced to 360. An LGBM model was built to predict the probability of severe VMS based on the relevant predictors.

Table C.1 displays the model performance from augmentation using the four generative models in terms of the number of synthetic observations that should be generated to achieve the optimal performance, baseline AUC, maximum AUC and maximum relative AUC in percentage.

| Model | $n'_{max}$ | Baseline AUC | Augmented AUC | Relative AUC (%) | Resampled AUC |
|---|---|---|---|---|---|
| SEQ | 25 | 0.7161 | 0.7488 | 4.57 | 0.7312 |
| BN | 198 | 0.7161 | 0.7497 | 4.69 | 0.6611 |
| **CTGAN** | **720** | **0.7161** | **0.7668** | **7.08** | **0.6477** |
| TVAE | 278 | 0.7161 | 0.7573 | 5.75 | 0.6783 |

**Table C.1:** Analysis results of augmentation performance for the Hot Flashes dataset. $n'_{max}$: $n'$ that leads to maximum AUC. Baseline AUC: baseline AUC from the base data. Augmented AUC: maximum AUC from the augmented data. Resampled AUC: AUC from the augmented data with a size of $n'_{max}$ using the bootstrap method.

The results show that augmentation has significantly improved the model performance as the data were augmented. In addition, CTGAN outperforms the other three models by boosting the ML performance by 7.08% with the baseline and maximum AUC values being 0.7161 and 0.7668 respectively, when 720 observations are simulated and added to the 360 observations, giving a total dataset of 1080 observations. The augmentation performance of bootstrap is overall shown to be worse than the four generative models, as the augmented AUC values are lower than the maximum AUC. It reveals that the diversifying the dataset has greater benefits in improving the model performance than simply replicating the observations to increase the sample size.

The baseline AUC for this dataset would be considered moderate to good [2]. Therefore, at this level of baseline performance augmentation does provide additional prognostic value.

## B.2  Application to Danish Colorectal Cancer Group Dataset

The Danish Colorectal Cancer Group (DCCG) database registered all patients in Denmark who were diagnosed with colorectal cancer or treated in a public Danish hospital since 2001 [3]. Patient data was acquired between 2001 and 2018 from the database, covering 12,855 clinical, surgical, radiological, and pathological records. After data preprocessing and removing the missingness in the postoperative medical complication, a total of 7,948 patient records remained, from which a random sample of 700 observations was drawn. In this analysis, we were interested in constructing a model to predict the probability of getting a postoperative medical complication based on associated risk factors, including age, gender, ASA score (i.e., pre-operative fitness score), localization of the tumor, procedure, pathological tumor (T) stage and node (N) stage, number of removed lymph nodes and number of lymph nodes with metastasis and unplanned intraoperative adverse event. A summary of statistics for the DCCG variables is presented in Table E.15. Similar to the Hot Flashes dataset, nested cross-validation was used to tune and evaluate an LGBM model.

Table C.2 summarizes the results for the DCCG dataset. Given the characteristics of the DCCG dataset, data augmentation was again recommended. It was found that by simulating another 720 synthetic observations from the TVAE generator on top of the original 700 observations, the optimal performance reaches an AUC of 0.7780 from the baseline AUC of 0.7171, leading to a relative AUC increase of 8.50%. In other words, the ML performance was enhanced from augmentation by 8.50% compared to the baseline performance without augmentation. Compared to TVAE, CTGAN performs similarly by improving the performance by up to 8.32%. Meanwhile, the performance of augmenting the analysis data to the same size $n'_{max}$ using the bootstrap method is inferior to using the four generative models, which further highlights the value of making the dataset more diverse.

| Model | $n'_{max}$ | Baseline AUC | Augmented AUC | Relative AUC (%) | Resampled AUC |
|---|---|---|---|---|---|
| SEQ | 8 | 0.7171 | 0.7630 | 6.41 | 0.7344 |
| BN | 26 | 0.7171 | 0.7583 | 5.75 | 0.7405 |
| CTGAN | 820 | 0.7171 | 0.7768 | 8.32 | 0.6995 |
| **TVAE** | **720** | **0.7171** | **0.7780** | **8.50** | **0.7077** |

**Table C.2:** Analysis results of augmentation performance for the Danish Colorectal Cancer Group dataset. $n'_{max}$: $n'$ that leads to maximum AUC. Baseline AUC: baseline AUC from the base data. Augmented AUC: maximum AUC from the augmented data. Resampled AUC: AUC from the augmented data with a size of $n'_{max}$ using the bootstrap method.

Similar to the previous case study, the baseline AUC for this dataset would be considered moderate to good [2]. Therefore, at this level of performance augmentation does provide additional prognostic value.

### B.3 Application to Breast Cancer Coimbra Dataset

The third dataset to be analyzed is Breast Cancer Coimbra dataset, in which a total of 116 women were recruited and screened with breast cancer by the Gynaecology Department of the University Hospital Centre of Coimbra between 2009 and 2013 [4]. The data preprocessing results in 10 variables across 116 records, covering patient's demographics, anthropometric measurements and information gathered by the routine blood tests. Table E.16 displays the detailed description of the chosen variables in the analysis. After computing the data characteristics for the Breast Cancer Coimbra analysis dataset, our decision model suggests that augmentation should be performed. We are building an LGBM model to predict the presence of breast cancer using age, body mass index, glucose, insulin, homeostasis model assessment, leptin, adiponectin, resistin, and MCP-1.

| Model | $n'_{max}$ | Baseline AUC | Augmented AUC | Relative AUC (%) | Resampled AUC |
|---|---|---|---|---|---|
| SEQ | 32 | 0.7392 | 0.8218 | 11.18 | 0.7599 |
| **BN** | **53** | **0.7392** | **0.8722** | **18.00** | **0.8291** |
| CTGAN | 71 | 0.7392 | 0.8416 | 13.85 | 0.7747 |
| TVAE | 94 | 0.7392 | 0.8490 | 14.86 | 0.7621 |

**Table C.3:** Analysis results of augmentation performance for the Breast Cancer Coimbra dataset. $n'_{max}$: $n'$ that leads to maximum AUC. Baseline AUC: baseline AUC from the base data. Augmented AUC: maximum AUC from the augmented data. Resampled AUC: AUC from the augmented data with a size of $n'_{max}$ using the bootstrap method.

The augmentation results for the Breast Cancer Coimbra dataset are summarized in Table C.3. The optimal performance was achieved by generating 53 more observations using Bayesian networks, leading to the maximum AUC value of 0.8722, compared to the baseline AUC of 0.7392. That leads to a remarkable increase in the relative AUC of 18.00%. In comparison, the performance of using bootstrap method to augment the original data is conspicuously much worse with the AUC being 0.8291. It demonstrates the value of incorporating diverse records rather than simply duplicating the records and the worth of using generative models to augment the data.

Similar to the previous case studies, the baseline AUC for this dataset would be considered moderate to good [2]. Therefore, at this level of performance augmentation does provide remarkable additional prognostic value.

### B.4 Application to Breast Cancer Dataset

The Breast Cancer dataset is a dataset comprising the health information related to the recurrence of breast cancer, provided by the University Medical Centre, Institute of Oncology in Yugoslavia [5]. The original and preprocessed data share the same size of 277 records and dimension of 10 variables. The statistical descriptives of the variables for the dataset is summarized in Table E.17. We used this dataset to predict the recurrence of breast cancer based on its associated demographic and clinical factors.

| Model | $n'_{max}$ | Baseline AUC | Augmented AUC | Relative AUC (%) | Resampled AUC |
|---|---|---|---|---|---|
| SEQ | 34 | 0.7143 | 0.7330 | 2.62 | 0.6500 |
| BN | 16 | 0.7143 | 0.7413 | 3.79 | 0.7060 |
| **CTGAN** | **25** | **0.7143** | **0.7451** | **4.31** | **0.6729** |
| TVAE | 21 | 0.7143 | 0.7406 | 3.69 | 0.7323 |

**Table C.4:** Analysis results of augmentation performance for the Breast Cancer dataset. $n'_{max}$: $n'$ that leads to maximum AUC. Baseline AUC: baseline AUC from the base data. Augmented AUC: maximum AUC from the augmented data. Resampled AUC: AUC from the augmented data with a size of $n'_{max}$ using the bootstrap method.

The analysis results are reported in Table C.4. The baseline AUC was found to be 0.7143. Among the four generative models, CTGAN outperformed the other three in terms of the maximum AUC of 0.7451, which was acquired by simulating additional 25 observations, resulting in relative change in AUC of 4.31%. Thus, the ML model performance was improved by 4.31% from augmenting the original data with extra 25 observations, compared to the baseline performance using the original data only. Moreover, the augmentation performance using bootstrap resampling is overall worse than using any generative model further spotlights the benefits of adding diverse information from the synthetic generative models. The fact that the augmentation performance using bootstrap resampling is even worse than the baseline performance further reveals that the bootstrap method might not be a good choice to enhance the model performance.

Similar to the previous case studies, the baseline AUC for this dataset would be considered moderate to good [2]. Therefore, at this level of performance augmentation does provide additional prognostic value.

### B.5 Application to Colposcopy/Schiller Dataset

The Colposcopy/Schiller dataset is one of the three modality colposcopy data that examines the subjective quality assessment of digital colposcopies collected by Hospital Universitario de Caracas [6]. The dataset contains 287 observations with one target variable and 62 variables as predictors related to color information, image area and coverage in digital cervical imaging. The target variable is considered as the outcome in the analysis of whether the subjective judgment is good or bad. A detailed description of the outcome and remaining variables is reported in Table E.18. The LGBM model was used to predict the subjective quality assessment based on the sixty-two imaging variables.

| Model | $n'_{max}$ | Baseline AUC | Augmented AUC | Relative AUC (%) | Resampled AUC |
|---|---|---|---|---|---|
| SEQ | 11 | 0.5125 | 0.6483 | 26.49 | 0.5454 |
| BN | 44 | 0.5125 | 0.6477 | 26.37 | 0.5147 |
| **CTGAN** | **2205** | **0.5125** | **0.7341** | **43.23** | **0.6116** |
| TVAE | 38 | 0.5125 | 0.6572 | 28.22 | 0.5628 |

Table C.5: Analysis results of augmentation performance for the Colposcopy/Schiller dataset. $n'_{max}$: $n'$ that leads to maximum AUC. Baseline AUC: baseline AUC from the base data. Augmented AUC: maximum AUC from the augmented data. Resampled AUC: AUC from the augmented data with a size of $n'_{max}$ using the bootstrap method.

The results are presented in Table C.5. The baseline performance yielded an AUC value of 0.5125. The best performance was achieved when the CTGAN was used to generate another set of 2,205 observations, resulting in an augmented AUC of 0.7341 and a substantial percentage improvement of 43.23% in the model performance. Moreover, the generative models also outperformed the bootstrap method. It reveals again that increasing the sample size without diversifying the dataset is less beneficial than increasing both the data size and diversity at the same time.

Similar to the previous case studies, the baseline AUC for this dataset would be considered poor [2]. Therefore, at this level of performance augmentation does provide pronounced additional prognostic value.

### B.6 Application to Diabetic Retinopathy Dataset

The Diabetic Retinopathy dataset describes detections of diabetic retinopathy represented by either detected lesions or image-level descriptors from Messidor image set [7]. The total size of the original data is 1,151, from which a sample with a size of 600 was randomly selected. The detailed description of the analysis dataset is described in Table E.19. We built an LGBM prediction model to predict the signs of diabetic retinopathy using all the available imaging features.

| Model | n'<sub>max</sub> | Baseline AUC | Augmented AUC | Relative AUC (%) | Resampled AUC |
|---|---|---|---|---|---|
| SEQ | 47 | 0.7400 | 0.7523 | 1.66 | 0.7281 |
| **BN** | **11534** | **0.7400** | **0.7974** | **7.75** | **0.7299** |
| CTGAN | 84 | 0.7400 | 0.7550 | 2.02 | 0.6943 |
| TVAE | 40737 | 0.7400 | 0.7600 | 2.70 | 0.7235 |

**Table C.6:** Analysis results of augmentation performance for the Diabetic Retinopathy dataset. $n'_{max}$: $n'$ that leads to maximum AUC. Baseline AUC: baseline AUC from the base data. Augmented AUC: maximum AUC from the augmented data. Resampled AUC: AUC from the augmented data with a size of $n'_{max}$ using the bootstrap method.

Table C.6 displays the augmentation performances using the four generative models and the bootstrap method. Given the baseline AUC is 0.7400, Bayesian networks led to the optimal performance with the maximum AUC being 0.7974 and the highest relative AUC increase of 7.75% when an additional 11,534 synthetic set was simulated and incorporated. On the other hand, the bootstrap method performed worse than the synthetic generative models and even worse than the baseline performance in most of the scenarios, which again shows the value of generating diverse data.

Similar to the previous case studies, the baseline AUC for this dataset would be considered moderate to good [2]. Therefore, at this level of performance augmentation does provide additional prognostic value.

## B.7    Application to Thoracic Surgery Dataset

The last case study uses the Thoracic Surgery dataset collected retrospectively by Wroclaw Thoracic Surgery Centre in Poland, which describes the post-operative life expectancy of lung cancer patients who underwent lung resections between 2007 and 2011 [8]. After data preprocessing, the analysis dataset remains 470 as the original data size, covering one outcome, 1-year survival, and sixteen variables related to demographics, clinical diagnostics, and post-operative complications. The characteristics of the variables are presented in Table E.20. The goal of this analysis is to predict the survival after 1 year period based on the associated factors.

| Model | n'<sub>max</sub> | Baseline AUC | Augmented AUC | Relative AUC (%) | Resampled AUC |
|---|---|---|---|---|---|
| SEQ | 46 | 0.5584 | 0.6151 | 10.14 | 0.5498 |
| BN | 3144 | 0.5584 | 0.6668 | 19.41 | 0.6731 |
| CTGAN | 1028 | 0.5584 | 0.6380 | 14.25 | 0.6446 |
| **TVAE** | **6602** | **0.5584** | **0.6700** | **19.98** | **0.6914** |

**Table C.7:** Analysis results of augmentation performance for the Thoracic Surgery dataset. $n'_{max}$: $n'$ that leads to maximum AUC. Baseline AUC: baseline AUC from the base data. Augmented AUC: maximum AUC from the augmented data. Resampled AUC: AUC from the augmented data with a size of $n'_{max}$ using the bootstrap method.

The analysis results are summarized in Table C.7. TVAE turns out to be the best generative model to achieve the optimal augmentation performance for the Thoracic Surgery dataset. The maximum AUC that TVAE can achieve is 0.6700, which is approximately 19.98% higher than the baseline AUC of 0.5584. In other words, the ML model performance was substantially enhanced by 19.98% from augmenting the original data, relative to the baseline performance using the original data only. However, the bootstrap performed slightly better than the generative models for this dataset.

Similar to the previous case studies, the baseline AUC for this dataset would be considered poor [2]. Therefore, at this level of performance augmentation does provide significant additional prognostic value.

# Appendix D - Hyperparameters

The following are the hyperparameters, their default values, and range for tuning the ensemble models.

| LGBM | | | | |
|---|---|---|---|---|
| **Hyperparameter** | **Default** | **Lower bound** | **Upper bound** | **Transform** |
| booster | 1 (gbdt) | 1 (gbdt) | 2 (goss) | 2^learning_rate |
| max_depth | 6 | 1 | 15 | |
| learning_rate | log2 (0.3) | -10 | 0 | |
| early_stopping_rounds | 7 | 7 | 30 | |
| min_data_in_leaf | 10 | 1 | 60 | |
| num_leaves | 15 | 4 | 60 | |
| **Random forest** | | | | |
| **Hyperparameter** | **Default** | **Lower bound** | **Upper bound** | **Transform** |
| num.trees | 500 | 1 | 2000 | round(n^min.node.size), where n is the number of observations |
| min.node.size | 0.5 | 0 | 1 | |
| max.depth | 15 | 1 | 50 | |
| min.bucket | 10 | 1 | 60 | |
| **XGBoost** | | | | |
| **Hyperparameter** | **Default** | **Lower bound** | **Upper bound** | **Transform** |
| gamma | 0 | -15 | 3 | 2^gamma |
| eta | log2(0.3) | -10 | 0 | 2^eta |
| max_depth | 6 | 1 | 15 | |
| early_stopping_rounds | 7 | 7 | 30 | |
| max_leaves | 15 | 4 | 60 | |
| min_child_weight | 1 | 0 | 7 | 2^min_child_weight |

**Table D.1:** The default values and ranges of hyperparameters in the models.

# Appendix E – Details of the datasets

The following are the details for the preprocessed datasets that were used prior to model training in this study.

## E.1    Canadian COVID-19

The first dataset is the Canadian COVID-19 dataset from the Public Health Agency of Canada. It contains over 1 million health records of individuals who have tested positive for COVID-19. We are interested in fitting a model that predicts mortality caused by COVID-19. The binary outcome of interest is derived from the case status in the dataset, and a value of 1 is assigned if the patient has died due to COVID-19 while a value 0 is assigned if the patient has recovered. The selected predictors for modeling include the following variables: date, age group, gender, region, exposure, province. Table E.1 presents an overview of the variables that are included in the binary model.

| Variable | Description | Type | Mean (SD) or level count (% of total size) or number of categories | Missingness (% of the total size) |
|---|---|---|---|---|
| Date | The date when a case is reported | Numeric (computed as the number of days since 1st January 2020) | 348.07 (96.16) | 0.00 |
| Age group | Patient's age group in years | Numeric | <20: 17.92%<br>20-29: 20.26%<br>30-39: 17.08%<br>40-49: 14.82%<br>50-59: 13.53%<br>60-69: 8.26%<br>70-79: 3.99%<br>>80: 2.16% | 1.99 |
| Gender | Patient's gender | Categorical | Female: 43.39%<br>Male: 50.05% | 0.00 |
| Region | Health unit in Canada | Categorical | 40 | 0.00 |
| Exposure | The type of being exposed to someone with COVID-19 | Categorical | Close contact: 31.21%<br>Outbreak: 10.78%<br>Travel-related: 1.14% | 0.00 |
| Province | Province in Canada where case is reported | Categorical | Ontario: 70.25%<br>Alberta: 29.75% | 0.00 |
| Case status | The status of a patient | Categorical | Recovered: 1.48 %<br>Deceased: 98.52% | 0.00 |

Note: SD: standard deviation

**Table E.1:** Descriptive statistics for the COVID-19 dataset.

## E.2 Canadian Community Health Survey

The CCHS data is a cross-sectional telephone survey administered by Statistics Canada that collects information on the health status, health care utilization and health determinants of Canadians. This dataset is a pooled version of survey data from 2001 to 2013, and the variables we are using are presented in Table E.2.

The model outcome is cardiovascular health and the covariates are age, sex, education, house income, household size, and immigration as predictors to predict the ideal state of cardiovascular health using variables from the dataset [9]. To assess cardiovascular health, we follow the definition of ideal cardiovascular health introduced by the American Heart Association to calculate the Cardiovascular Health in Ambulatory Care Research Team (CANHEART) health index score, which is determined by 7 health factors including smoking, obesity, hypertension, diabetes, physical activity, and fruit and vegetable consumption [10]. The final CANHEART index score ranges from 0 (worst) to 6 (best). The outcome is assigned to be 1 if the score is above 3 [11], which is considered to be an intermediate or ideal state of cardiovascular health and 0 otherwise.

| Variable | Description | Type | Mean (SD) or level count (% of total size) | Missingness (% of the total size) |
|---|---|---|---|---|
| Age | Patient's age in years | Numeric | 47.24 (20.19) | 0.00 |
| Sex | Patient's gender | Categorical | Female: 45.84%<br>Male: 54.16% | 0.00 |
| Education | Patient's highest level of education | Categorical | < Secondary school graduate: 26.14%<br>Secondary school graduate: 16.93%<br>Some post-secondary education: 6.65%<br>Post-secondary certificate: 48.55% | 1.72 |
| Marital status | Patient's marital status | Categorical | Married: 43.41%<br>Common-law: 8.12%<br>Widow/separation/divorce: 19.55%<br>Single/never married: 28.74% | 0.17 |
| House income | Total household income from all sources | Numeric | 57603.42 (32061.49) | 9.59 |
| Household size | Size of entire household | Numeric | 2.39 (1.23) | 15.66 |
| Immigration | Whether a patient is an immigrant | Categorical | Immigrant: 13.50%<br>Non-immigration: 84.11% | 1.81 |
| Smoking | Type of smoking | Categorical | Daily smoking: 17.92%<br>Occasional smoking: 2.64%<br>Always occasional smoking: 1.68%<br>Former daily smoking: 25.76%<br>Former occasional smoking: 14.56%<br>Never smoked: 36.91% | 0.00 |
| Obesity | Patient's self-reported BMI | Numeric | 25.80 (5.15) | 0.00 |
| Hypertension | Whether a patient was diagnosed with hypertension | Categorical | Yes: 20.23%<br>No: 79.46% | 0.00 |
| Diabetes | Whether a patient was diagnosed with diabetes | Categorical | Yes: 6.99%<br>No: 92.91% | 0.00 |
| Physical activity | Daily energy expenditure | Numeric | 2.20 (2.43) | 0.00 |
| Fruit and vegetable consumption | Daily consumption of fruits and vegetables | Numeric | 4.78 (2.53) | 0.00 |

| CANHEART | Whether a patient is in ideal cardiovascular health; this is a sum of the prior six variables. | Categorical | Ideal: 63.95% Non-ideal: 36.05% | 0.00 |

Note: SD: standard deviation

**Table E.2:** Descriptive statistics for the CCHS dataset.

### E.3    COVID-19 Survival

The COVID-19 survival dataset that is used in the study is a web-based survey data collected by the research team by Nexoid, a company in the United Kingdom. They collect demographic, socioeconomic and health-related information of individuals to predict two crucial aspects related to COVID-19: the probability of being infected with COVID-19 as well as the probability of mortality associated with COVID-19. In our study, we focus on the probability of COVID-19 infection using important demographic, behavioral and health factors including age, sex, race, smoking, nursing home, COVID-19 symptoms, COVID-19 contact, health worker, and the presence of comorbidities such as asthma, kidney disease, liver disease, heart disease, lung disease, diabetes, and hypertension. The outcome of interest is determined by the risk scores of getting infected with COVID-19. The patients whose risk scores exceed the mean risk score are considered as having a high risk of contracting COVID-19, while those with scores below the mean are classified as having a low risk. Table E.3 summarizes the basic statistics of the selected variables.

| Variable | Description | Type | Mean (SD) or level count (% of total size) | Missingness (% of the total size) |
|---|---|---|---|---|
| Age | Age group in years | Numeric | 0_10: 0.58%<br>10_20: 3.35%<br>20_30: 21.56%<br>30_40: 29.84%<br>40_50: 21.09%<br>50_60: 12.48%<br>60_70: 7.47%<br>70_80: 2.97%<br>80_90: 0.54%<br>90_100: 0.12%<br>100_110: 0.01% | 0.00 |
| Sex | Patient's gender | Categorical | Female: 63.13%<br>Male: 36.53% | 0.34 |
| Race | Patient's race | Categorical | White: 24.19%<br>Hispanic: 1.40%<br>Asian: 1.21%<br>Mixed: 0.96%<br>Black: 0.46%<br>Other: 0.32% | 71.45 |
| Smoking | Type of smoking | Categorical | Heavy: 1.68%<br>Medium: 7.64%<br>Light: 4.44%<br>Quit0: 5.54%<br>Quit5: 6.58%<br>Quit10: 9.23%<br>Vape: 5.95%<br>Never smoked: 58.74% | 0.19 |
| BMI | Body mass index | Numeric | 29.37 (7.81) | 0.00 |
| House count | House person count | Numeric | 3.14 (1.57) | 0.00 |
| Public transport count | Number of public transports used | Numeric | 0.38 (1.70) | 71.12 |
| Nursing home | Whether it is a nursing home | Categorical | 1: 0.07%<br>0: 99.93% | 0.00 |
| COVID-19 symptoms | Whether a patient shows symptoms of COVID-19 | Categorical | 1: 2.04%<br>0: 97.96% | 0.00 |
| COVID-19 contact | Whether a patient has close contact with someone infected with COVID-19 | Categorical | 1: 4.33%<br>0: 95.67% | 0.00 |
| Health worker | Whether a patient is a healthcare worker | Categorical | 1: 1.79%<br>0: 98.21% | 0.00 |

| Asthma | Whether a patient has asthma | Categorical | 1: 15.26%<br>0: 84.74% | 0.00 |
|---|---|---|---|---|
| Kidney disease | Whether a patient has kidney disease | Categorical | 1: 0.36%<br>0: 99.64% | 0.00 |
| Liver disease | Whether a patient has liver disease | Categorical | 1: 0.21%<br>0: 99.79% | 0.00 |
| Heart disease | Whether a patient has heart disease | Categorical | 1: 1.87%<br>0: 98.13% | 0.00 |
| Lung disease | Whether a patient has lung disease | Categorical | 1: 1.45%<br>0: 98.55% | 0.00 |
| Diabetes | Whether a patient has diabetes | Categorical | 1: 6.17%<br>0: 93.83% | 0.00 |
| Hypertension | Whether a patient has hypertension | Categorical | 1: 13.83%<br>0: 86.17% | 0.00 |
| Outcome | Whether a patient has a high risk of getting infected with COVID-19 | Categorical | Yes: 39.20%<br>No: 60.80% | 0.00 |

Note: SD: standard deviation

**Table E.3:** A summary of descriptive statistics for the COVID survival dataset.

### E.4 FDA Adverse Events

The next dataset contains the reports submitted to the FDA Adverse Event Reporting System for patients with adverse events. The binary outcome of interest for this dataset is whether or a patient has died. Our primary goal with this dataset is to explore the relationship between patient mortality and various predictors, including event date, gender, age, weight, drug name and the indication for drug use. Detailed statistics for these variables can be found in Table E.4.

| Variable | Description | Type | Mean (SD) or level count (% of total size) or number of categories | Missingness (% of the total size) |
|---|---|---|---|---|
| Outcome | Whether a patient has died | Categorical | Death: 9.94% Non-death: 90.06% | 0.00 |
| Event date | Date the adverse event occurred | Numeric (difference from 1/1/2020) | 466.52 (827.94) | 62.26 |
| Gender | Patient's gender | Categorical | Female: 51.82% Male: 37.85% | 10.33 |
| Age | Patient's age in years | Numeric | 55.90 (20.80) | 33.41 |
| Weight | Patient's weight in kg | Numeric | 73.05 (25.70) | 74.13 |
| Drug name | Name of medicinal product | Categorical | 10,545 | 0.00 |
| Indication | Medical terminology describing the indication for use | Categorical | 4,287 | 0.00 |

Note: SD: standard deviation

**Table E.4:** Descriptive statistics for the FAERS dataset.

### E.5  Texas Inpatients (2012)

Texas inpatient dataset contains 75 variables. Similar to the Washington state hospital discharge data, in this dataset, we explore the relationship between those demographic and health factors and the length of stay in the Texas hospitals. The involved covariates include age, sex, race, ethnicity, location, weekday, risk mortality, severity, DRG and fees with detailed descriptions in Table E.5. According to their length of stay in the hospital, the patients are classified into two groups, and the outcome is assigned a value of 1 if the patient's length of stay is greater than or equal to 3 days and 0 otherwise.

| Variable | Description | Type | Mean (SD) or level count (% of total size) or number of categories | Missingness (% of the total size) |
|---|---|---|---|---|
| Outcome | Whether a patient's length of stay is greater than 3 days | Categorical | 0: 40.42%<br>1: 59.58% | 0.00 |
| Age | Patient's age groups | Numeric | 0: 11.95%<br>1: 1.66%<br>2: 1.62%<br>3: 1.08%<br>4: 1.39%<br>5: 1.58%<br>6: 1.58%<br>7: 4.84%<br>8: 5.28%<br>9: 4.94%<br>10: 3.75%<br>11: 3.41%<br>12: 3.93%<br>13: 4.96%<br>14: 5.58%<br>15: 5.95%<br>16: 6.47%<br>17: 6.01%<br>18: 5.78%<br>19: 5.32%<br>20: 3.94%<br>21: 2.40%<br>22: 0.21%<br>23: 2.68%<br>24: 2.86%<br>25: 0.54%<br>26: 0.28% | 0.00 |

| Sex | Patient's gender | Categorical | Female: 56.25% | 6.58 |
| --- | --- | --- | --- | --- |
| | | | Male: 37.16% | |
| Race | Patient's race | Categorical | 1 American Indian/Eskimo/Aleut: 0.77% | 0.13 |
| | | | 2 Asian or Pacific Islander: 1.68% | |
| | | | 3 Black: 12.61% | |
| | | | 4 White: 61.45% | |
| | | | 5 Other: 23.35% | |
| Ethnicity | Whether a patient is of Hispanic origin | Categorical | 1 Hispanic Origin: 28.14% | 1.41 |
| | | | 2 Not of Hispanic Origin: 70.45% | |
| Location | Patient's mailing address in Texas and contiguous states | Categorical | AR: 0.48% | 0.01 |
| | | | FC: 0.25% | |
| | | | LA: 0.21% | |
| | | | NM: 0.57% | |
| | | | OK: 0.32% | |
| | | | TX: 97.21% | |
| | | | XX: 0.02% | |
| | | | ZZ: 0.92% | |
| Weekday | The day of week a patient is admitted | Categorical | 1 Monday: 16.97% | 0.00 |
| | | | 2 Tuesday: 17.22% | |
| | | | 3 Wednesday: 16.38% | |
| | | | 4 Thursday: 15.89% | |
| | | | 5 Friday: 14.98% | |
| | | | 6 Saturday: 9.42% | |
| | | | 7 Sunday: 9.14% | |
| Risk mortality | Risk of mortality score from the All Patient Refined (APR) Diagnosis Related Group (DRG) from the 3M™ APR-DRG Grouper. | Categorical | 0 No class specified: 0.10% | 0.00 |
| | | | 1 Minor: 60.15% | |
| | | | 2 Moderate: 20.26% | |
| | | | 3 Major: 13.26% | |

| | | | 4 Extreme: 6.22% | |
| Severity | Severity of illness score from the All Patient Refined (APR) Diagnosis Related Group (DRG) from the 3M™ APR-DRG Grouper. | Categorical | 0 No class specified: 0.10% <br> 1 Minor: 35.40% <br> 2 Moderate: 33.39% <br> 3 Major: 22.76% <br> 4 Extreme: 8.35% | 0.00 |
| DRG | All Patient Refined (APR) Diagnosis Related Group (DRG) as assigned by 3M APR-DRG Grouper | Categorical | 316 | 0.00 |
| Fees | Total non-covered amount of the charge | Numeric | 57.51 (1375.47) | 0.02 |

Note: SD: standard deviation

**Table E.5:** Descriptive statistics for the Texas inpatient dataset.

## E.6   Washington State Hospital Discharges (2007)

The seventh dataset, Washington State Hospital Discharge dataset, contains over 350 variables. Among these, we model the relationship between those demographic and health factors and the length of stay in the hospital. The covariates were: age, atype, aweekend, died, DRG, primary diagnosis code, and ZIP code. A detailed description of these variables is displayed in Table E.6. The outcome of our study categorizes patients into two groups based on their length of stay. A value of 1 is assigned if the patient's length of stay is greater than or equal to 3 days and 0 otherwise.

| Variable | Description | Type | Mean (SD) or 1 count (% of total size) or number of categories | Missingness (% of the total size) |
|---|---|---|---|---|
| Outcome | Whether a patient's length of stay is greater than 3 days | Categorical | Yes: 49.07% <br> No: 50.93% | 0.00 |
| Age | Patient's age in years | Numeric | 45.58 (28.45) | 0.01 |
| Atype | Admission type | Categorical | 1: 34.69% <br> 2: 18.03% <br> 3: 34.23% <br> 4: 12.81% <br> 5: 0.23% | 0.00 |
| Aweekend | Whether admission occurs on a weekend | Categorical | 1: 19.32% <br> 0: 80.68% | 0.00 |

| Died | Whether a patient died during hospitalization | Categorical | 1: 1.99%<br>0: 98.01% | 0.00 |
| --- | --- | --- | --- | --- |
| DRG | Diagnosis-related-group (DRG) in effect on discharge date | Categorical | 862 | 0.00 |
| DX1 | Primary diagnosis | Categorical | 5863 | 15.14 |
| ZIP | Patient's ZIP code | Categorical | 4271 | 0.06 |

Note: SD: standard deviation

**Table E.6:** Descriptive statistics for the Washington state hospital discharge dataset.

### E.7 Basic Stand Alone (BSA) Inpatient Claims

This dataset contains the claim-level information with each recording being an inpatient claim chosen from a 5% random sample of Medicare beneficiaries during 2008. In this study, we choose the variables including age, gender, DRG, ICD-9 primary procedure code, Medicare payment and the length of stay and explore the relationship between the length of stay and its relevant demographic and claim-related factors. The outcome is defined as a binary variable taking a value of 1 if the length of stay on the file is greater than or equal to 2.5 days, and 0 otherwise. Table E.7 provides an overview of the detailed statistics for these variables.

| Variable | Description | Type | Mean (SD) or 1 count (% of total size) or number of categories | Missingness (% of the total size) |
|---|---|---|---|---|
| Outcome | Whether the length of stay on a claim is greater than 2.5 days | Categorical | Yes: 57.35%<br>No: 42.65% | 0.00 |
| Age | The beneficiary's age | Numeric | 1 Under 65: 19.73%<br>2 65- 69: 13.19%<br>3 70-74: 14.65%<br>4 75-79: 15.55%<br>5 80-84: 16.10%<br>6 85 & older: 20.78% | 0.00 |
| Gender | The beneficiary's gender | Categorical | 1 Male: 56.12%<br>2 Female: 43.88% | 0.00 |
| DRG | Diagnostic related groups to which a hospital claim belongs for prospective payment purposes | Categorical | 311 | 0.00 |
| ICD-9 | Primary procedure (primarily surgical procedures) performed during the inpatient stay | Categorical | 85 | 47.00 |
| Payment | Quintile value (or code) to which the actual Medicare payment amount on the claim belongs | Categorical | 1: 19.97%<br>2: 20.12%<br>3: 20.00%<br>4: 19.82%<br>5: 20.09% | 0.00 |

Note: SD: standard deviation

**Table E.7:** Descriptive statistics for the Basic Stand Alone inpatient claims dataset.

## E.8 Washington State Hospital Discharges (2008)

This dataset contains 652,340 inpatient discharge records in 2008 from community hospitals in Washington from State Inpatient Databases that are used to track the trends in healthcare utilization, access, charges, quality and outcomes in the United States. We are interested in examining the relationship between length of stay and demographic and health factors. Specifically, the covariates of interest include age, female, race, admission type, aweekend, DRG, DX1, primary payer, total charges, zip code, chronic conditional indicators, and procedure classes for ICD-10-PCS procedure codes, comorbidity measures for alcohol abuse, depression, hypertension and obesity. The outcome is created by classifying the patients into two groups based on the median of their length of stay. A value of 1 is assigned if the patient's length of stay is greater than or equal to 2 days and 0 otherwise. Detailed statistics of the variables are displayed in Table E.8.

| Variable | Description | Type | Mean (SD) or level count (% of total size) or number of categories | Missingness (% of the total size) |
|---|---|---|---|---|
| Outcome | Whether a patient's length of stay is greater than 2 days | Categorical | 1: 74.58%<br>0: 25.42% | 0.00 |
| AGE | Patient's age in years | Numeric | 45.79 (28.43) | 0.01 |
| FEMALE | Whether a patient's gender is female | Categorical | 1 Female: 58.69%<br>0 Male: 41.31% | 0.01 |
| RACE | Patient's race | Categorical | 1 White: 23.86%<br>2 Black: 1.18%<br>3 Hispanic: 2.80%<br>4 Asian or Pacific Islander: 1.31%<br>5 Native American: 0.48%<br>6 Other: 0.02% | 70.34 |
| ATYPE | Admission type | Categorical | 1 Emergency: 35.59%<br>2 Urgent: 17.80%<br>3 Elective: 33.22%<br>4 Newborn: 12.74%<br>5 Trauma Center: 0.65% | 0.00 |
| AWEEKEND | Whether a patient's admission day is on a weekend | Categorical | 1 Admitted Saturday - Sunday: 19.50%<br>0 Admitted Monday - Friday: 80.50% | 0.00 |
| DRG | Diagnosis Related Group | Categorical | 746 | 0.00 |
| DX1 | ICD-9-CM Diagnosis | Categorical | 6149 | 0.00 |
| PAY1 | Expected primary payer (Medicare, Medicaid, private insurances, etc.) | Categorical | 1 Medicare: 31.21%<br>2 Medicaid: 20.03%<br>3 Private insurance: 42.89%<br>4 Self-pay: 2.80%<br>5 No charge: 0.60%<br>6 Other: 2.46% | 0.00 |
| TOTCHG | Total charges | Numeric | 26040.52 (43943.20) | 0.01 |
| ZIP | Zip code | Categorical | 4192 | 0.00 |
| CHRON1 | ICD-9-CM Chronic | Categorical | 1 Chronic condition: 35.16% | 0.01 |

| | Condition Indicators | | 0 Non-chronic condition: 64.82% | |
| --- | --- | --- | --- | --- |
| CHRONB1 | Chronic Condition Indicators - body system | Categorical | 18 | 0.01 |
| PCLASS1 | Procedure Classes Refined for ICD-10-PCS procedure codes | Categorical | 1 Minor diagnostic: 6.66%<br>2 Minor therapeutic: 24.98%<br>3 Major diagnostic: 0.44%<br>4 Major Therapeutic: 31.07% | 36.84 |
| CM_ALCOH | AHRQ comorbidity measure for ICD-9-CM codes: alcohol abuse | Categorical | 1 Comorbidity is present: 3.01%<br>0 Comorbidity is not present: 96.99% | 0.00 |
| CM_DEPRE | AHRQ comorbidity measure for ICD-9-CM codes: depression | Categorical | 1 Comorbidity is present: 6.31%<br>0 Comorbidity is not present: 93.69% | 0.00 |
| CM_HTN_C | AHRQ comorbidity measure for ICD-9-CM codes: hypertension (combine uncomplicated and complicated) | Categorical | 1 Comorbidity is present: 28.69%<br>0 Comorbidity is not present: 71.31% | 0.00 |
| CM_OBESE | AHRQ comorbidity measure for ICD-9-CM codes: obesity | Categorical | 1 Comorbidity is present: 5.55%<br>0 Comorbidity is not present: 94.45% | 0.00 |

Note: SD: standard deviation

**Table E.8:** Descriptive statistics for the hospital Washington dataset.

## E.9  California Hospital Discharges (2008)

This dataset contains over 4 million inpatient discharge records in 2008 from community hospitals in California from State Inpatient Databases that are used to track the trends in healthcare utilization, access, charges, quality and outcomes in United States. We are interested in exploring the relationship between length of stay and its demographic and health factors. Specifically, the covariates of interest include age, female, race, aweekend, DRG, DX1, primary payer, total charges, chronic conditional indicators, and procedure classes for ICD-10-PCS procedure codes, comorbidity measures for alcohol abuse, depression, hypertension and obesity. The outcome is generated by dividing the patients into two groups based on the median of their length of stay. A value of 1 is assigned if the patient's length of stay is greater than or equal to 3 days and 0 otherwise. Detailed statistics of the variables are displayed in Table E.9.

| Variable | Description | Type | Mean (SD) or level count (% of total size) or number of categories | Missingness (% of the total size) |
|---|---|---|---|---|
| Outcome | Whether a patient's length of stay is greater than 3 days | Categorical | 1: 54.10%<br>0: 45.90% | 0.00 |
| AGE | Patient's age in years | Numeric | 44.59 (28.58) | 0.95 |
| FEMALE | Whether a patient's gender is female | Categorical | 1 Female: 57.24%<br>0 Male: 39.85% | 2.91 |
| RACE | Patient's race | Categorical | 1 White: 46.72%<br>2 Black: 7.27%<br>3 Hispanic: 28.52%<br>4 Asian or Pacific Islander: 7.07%<br>5 Native American: 0.07%<br>6 Other: 2.04% | 8.31 |
| AWEEKEND | Whether a patient's admission day is on a weekend | Categorical | 1 Admitted Saturday - Sunday: 20.39%<br>0 Admitted Monday - Friday: 79.61% | 0.00 |
| DRG | Diagnosis Related Group | Categorical | 746 | 0.00 |
| DX1 | ICD-9-CM Diagnosis | Categorical | 8548 | 0.00 |
| PAY1 | Expected primary payer (Medicare, Medicaid, private insurances, etc.) | Categorical | 1 Medicare: 31.13%<br>2 Medicaid: 25.57%<br>3 Private insurance: 34.79%<br>4 Self-pay: 3.41%<br>6 Other: 5.09% | 0.02 |
| TOTCHG | Total charges | Numeric | 45065.28 (78294.38) | 12.00 |
| CHRON1 | ICD-9-CM Chronic Condition Indicators | Categorical | 1 Chronic condition: 34.22%<br>0 Non-chronic condition: 65.78% | 0.00 |
| CHRONB1 | Chronic Condition Indicators - body system | Categorical | 18 | 0.00 |
| PCLASS1 | Procedure Classes Refined for ICD-10-PCS procedure codes | Categorical | 1 Minor diagnostic: 9.65%<br>2 Minor therapeutic: 28.12% | 35.40 |

|  |  |  | 3 Major diagnostic: 0.46% |  |
|  |  |  | 4 Major Therapeutic: 26.37% |  |
| CM_ALCOH | AHRQ comorbidity measure for ICD-9-CM codes: alcohol abuse | Categorical | 1 Comorbidity is present: 3.84%<br>0 Comorbidity is not present: 96.16% | 0.00 |
| CM_DEPRE | AHRQ comorbidity measure for ICD-9-CM codes: depression | Categorical | 1 Comorbidity is present: 5.80%<br>0 Comorbidity is not present: 94.20% | 0.00 |
| CM_HTN_C | AHRQ comorbidity measure for ICD-9-CM codes: hypertension (combine uncomplicated and complicated) | Categorical | 1 Comorbidity is present: 33.31%<br>0 Comorbidity is not present: 66.69% | 0.00 |
| CM_OBESE | AHRQ comorbidity measure for ICD-9-CM codes: obesity | Categorical | 1 Comorbidity is present: 7.23%<br>0 Comorbidity is not present: 92.77% | 0.00 |

Note: SD: standard deviation

**Table E.9:** Descriptive statistics for the hospital California dataset.

## E.10 Florida Hospital Discharges (2007)

This dataset contains over 2.3 million inpatient discharge records in 2007 from community hospitals in Florida from State Inpatient Databases that are used to track the trends in healthcare utilization, access, charges, quality and outcomes in United States. We are interested in exploring the relationship between length of stay and its demographic and health factors. Specifically, the covariates of interest include age, female, race, admission type, aweekend, DRG, DX1, primary payer, total charges and zip code. The outcome is created by classifying the patients into two groups based on the median of their length of stay. A value of 1 is assigned if the patient's length of stay is greater than or equal to 3 days and 0 otherwise. Detailed statistics of the variables are displayed in Table E.10.

| Variable | Description | Type | Mean (SD) or level count (% of total size) or number of categories | Missingness (% of the total size) |
|---|---|---|---|---|
| Outcome | Whether a patient's length of stay is greater than 3 days | Categorical | 1: 60.46%<br>0: 39.54% | 0.00 |
| AGE | Patient's age in years | Numeric | 51.23 (27.04) | 0.00 |
| FEMALE | Whether a patient's gender is female | Categorical | 1 Female: 56.02%<br>0 Male: 43.98% | 0.00 |
| RACE | Patient's race | Categorical | 1 White: 65.53%<br>2 Black: 16.92%<br>3 Hispanic: 13.28%<br>4 Asian or Pacific Islander: 0.76%<br>5 Native American: 0.27%<br>6 Other: 2.47% | 0.76 |
| ATYPE | Admission type | Categorical | 1 Emergency: 54.15%<br>2 Urgent: 16.43%<br>3 Elective: 20.73%<br>4 Newborn: 8.17%<br>5 Trauma Center: 0.52% | 0.00 |
| AWEEKEND | Whether a patient's admission day is on a weekend | Categorical | 1 Admitted Saturday - Sunday: 19.51%<br>0 Admitted Monday - Friday: 80.49% | 0.00 |
| DRG | Diagnosis Related Group | Categorical | 861 | 0.00 |
| DX1 | ICD-9-CM Diagnosis | Categorical | 7380 | 0.00 |
| PAY1 | Expected primary payer (Medicare, Medicaid, private insurances, etc.) | Categorical | 1 Medicare: 42.71%<br>2 Medicaid: 17.50%<br>3 Private insurance: 27.52%<br>4 Self-pay: 6.23%<br>5 No charge: 2.28%<br>6 Other: 3.77% | 0.00 |
| TOTCHG | Total charges | Numeric | 33604.48 (52812.95) | 0.01 |
| ZIP | Zip code | Categorical | 14729 | 0.00 |

Note: SD: standard deviation

**Table E.10:** Descriptive statistics for the hospital Florida dataset.

## E.11   New York Hospital Discharges (2007)

This dataset consists of over 2.6 million inpatient discharge records in 2007 from community hospitals in New York from State Inpatient Databases that are used to track the trends in healthcare utilization, access, charges, quality and outcomes in the United States. We are interested in examining the relationship between length of stay and demographic and health factors. Specifically, the covariates of interest include age, female, race, admission type, aweekend, DRG, DX1, primary payer, total charges, zip code, chronic conditional indicators and procedure classes for ICD-10-PCS procedure codes. The outcome is created by classifying the patients into two groups based on the median of their length of stay. A value of 1 is assigned if the patient's length of stay is greater than or equal to 3 days and 0 otherwise. Detailed statistics of the variables are displayed in Table E.11.

| Variable | Description | Type | Mean (SD) or level count (% of total size) or number of categories | Missingness (% of the total size) |
|---|---|---|---|---|
| Outcome | Whether a patient's length of stay is greater than 3 days | Categorical | 1: 61.82%<br>0: 38.18% | 0.00 |
| AGE | Patient's age in years | Numeric | 48.87 (27.36) | 0.00 |
| FEMALE | Whether a patient's gender is female | Categorical | 1 Female: 56.68%<br>0 Male: 43.32% | 0.00 |
| RACE | Patient's race | Categorical | 1 White: 56.73%<br>2 Black: 17.43%<br>3 Hispanic: 13.56%<br>4 Asian or Pacific Islander: 3.38%<br>5 Native American: 1.01%<br>6 Other: 5.90% | 1.98 |
| ATYPE | Admission type | Categorical | 1 Emergency: 60.22%<br>2 Urgent: 9.86%<br>3 Elective: 20.86%<br>4 Newborn: 8.90%<br>5 Trauma Center: 0.00% | 0.16 |
| AWEEKEND | Whether a patient's admission day is on a weekend | Categorical | 1 Admitted Saturday - Sunday: 19.26%<br>0 Admitted Monday - Friday: 80.74% | 0.00 |
| DRG | Diagnosis Related Group | Categorical | 863 | 0.00 |
| DX1 | ICD-9-CM Diagnosis | Categorical | 7956 | 0.00 |
| PAY1 | Expected primary payer (Medicare, Medicaid, private insurances, etc.) | Categorical | 1 Medicare: 36.08%<br>2 Medicaid: 23.69%<br>3 Private insurance: 32.38%<br>4 Self-pay: 5.44%<br>5 No charge: 0.17%<br>6 Other: 2.23% | 0.00 |
| TOTCHG | Total charges | Numeric | 24628.84 (43545.43) | 0.01 |
| ZIP | Zip code | Categorical | 10814 | 0.00 |

| CHRON1 | ICD-9-CM Chronic Condition Indicators | Categorical | 1 Chronic condition: 40.86%<br>0 Non-chronic condition: 59.14% | 0.00 |
|---|---|---|---|---|
| CHRONB1 | Chronic Condition Indicators - body system | Categorical | 18 | 0.00 |
| PCLASS1 | Procedure Classes Refined for ICD-10-PCS procedure codes | Categorical | 1 Minor diagnostic: 15.19%<br>2 Minor therapeutic: 31.30%<br>3 Major diagnostic: 0.54%<br>4 Major Therapeutic: 25.86% | 27.12 |

Note: SD: standard deviation

**Table E.11:** Descriptive statistics for the hospital New York dataset.

### E.12 Better Outcomes Registry & Network

Data are collected from the BORN Ontario birth registry that covers about 1 million records regarding Ontario's maternal demographic characteristics, obstetrical history, health behaviors, prenatal screening and newborn care information. We combine the pregnancy and infant datasets and examine the association between low birthweight and its related risk factors. The relevant factors include as gestational age, maternal age, maternal body mass index, total number of pregnancies a mother has experienced, number of previous preterm pregnancies, number of previous abortions, maternal smoking status, alcohol exposure, prenatal screening, mental health concerns for addiction, anxiety, depression, maternal health conditions for diabetes and genetics and drug exposure to Cocaine, Hallucinogens and Opioids. We follow the definition of low birthweight[25] and classify the newborns whose birth weights are less than 2,500 grams as infants with low birthweight. A value of 1 is given for newborns with low birthweight and 0 otherwise. A summary of descriptive statistics for the variables is presented in Table E.12.

| Variable | Description | Type | Mean (SD) or level count (% of total size) or number of categories | Missingness (% of the total size) |
|---|---|---|---|---|
| Birth weight | Whether a newborn baby has low birthweight (<2,500 grams) | Categorical | 1: 7.01%<br>0: 92.99% | 0.00 |
| Gestational age | Gestational age of a newborn baby | Numeric | 1 < 34 weeks: 2.32%<br>2 34-36 weeks: 6.00%<br>3 37-38 weeks: 27.15%<br>4 39-41 weeks: 64.11%<br>5 >=42 weeks: 0.43% | 0.00 |
| Maternal age | Maternal age in years at time of stillbirth or live birth | Numeric | 1 <= 19: 2.15%<br>2 20-34: 74.12%<br>3 35-39: 19.31%<br>4 >=40: 4.36% | 0.06 |
| Maternal BMI | Maternal pre-pregnancy body mass index | Numeric | 1 <18.5: 4.53%<br>2 18.5-24.9: 43.44%<br>3 25-29.9: 20.22%<br>4 >=30: 15.86% | 15.94 |
| Parity | Total number of pregnancies a mother has experienced | Numeric | 0: 42.82%<br>1: 34.57%<br>2: 13.98%<br>3: 4.64%<br>>=4: 2.89% | 1.10 |
| Preterm birth | Number of previous preterm pregnancies | Numeric | 0: 93.57%<br>1: 4.41%<br>2: 0.71%<br>3: 0.13%<br>>=4: 0.05% | 1.13 |
| Abortions | Number of previous abortions | Numeric | 0: 65.77%<br>1: 20.85%<br>2: 7.25%<br>3: 2.50%<br>>=4: 1.50% | 2.13 |
| Smoking | Maternal smoking status at time of admission | Categorical | Yes: 7.64%<br>No: 88.14% | 4.21 |

| Alcohol | Alcohol exposure in pregnancy | Categorical | Yes: 2.23%<br>No: 92.54% | 5.23 |
|---|---|---|---|---|
| Prenatal screening | Whether a mother has prenatal screening during pregnancy | Categorical | Yes: 66.30%<br>No: 33.70% | 0.00 |
| Addiction | Mental health concern regarding addiction | Categorical | Yes: 0.60%<br>No: 93.72% | 5.68 |
| Anxiety | Mental health concern regarding anxiety | Categorical | Yes: 8.99%<br>No: 85.33% | 5.68 |
| Depression | Mental health concern regarding depression | Categorical | Yes: 7.58%<br>No: 86.74% | 5.68 |
| Diabetes | Maternal health condition regarding diabetes | Categorical | Yes: 1.00%<br>No: 93.27% | 5.73 |
| Genetics | Maternal health condition regarding genetics | Categorical | Yes: 0.00%<br>No: 94.26% | 5.73 |
| Cocaine drug | Drug exposure to Cocaine in pregnancy | Categorical | Yes: 0.25%<br>No: 94.65% | 5.10 |
| Hallucinogens drug | Drug exposure to Hallucinogens in pregnancy | Categorical | Yes: 0.02%<br>No: 94.88% | 5.10 |
| Opioids drug | Drug exposure to Opioids in pregnancy | Categorical | Yes: 0.42%<br>No: 94.48% | 5.10 |

Note: SD: standard deviation

**Table E.12:** Descriptive statistics for the BORN dataset.

### E.13 Medical Information Mart for Intensive Care III

The dataset is extracted from the MIMIC-III relational database (version 1.4), which contains deidentified clinical data of the patients who were admitted to the Beth Israel Deaconess Medical Center in Boston, Massachusetts [12–14]. It contains various tables of patient data regarding demographics, admission information, lab tests, diagnosis codes, caregiver information, and discharge notes. We use this dataset to investigate the relationship between 30-day readmission and its related demographics, vital signs and lab test values. The demographics include the age of the patients when they were first admitted to the ICU, their ethnicity group and admission type. The vital signs consider the (systolic and diastolic) blood pressure, heart rate and respiration rate. Several lab measurements are also incorporated into the analysis. The selection criteria for readmitted patients are to include those who were readmitted within 30-day of initial hospital discharge from the ICU. The patients who were readmitted to the ICU are assigned a label of 1, while those who were not readmitted are assigned a label of 0. Table E.13 summarizes the descriptive statistics of the selected variables.

| Variable | Description | Type | Mean (SD) or level count (% of total size) or number of categories | Missingness (% of the total size) |
|---|---|---|---|---|
| Readmission | Whether a patient is re-admitted to ICU | Categorical | 1 Yes: 21.01%<br>0 No: 78.99% | 0.00 |
| Age | Patient's age in the time of first admission | Numeric | 63.43 (16.16) | 0.00 |
| Ethnicity | Patient's ethnicity group | Categorical | 38 | 15.14% |
| Admission type | Patient's admission type | Categorical | Elective: 18.10%<br>Emergency: 78.52%<br>Urgent: 3.37% | 0.00 |
| Heart rate | Vital sign for heart rate | Numeric | 87.86 (15.89) | 0.47% |
| NT-proBNP | Lab test for N-terminal prohormone of brain natriuretic peptide | Numeric | 4.10 (1.17) | 43.55% |
| Creatinine | Lab test for serum creatinine | Numeric | 4.10 (1.17) | 43.54% |
| Bun | Lab test for blood urea nitrogen | Numeric | 4.10 (1.17) | 43.53% |
| Potassium | Lab test for potassium | Numeric | 4.10 (1.17) | 43.52% |
| Cholesterol | Lab test for cholesterol | Numeric | 4.10 (1.17) | 43.54% |

Note: SD: standard deviation

**Table E.13:** Descriptive statistics for the MIMIC-III dataset.

## E.14 Hot Flashes

The Hot Flashes is a survey dataset that was collected from 373 patients with early breast cancer to understand the frequency and severity of vasomotor symptoms (VMS) and the effectiveness of previously applied interventions between June 5, 2020 and March 5, 2021 at two cancer in Ontario [1]. The outcome of interest is a binary variable, representing whether the severity of the VMS problem for a patient is high (coded as 1) or not (coded as 0). The dataset also collected a number of associated risk factors, including demographics, symptoms of having hot flashes, and medical treatments for hot flashes. A summary of the selected analysis variables is reported in Table E.14.

| Variable | Description | Type | Mean (SD) or level count (% of total size) | Missingness (% of the total size) |
|---|---|---|---|---|
| Severity | Whether a patient had severe VMS (outcome) | Categorical | 1: 49.44% 0: 50.56% | 0.00% |
| Age | Age in years | Numeric | 56.31 (10.52) | 0.00% |
| Assessment of VMS | Whether a patient is asked about/assessed for symptoms of hot flashes by HCP during clinic visits | Categorical | 1: 58.33% 0: 35.83% | 5.83% |
| Hot flashes per week | The number of hot flashes in a week that occurred in the past week | Numeric | 28.74 (47.00) | 18.06% |
| Menopausal status | Current self-reported menopausal status | Categorical | 1: 55.00% 0: 36.11% | 8.89% |
| Feeling extremely hot/sweaty | Whether a patient had the symptom of feeling extremely hot/sweaty | Categorical | 1: 76.67% 0: 23.33% | 0.00% |
| Redness of my face/chest | Whether a patient had the symptom of redness on the face/chest | Categorical | 1: 18.89% 0: 81.11% | 0.00% |
| Feeling chills/clammy after hot flashes have passed | Whether a patient had the symptom of feeling chills/clammy after hot flashes passed | Categorical | 1: 26.94% 0: 73.06% | 0.00% |
| Waking up at night/difficulty sleeping | Whether a patient had the symptom of wake-up at night/having difficulty sleeping | Categorical | 1: 55.56% 0: 44.44% | 0.00% |
| Irritability | Whether a patient had irritability problems | Categorical | 1: 12.22% 0: 87.78% | 0.00% |
| Memory problems | Whether a patient had memory problems | Categorical | 1: 10.83% 0: 89.17% | 0.00% |
| Endocrine therapy | Endocrine therapy treatment for breast cancer | Categorical | 1: 88.61% 0: 11.39% | 0.00% |
| Ovarian function suppression | Ovarian function suppression treatment for breast cancer | Categorical | 1: 19.44% 0: 80.56% | 0.00% |
| Chemotherapy | Chemotherapy treatment for breast cancer | Categorical | 1: 56.94% 0: 43.06% | 0.00% |
| Change to breast cancer treatment | Changes made to breast cancer treatment due to hot flashes | Categorical | 1: 18.33% 0: 81.67% | 0.00% |
| Drug treatment for VMS | VMS treatments prescribed, recommended or tried by the patient | Categorical | 1: 31.11% 0: 68.89% | 0.00% |
| CAM therapies for VMS | Complementary treatments prescribed, recommended or tried by the patient | Categorical | 1: 17.22% 0: 82.78% | 0.00% |

| Referral to a menopause clinic | Patient referred or seen by a gynecologist or dedicated menopause clinic to assist in managing hot flashes | Categorical | 1: 6.67%<br>0: 93.06% | 0.28% |

Note: SD: standard deviation

**Table E.14:** A summary of descriptive statistics for the Hot Flashes dataset. SD: standard deviation.

## E.15   Danish Colorectal Cancer Group

The Danish Colorectal Cancer Group (DCCG) database registered all patients in Denmark who were diagnosed with colorectal cancer or treated in a public Danish hospital since 2001 [3]. The original data was obtained from the database with a total of 12,855 observations. After data preprocessing, we draw a random sample of 700 observations for analysis. The outcome of interest is whether a patient had a postoperative medical complication, and associated risk factors include age, gender, ASA score (i.e., pre-operative fitness score), localization of the tumor, procedure, pathological tumor (T) stage and node (N) stage, number of removed lymph nodes and number of lymph nodes with metastasis and unplanned intraoperative adverse event. The statistics descriptives for the chosen DCCG variables is displayed in Table E.15.

| Variable | Description | Type | Mean (SD) or level count (% of total size) | Missingness (% of the total size) |
|---|---|---|---|---|
| Postoperative medical complication | Whether a patient had an unwanted medical condition (outcome) | Categorical | 1: 16.26%<br>0: 83.74% | 0.00% |
| Age | Patient's age | Numeric | 72.44 (9.91) | 0.00% |
| Gender | Patient's gender | Categorical | Female: 55.40%<br>Male: 44.60% | 0.00% |
| ASA score | Score to evaluate the overall health status according to the American Society of Anesthesiologists Physical Status Classification System | Categorical | ASA1: 17.44%<br>ASA2: 54.79%<br>ASA3: 24.96%<br>ASA4: 1.26% | 1.55% |
| Localization of the tumor | A place in the body where the tumor was located | Categorical | Ascending colon: 34.41%<br>Caecum: 40.17%<br>Hepatic flexure: 14.15%<br>Splenic flexure: 0.08%<br>Transverse colon: 11.19% | 0.00% |
| Procedure | Surgical procedure that was performed | Categorical | Right hemicolectomy: 85.52%<br>Extended right hemicolectomy: 14.48% | 0.00% |
| Pathological T stage | Measure to classify the extent of cancer (T stage) by the TNM staging system | Categorical | pT1 stage: 5.60%<br>pT2 stage: 13.12%<br>pT3 stage: 60.32%<br>pT4 stage: 18.39%<br>pTx or pT0 stage: 0.58% | 1.99% |
| Pathological N stage | Measure to classify the extent of cancer spread to regional lymph nodes (N stage) by the TNM staging system | Categorical | pN1 stage: 21.62%<br>pN2 stage: 14.83%<br>pNx or pN0 stage: 63.16% | 0.39% |
| Number of removed lymph nodes | Pathologically shown total number of removed lymph nodes | Numeric | 39.05 (108.69) | 0.00% |
| Number of lymph nodes with metastasis | Pathologically shown total number of lymph nodes with metastasis | Numeric | 1.71 (4.14) | 1.26% |
| Unplanned intraoperative adverse event | Whether there was an unexpected intraoperative adverse event | Categorical | Yes: 1.91%<br>No: 91.53% | 6.56% |

Note: SD: standard deviation

Table E.15: A summary of descriptive statistics for the DCCG dataset.

### E.16   Breast Cancer Coimbra

The Breast Cancer Coimbra dataset comprises the health information of women with breast cancer measured by the Gynaecology Department of the University Hospital Centre of Coimbra in the years from 2009 to 2013 [4]. It has 116 observations with 10 variables. The outcome describes whether the female patient is with breast cancer. The remaining 9 variables include patient's age, body mass index, glucose, insulin, homeostasis model assessment, leptin, adiponectin, resistin and MCP-1. The detailed characteristics of the variables are described in Table E.16.

| Variable | Description | Type | Mean (SD) or level count (% of total size) | Missingness (% of the total size) |
|---|---|---|---|---|
| Outcome | Whether a patient had breast cancer | Categorical | 1: 44.83% <br> 0: 55.17% | 0.00% |
| Age | Patient's age | Numeric | 57.30 (16.11) | 0.00% |
| Body mass index | Patient's body mass index | Numeric | 27.58 (5.02) | 0.00% |
| Glucose | Patient's serum glucose | Numeric | 97.79 (22.53) | 0.00% |
| Insulin | Patient's serum insulin | Numeric | 10.01 (10.07) | 0.00% |
| Homeostasis model assessment | An index to evaluate insulin resistance | Numeric | 2.69 (3.64) | 0.00% |
| Leptin | Serum values of Leptin | Numeric | 26.62 (19.18) | 0.00% |
| Adiponectin | Serum values of Adiponectin | Numeric | 10.18 (6.84) | 0.00% |
| Resistin | Serum values of Resistin | Numeric | 14.73 (12.39) | 0.00% |
| MCP-1 | Serum values of Chemokine Monocyte Chemoattractant Protein 1 | Numeric | 534.65 (345.91) | 0.00% |

Note: SD: standard deviation

Table E.16: A summary of descriptive statistics for the Breast Cancer Coimbra dataset.

### E.17   Breast Cancer

The Breast Cancer dataset is a dataset related to the recurrence of breast cancer patients, provided by the University Medical Centre, Institute of Oncology in Yugoslavia [5]. The dataset consists of 277 health records. The target variable is whether a patient experienced breast cancer recurrence and the relevant factors cover the patient's age, menopausal status, tumor size, node caps, breast, breast quad, irradiation, lymph nodes, and degree of malignancy. The full description of the variables is in Table E.17.

| Variable | Description | Type | Mean (SD) or level count (% of total size) | Missingness (% of the total size) |
|---|---|---|---|---|
| Outcome | Whether a patient had a recurrence event | Categorical | 1: 29.24%<br>0: 70.76% | 0.00% |
| Age | Patient's age | Numeric | 3.64 (1.01) | 0.00% |
| Menopausal status | Patient's menopausal status | Categorical | Ge40: 44.40%<br>Lt40: 1.81%<br>Premeno: 53.79% | 0.00% |
| Tumor size | Size of a tumor | Numeric | 0-4: 2.89%<br>5-9: 1.44%<br>10-14: 10.11%<br>15-19: 10.47%<br>20-24: 17.33%<br>25-29: 18.41%<br>30-34: 20.58%<br>35-39: 6.86%<br>40-44: 7.94%<br>45-49: 1.08%<br>50-54: 2.89% | 0.00% |
| Node caps | Capsule of a lymph node | Categorical | Yes: 20.22%<br>No: 79.78%s | 0.00% |
| Breast | Location of the breast | Categorical | Left: 52.35%<br>Right: 47.65% | 0.00% |
| Breast quad | Division of the breast into four sections or quadrants | Categorical | Central: 7.58%<br>Left low: 38.27%<br>Left up: 33.94%<br>Right low: 8.30%<br>Right up: 11.91% | 0.00% |
| Irradiation | Whether a patient undergone radiation therapy | Categorical | Yes: 22.38%<br>No: 77.62% | 0.00% |
| lymph nodes | Involved lymph nodes | Numeric | 0-2: 75.45%<br>6-8: 20.94%<br>12-14: 1.08%<br>15-17: 2.17%<br>24-26: 0.36% | 0.00% |
| Degree of malignancy | Severity of a malignant tumor | Numeric | 1: 23.83%<br>2: 46.57%<br>3: 29.60% | 0.00% |

Note: SD: standard deviation

**Table E.17:** A summary of descriptive statistics for the Breast Cancer dataset.

## E.18 Colposcopy/Schiller

The Colposcopy/Schiller dataset is one of the three modality colposcopy data that evaluates the subjective quality assessment of digital colposcopies collected by Hospital Universitario de Caracas [6].

The data have 287 health records. The outcome is whether the subjective assessment is good or not. The remaining 62 variables are associated with color information, image area and coverage in digital cervical imaging. Table E.18 summarizes the description of the dataset variables.

| Variable | Description | Type | Mean (SD) or level count (% of total size) | Missingness (% of the total size) |
|---|---|---|---|---|
| Outcome | Whether the subjective assessment is good | Categorical | 1: 72.83% 0: 23.17% | 0.00% |
| Cervix area | Image area with cervix | Numeric | 0.49 (0.24) | 0.00% |
| Os area | Image area with external os | Numeric | 0.01 (0.01) | 0.00% |
| Walls area | Image area with vaginal walls | Numeric | 0.18 (0.19) | 0.00% |
| Speculum area | Image area with the speculum | Numeric | 0.26 (0.19) | 0.00% |
| Artifacts area | Image area with artifacts | Numeric | 0.04 (0.04) | 0.00% |
| Cervix artifacts area | Cervix area with the artifacts | Numeric | 0.03 (0.03) | 0.00% |
| Os artifacts area | External os area with the artifacts | Numeric | 0.04 (0.13) | 0.00% |
| Walls artifacts area | Vaginal walls with the artifacts | Numeric | 0.03 (0.08) | 0.00% |
| Speculum artifacts area | Speculum area with the artifacts | Numeric | 0.01 (0.06) | 0.00% |
| Cervix specularities area | Cervix area with the specular reflections | Numeric | 0.01 (0.02) | 0.00% |
| Os specularities area | External os area with the specular reflections | Numeric | 0.01 (0.02) | 0.00% |
| Walls specularities area | Vaginal walls area with the specular reflections | Numeric | 0.02 (0.05) | 0.00% |
| Speculum specularities area | Speculum area with the specular reflections | Numeric | 0.14 (0.15) | 0.00% |
| specularities area | Total area with specular reflections | Numeric | 0.05 (0.05) | 0.00% |
| Area h max diff | Maximum area differences between the four cervix quadrants | Numeric | 0.21 (0.16) | 0.00% |
| Rgb cervix r mean | Average color information in the cervix (R channel) | Numeric | 48.39 (26.78) | 0.00% |
| Rgb cervix r std | Stddev color information in the cervix (R channel) | Numeric | 33.28 (14.89) | 0.00% |
| Rgb cervix r mean minus std | (Avg - stddev) color information in the cervix (R channel) | Numeric | 15.11 (24.53) | 0.00% |
| Rgb cervix r mean plus std | (Avg + stddev) information in the cervix (R channel) | Numeric | 81.67 (35.72) | 0.00% |
| Rgb cervix g mean | Average color information in the cervix (G channel) | Numeric | 41.70 (23.03) | 0.00% |
| Rgb cervix g std | Stddev color information in the cervix (G channel) | Numeric | 30.92 (15.34) | 0.00% |
| Rgb cervix g mean | (Avg - stddev) color | Numeric | 10.78 (19.12) | 0.00% |
| minus std | information in the cervix (G channel) | | | |
| Rgb cervix g mean plus std | (Avg + stddev) information in the cervix (G channel) | Numeric | 72.62 (34.15) | 0.00% |
| Rgb cervix b mean | Average color information in the cervix (B channel) | Numeric | 84.10 (43.83) | 0.00% |
| Rgb cervix b std | Stddev color information in the cervix (B channel) | Numeric | 47.43 (20.61) | 0.00% |
| Rgb cervix b mean minus std | (Avg - stddev) color information in the cervix (B channel) | Numeric | 36.67 (31.48) | 0.00% |
| Rgb cervix b mean plus std | (Avg + stddev) information in the cervix (B channel) | Numeric | 131.54 (60.83) | 0.00% |
| Rgb total r mean | Average color information in the image (R channel) | Numeric | 75.09 (27.43) | 0.00% |
| Rgb total r std | Stddev color information in the image (R channel) | Numeric | 55.43 (19.24) | 0.00% |
| Rgb total r mean minus std | (Avg - stddev) color information in the image (R channel) | Numeric | 19.66 (22.76) | 0.00% |
| Rgb total r mean plus std | (Avg + stddev) color information in the image (R channel) | Numeric | 130.52 (41.56) | 0.00% |
| Rgb total g mean | Average color information in the image (G channel) | Numeric | 65.66 (26.87) | 0.00% |
| Rgb total g std | Stddev color information in the image (G channel) | Numeric | 50.91 (20.65) | 0.00% |
| Rgb total g mean minus std | (Avg - stddev) color information in the image (G channel) | Numeric | 14.75 (17.84) | 0.00% |
| Rgb total g mean plus std | (Avg + stddev) color information in the image (G channel) | Numeric | 116.57 (44.48) | 0.00% |
| Rgb total b mean | Average color information in the image (B channel) | Numeric | 102.45 (31.72) | 0.00% |
| Rgb total b std | Stddev color information in the image (B channel) | Numeric | 64.26 (16.25) | 0.00% |
| Rgb total b mean minus std | (Avg - stddev) color information in the image (B channel) | Numeric | 38.18 (26.79) | 0.00% |
| Rgb total b mean plus std | (Avg + stddev) color information in the image (B channel) | Numeric | 166.71 (42.70) | 0.00% |
| hsv cervix h mean | Average color information in the cervix (H channel) | Numeric | 3.92 (0.38) | 0.00% |
| hsv cervix h std | Stddev color information in the cervix (H channel) | Numeric | 2.44 (0.56) | 0.00% |
| hsv cervix s mean | Average color information in the cervix (S channel) | Numeric | 127.81 (51.19) | 0.00% |
| hsv cervix s std | Stddev color information in the cervix (S channel) | Numeric | 48.70 (23.93) | 0.00% |
| hsv cervix v mean | Average color information in the cervix (V channel) | Numeric | 86.19 (43.67) | 0.00% |
| hsv cervix v std | Stddev color information in the cervix (V channel) | Numeric | 47.20 (20.44) | 0.00% |
| hsv total h mean | Average color information in the image (H channel) | Numeric | 4.00 (0.33) | 0.00% |
| hsv total h std | Stddev color information in the image (H channel) | Numeric | 2.51 (0.19) | 0.00% |
| hsv total s mean | Average color information in the image (S channel) | Numeric | 111.00 (42.07) | 0.00% |
| hsv total s std | Stddev color information in the image (S channel) | Numeric | 53.33 (20.95) | 0.00% |
| hsv total v mean | Average color information in the image (V channel) | Numeric | 106.63 (32.58) | 0.00% |
| hsv total v std | Stddev color information in the image (V channel) | Numeric | 64.24 (16.39) | 0.00% |
| Fit cervix hull rate | Coverage of the cervix convex hull by the cervix | Numeric | 0.90 (0.20) | 0.00% |
| Fit cervix hull total | Image coverage of the cervix convex hull | Numeric | 0.52 (0.24) | 0.00% |
| Fit cervix bbox rate | Coverage of the cervix bounding box by the cervix | Numeric | 0.77 (0.19) | 0.00% |
| Fit cervix bbox total | Image coverage of the cervix bounding box | Numeric | 0.60 (0.25) | 0.00% |
| Fit circle rate | Coverage of the cervix circle by the cervix | Numeric | 0.57 (0.14) | 0.00% |
| Fit circle total | Image coverage of the cervix circle | Numeric | 0.84 (0.44) | 0.00% |
| Fit ellipse rate | Coverage of the cervix ellipse by the cervix | Numeric | 0.93 (0.21) | 0.00% |
| Fit ellipse total | Image coverage of the cervix ellipse | Numeric | 0.50 (0.23) | 0.00% |
| Fit ellipse goodness | Goodness of the ellipse fitting | Numeric | 135.51 (87.55) | 0.00% |
| Distance to center cervix | Distance between the cervix center and the image center | Numeric | 0.50 (0.26) | 0.00% |
| Distance to center os | Distance between the cervical os center and the image center | Numeric | 0.46 (0.17) | 0.00% |

Note: SD: standard deviation

**Table E.18:** A summary of descriptive statistics for the Colposcopy/Schiller dataset.

### E.19  Diabetic Retinopathy

The Diabetic Retinopathy dataset comprises the detections of diabetic retinopathy represented by either detected lesions or image-level descriptors from Messidor image set [7]. The data size is 1,151. The outcome in the selected sample is whether a patient's image contains signs of diabetic retinopathy, and the other imaging information such as the numbers of the detection of microaneurysms found across various confidence levels, are considered as the relevant factors. and we randomly draw a sample with a size of 600. The outcome in the selected sample is whether  Table E.19 summarizes the statistics descriptives of the dataset.

| Variable | Description | Type | Mean (SD) or level count (% of total size) | Missingness (% of the total size) |
|---|---|---|---|---|
| Outcome | Whether a patient's image contains signs of diabetic retinopathy. | Categorical | 1: 54.67%<br>0: 45.33% | 0.00% |
| MA1 | Number of the detection of microaneurysms found at various confidence levels | Numeric | 39.67 (26.81) | 0.00% |
| MA2 | Number of the detection of microaneurysms found at various confidence levels | Numeric | 38.03 (25.03) | 0.00% |
| MA3 | Number of the detection of microaneurysms found at various confidence levels | Numeric | 36.17 (23.53) | 0.00% |
| MA4 | Number of the detection of microaneurysms found at various confidence levels | Numeric | 33.22 (21.64) | 0.00% |
| MA5 | Number of the detection of microaneurysms found at various confidence levels | Numeric | 29.58 (19.86) | 0.00% |
| MA6 | Number of the detection of microaneurysms found at various confidence levels | Numeric | 21.75 (15.27) | 0.00% |
| MA7 | Number of the detection of microaneurysms found at various confidence levels | Numeric | 65.14 (59.26) | 0.00% |
| Exu1 | Number of exudate pixels at various confidence levels | Numeric | 23.45 (22.00) | 0.00% |
| Exu2 | Number of exudate pixels at various confidence levels | Numeric | 8.92 (11.90) | 0.00% |
| Exu3 | Number of exudate pixels at various confidence levels | Numeric | 2.03 (4.51) | 0.00% |
| Exu4 | Number of exudate pixels at various confidence levels | Numeric | 0.70 (3.07) | 0.00% |
| Exu5 | Number of exudate pixels at various confidence levels | Numeric | 0.27 (1.31) | 0.00% |
| Exu6 | Number of exudate pixels at various confidence levels | Numeric | 0.11 (0.49) | 0.00% |
| Exu7 | Number of exudate pixels at various confidence levels | Numeric | 0.05 (0.22) | 0.00% |
| Euclidean distance | The Euclidean distance of the center of the macula and the center of the optic disk | Numeric | 0.52 (0.03) | 0.00% |
| Diameter of optic disk | The diameter of the optic disk | Numeric | 0.11 (0.02) | 0.00% |
| Quality | Quality assessment | Categorical | 1: 99.67% | 0.00% |

| | | | 0: 0.33% | |
|---|---|---|---|---|
| Pre-screening | Pre-screening for whether a patient had a severe retinal abnormality | Categorical | 1: 92.00%<br>0: 8.00% | 0.00% |
| AM_FM | Confidence of the detection of Diabetic Retinopathy indicated by the Amplitude-Modulation Frequency-Modulation-based classification | Categorical | 1: 33.67%<br>0: 66.33% | 0.00% |

Note: SD: standard deviation

**Table E.19:** A summary of descriptive statistics for the Diabetic Retinopathy dataset.

## E.20   Thoracic Surgery

The Thoracic Surgery dataset contains the post-operative life expectancy of the patients who underwent major lung resections between 2007 and 2011. The dataset was collected retrospectively by Wroclaw Thoracic Surgery Centre, which is associated with the Department of Thoracic Surgery of the Medical University of Wroclaw and Lower-Silesian Centre for Pulmonary Diseases, Poland, while the research database constitutes a part of the National Lung Cancer Registry, administered by the Institute of Tuberculosis and Pulmonary Diseases in Warsaw, Poland [8]. The dataset has 470 observations. The outcome is a binary variable indicating whether the patient survived after 1 year period. The remaining sixteen variables are diagnosis code, patient's age, forced vital capacity, exhaled volume, performance status, pain, haemoptysis, dyspnoea, coughing, weakness, tumor size, diabetes mellitus, peripheral arterial diseases, smoking and asthma. The detailed variable characteristics are displayed in Table E.20.

| Variable | Description | Type | Mean (SD) or level count (% of total size) | Missingness (% of the total size) |
|---|---|---|---|---|
| Outcome | Whether a patient died within one year after surgery | Categorical | 1: 14.89%<br>0: 85.11% | 0.00% |
| DGN | Diagnosis - specific combination of ICD-10 codes for primary and secondary as well multiple tumours if any | Categorical | DGN1: 0.21%<br>DGN2: 11.06%<br>DGN3: 74.26%<br>DGN4: 10.00%<br>DGN5: 3.19%<br>DGN6: 0.85%<br>DGN8: 0.43% | 0.00% |
| AGE | Patient's age s | Numeric | 62.53 (8.71) | 0.00% |
| PRE4 | Forced vital capacity | Numeric | 3.28 (0.87) | 0.00% |
| PRE5 | Volume that has been exhaled at the end of the first second of forced expiration | Numeric | 4.57 (11.77) | 0.00% |
| PRE6 | Performance status - Zubrod scale | Categorical | PRZ0: 27.66%<br>PRZ1: 66.60%<br>PRZ2: 5.74% | 0.00% |
| PRE7 | Pain before surgery | Categorical | 1: 6.60%<br>0: 93.40% | 0.00% |
| PRE8 | Haemoptysis before surgery | Categorical | 1: 14.47%<br>0: 85.53% | 0.00% |
| PRE9 | Dyspnoea before surgery | Categorical | 1: 6.60%<br>0: 93.40% | 0.00% |
| PRE10 | Cough before surgery | Categorical | 1: 68.72%<br>0: 31.28% | 0.00% |
| PRE11 | Weakness before surgery | Categorical | 1: 16.60%<br>0: 83.40% | 0.00% |
| PRE14 | T in clinical TNM - size of the original tumour, from OC11 (smallest) to OC14 (largest) | Categorical | OC11: 37.66%<br>OC12: 54.68%<br>OC13: 4.04%<br>OC14: 3.62% | 0.00% |
| PRE17 | Type 2 DM - diabetes mellitus | Categorical | 1: 7.45%<br>0: 92.55% | 0.00% |
| PRE19 | MI up to 6 months | Categorical | 1: 0.43%<br>0: 99.57% | 0.00% |
| PRE25 | PAD - peripheral arterial diseases | Categorical | 1: 1.70%<br>0: 98.30% | 0.00% |
| PRE30 | Smoking | Categorical | 1: 82.13%<br>0: 17.87% | 0.00% |
| PRE32 | Asthma | Categorical | 1: 0.43%<br>0: 99.57% | 0.00% |

Note: SD: standard deviation

**Table E.20:** A summary of descriptive statistics for the Thoracic Surgery dataset.

# F References